\documentclass{article}

\usepackage{microtype}
\usepackage{graphicx}
\usepackage{subfigure}
\usepackage{booktabs} %
\usepackage{multirow}
\usepackage{makecell}
\usepackage{capt-of}

\usepackage{hyperref}

\usepackage[accepted]{icml2020}

\icmltitlerunning{Learning Transferable Visual Models From Natural Language Supervision}

\begin{document}

\twocolumn[
\icmltitle{Learning Transferable Visual Models From Natural Language Supervision}

\icmlsetsymbol{equal}{*}

\begin{icmlauthorlist}
\icmlauthor{Alec Radford}{equal,openai}
\icmlauthor{Jong Wook Kim}{equal,openai}
\icmlauthor{Chris Hallacy}{openai}
\icmlauthor{Aditya Ramesh}{openai}
\icmlauthor{Gabriel Goh}{openai}
\icmlauthor{Sandhini Agarwal}{openai}
\icmlauthor{Girish Sastry}{openai}
\icmlauthor{Amanda Askell}{openai}
\icmlauthor{Pamela Mishkin}{openai}
\icmlauthor{Jack Clark}{openai}
\icmlauthor{Gretchen Krueger}{openai}
\icmlauthor{Ilya Sutskever}{openai}
\end{icmlauthorlist}

\icmlaffiliation{openai}{OpenAI, San Francisco, CA 94110, USA}

\icmlcorrespondingauthor{}{\{alec, jongwook\}@openai.com}

\icmlkeywords{}

\vskip 0.3in
]

\printAffiliationsAndNotice{\icmlEqualContribution} %

\begin{abstract}
State-of-the-art computer vision systems are trained to predict a fixed set of predetermined object categories. This restricted form of supervision limits their generality and usability since additional labeled data is needed to specify any other visual concept. Learning directly from raw text about images is a promising alternative which leverages a much broader source of supervision. We demonstrate that the simple pre-training task of predicting which caption goes with which image is an efficient and scalable way to learn SOTA image representations from scratch on a dataset of 400 million (image, text) pairs collected from the internet. After pre-training, natural language is used to reference learned visual concepts (or describe new ones) enabling zero-shot transfer of the model to downstream tasks. We study the performance of this approach by benchmarking on over 30 different existing computer vision datasets, spanning tasks such as OCR, action recognition in videos, geo-localization, and many types of fine-grained object classification. The model transfers non-trivially to most tasks and is often competitive with a fully supervised baseline without the need for any dataset specific training. For instance, we match the accuracy of the original ResNet-50 on ImageNet zero-shot without needing to use any of the 1.28 million training examples it was trained on.
We release our code and pre-trained model weights at \url{https://github.com/OpenAI/CLIP}.
\end{abstract}

\section{Introduction and Motivating Work}

Pre-training methods which learn directly from raw text have revolutionized NLP over the last few years \citep{dai2015semi,peters2018deep,howard2018universal,radford2018improving,devlin2018bert,raffel2019exploring}. Task-agnostic objectives such as autoregressive and masked language modeling have scaled across many orders of magnitude in compute, model capacity, and data, steadily improving capabilities. The development of ``text-to-text'' as a standardized input-output interface \citep{mccann2018natural,radford2019language,raffel2019exploring} has enabled task-agnostic architectures to zero-shot transfer to downstream datasets removing the need for specialized output heads or dataset specific customization. Flagship systems like GPT-3 \citep{brown2020language} are now competitive across many tasks with bespoke models while requiring little to no dataset specific training data.

These results suggest that the aggregate supervision accessible to modern pre-training methods within web-scale collections of text surpasses that of high-quality crowd-labeled NLP datasets. However, in other fields such as computer vision it is still standard practice to pre-train models on crowd-labeled datasets such as ImageNet \citep{imagenet_cvpr09}. Could scalable pre-training methods which learn directly from web text result in a similar breakthrough in computer vision? Prior work is encouraging.

\begin{figure*}[ht]
\begin{center}
\centerline{\includegraphics[width=\textwidth]{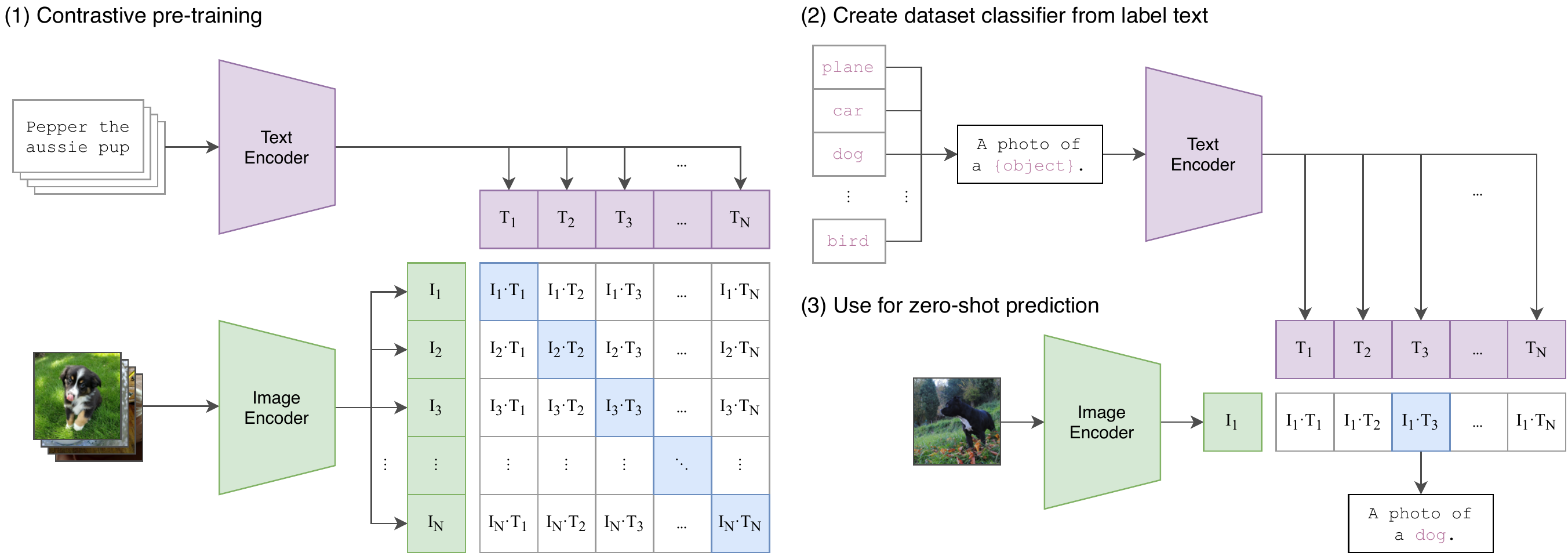}}
\caption{Summary of our approach. While standard image models jointly train an image feature extractor and a linear classifier to predict some label, CLIP jointly trains an image encoder and a text encoder to predict the correct pairings of a batch of (image, text) training examples. At test time the learned text encoder synthesizes a zero-shot linear classifier by embedding the names or descriptions of the target dataset's classes.}
\label{main_fig}
\end{center}
\vspace{-1em}
\end{figure*}

Over 20 years ago \citet{mori1999image} explored improving content based image retrieval by training a model to predict the nouns and adjectives in text documents paired with images. \citet{quattoni2007learning} demonstrated it was possible to learn more data efficient image representations via manifold learning in the weight space of classifiers trained to predict words in captions associated with images. \citet{srivastava2012multimodal} explored deep representation learning by training multimodal Deep Boltzmann Machines on top of low-level image and text tag features. \citet{joulin2016learning} modernized this line of work and demonstrated that CNNs trained to predict words in image captions learn useful image representations. They converted the title, description, and hashtag metadata of images in the YFCC100M dataset \citep{thomee2016yfcc100m} into a bag-of-words multi-label classification task and showed that pre-training AlexNet \citep{krizhevsky2012imagenet} to predict these labels learned representations which preformed similarly to ImageNet-based pre-training on transfer tasks. \citet{li2017learning} then extended this approach to predicting phrase n-grams in addition to individual words and demonstrated the ability of their system to zero-shot transfer to other image classification datasets by scoring target classes based on their dictionary of learned visual n-grams and predicting the one with the highest score. Adopting more recent architectures and pre-training approaches, VirTex \citep{desai2020virtex}, ICMLM \citep{bulent2020learning}, and ConVIRT \citep{zhang2020contrastive} have recently demonstrated the potential of transformer-based language modeling, masked language modeling, and contrastive objectives to learn image representations from text.

While exciting as proofs of concept, using natural language supervision for image representation learning is still rare. This is likely because demonstrated performance on common benchmarks is much lower than alternative approaches. For example, \citet{li2017learning} reach only 11.5\% accuracy on ImageNet in a zero-shot setting. This is well below the 88.4\% accuracy of the current state of the art \citep{xie2020self}. It is even below the 50\% accuracy of classic computer vision approaches \citep{ILSVRC2012}. Instead, more narrowly scoped but well-targeted uses of weak supervision have improved performance. \citet{mahajan2018exploring} showed that predicting ImageNet-related hashtags on Instagram images is an effective pre-training task. When fine-tuned to ImageNet these pre-trained models increased accuracy by over 5\% and improved the overall state of the art at the time. \citet{kolesnikov2019large} and \citet{dosovitskiy2020image} have also demonstrated large gains on a broader set of transfer benchmarks by pre-training models to predict the classes of the noisily labeled JFT-300M dataset.

This line of work represents the current pragmatic middle ground between learning from a limited amount of supervised ``gold-labels'' and learning from practically unlimited amounts of raw text. However, it is not without compromises. Both works carefully design, and in the process limit, their supervision to 1000 and 18291 classes respectively. Natural language is able to express, and therefore supervise, a much wider set of visual concepts through its generality. Both approaches also use static softmax classifiers to perform prediction and lack a mechanism for dynamic outputs. This severely curtails their flexibility and limits their ``zero-shot'' capabilities.

A crucial difference between these weakly supervised models and recent explorations of learning image representations directly from natural language is scale. While \citet{mahajan2018exploring} and \citet{kolesnikov2019large} trained their models for accelerator years on millions to billions of images, VirTex, ICMLM, and ConVIRT trained for accelerator days on one to two hundred thousand images. In this work, we close this gap and study the behaviors of image classifiers trained with natural language supervision at large scale. Enabled by the large amounts of publicly available data of this form on the internet, we create a new dataset of 400 million (image, text) pairs and demonstrate that a simplified version of ConVIRT trained from scratch, which we call CLIP, for Contrastive Language-Image Pre-training, is an efficient method of learning from natural language supervision. We study the scalability of CLIP by training a series of eight models spanning almost 2 orders of magnitude of compute and observe that transfer performance is a smoothly predictable function of compute \citep{hestness2017deep,kaplan2020scaling}. We find that CLIP, similar to the GPT family, learns to perform a wide set of tasks during pre-training including OCR, geo-localization, action recognition, and many others. We measure this by benchmarking the zero-shot transfer performance of CLIP on over 30 existing datasets and find it can be competitive with prior task-specific supervised models. We also confirm these findings with linear-probe representation learning analysis and show that CLIP outperforms the best publicly available ImageNet model while also being more computationally efficient. We additionally find that zero-shot CLIP models are much more robust than equivalent accuracy supervised ImageNet models which suggests that zero-shot evaluation of task-agnostic models is much more representative of a model's capability. These results have significant policy and ethical implications, which we consider in Section \ref{sec:broader-impacts}.

\begin{figure}[t]
\begin{center}
\centerline{\includegraphics[width=1.0\columnwidth]{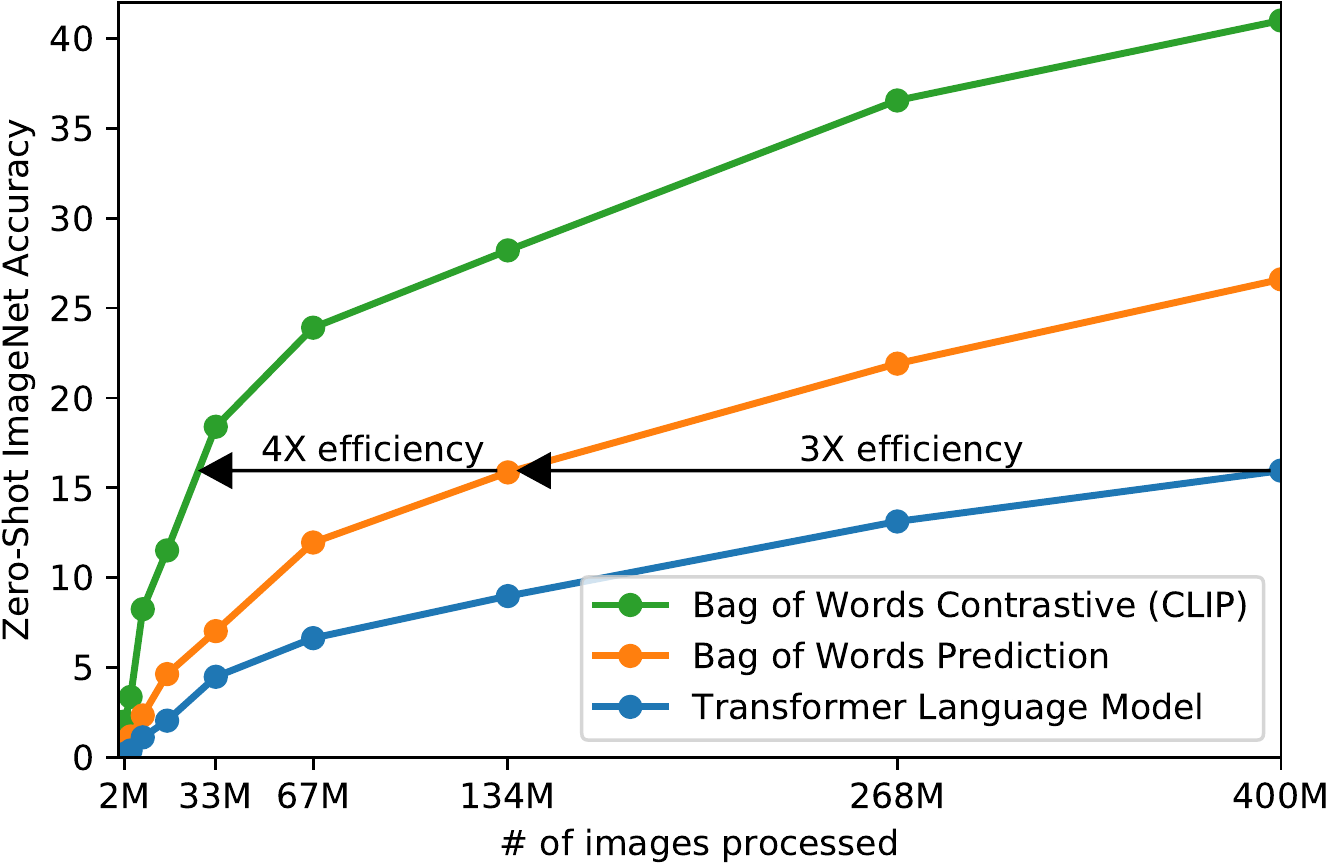}}
\caption{\textbf{CLIP is much more efficient at zero-shot transfer than our image caption baseline.} Although highly expressive, we found that transformer-based language models are relatively weak at zero-shot ImageNet classification. Here, we see that it learns 3x slower than a baseline which predicts a bag-of-words (BoW) encoding of the text \citep{joulin2016learning}. Swapping the prediction objective for the contrastive objective of CLIP further improves efficiency another 4x.}
\label{compare_objective_fig}
\end{center}
\vspace{-1em}
\end{figure}

\section{Approach}

\subsection{Natural Language Supervision}

At the core of our approach is the idea of learning perception from supervision contained in natural language. As discussed in the introduction, this is not at all a new idea, however terminology used to describe work in this space is varied, even seemingly contradictory, and stated motivations are diverse. \citet{zhang2020contrastive}, \citet{gomez2017self}, \citet{joulin2016learning}, and \citet{desai2020virtex} all introduce methods which learn visual representations from text paired with images but describe their approaches as unsupervised, self-supervised, weakly supervised, and supervised respectively.

We emphasize that what is common across this line of work is not any of the details of the particular methods used but the appreciation of natural language as a training signal. All these approaches are learning from \textit{natural language supervision}. Although early work wrestled with the complexity of natural language when using topic model and n-gram representations, improvements in deep contextual representation learning suggest we now have the tools to effectively leverage this abundant source of supervision \citep{mccann2017learned}.

Learning from natural language has several potential strengths over other training methods. It's much easier to scale natural language supervision compared to standard crowd-sourced labeling for image classification since it does not require annotations to be in a classic ``machine learning compatible format'' such as the canonical 1-of-N majority vote ``gold label''. Instead, methods which work on natural language can learn passively from the supervision contained in the vast amount of text on the internet. Learning from natural language also has an important advantage over most unsupervised or self-supervised learning approaches in that it doesn't ``just'' learn a representation but also connects that representation to language which enables flexible zero-shot transfer. In the following subsections, we detail the specific approach we settled on.

\subsection{Creating a Sufficiently Large Dataset}\label{subsection:creating-large-dataset}

Existing work has mainly used three datasets, MS-COCO \citep{lin2014microsoft}, Visual Genome \citep{krishna2017visual}, and YFCC100M \citep{thomee2016yfcc100m}. While MS-COCO and Visual Genome are high quality crowd-labeled datasets, they are small by modern standards with approximately 100,000 training photos each. By comparison, other computer vision systems are trained on up to 3.5 \textit{billion} Instagram photos \citep{mahajan2018exploring}. YFCC100M, at 100 million photos, is a possible alternative, but the metadata for each image is sparse and of varying quality. Many images use automatically generated filenames like \texttt{20160716\_113957.JPG} as ``titles'' or contain ``descriptions'' of camera exposure settings. After filtering to keep only images with natural language titles and/or descriptions in English, the dataset shrunk by a factor of 6 to only 15 million photos. This is approximately the same size as ImageNet.

A major motivation for natural language supervision is the large quantities of data of this form available publicly on the internet. Since existing datasets do not adequately reflect this possibility, considering results only on them would underestimate the potential of this line of research. To address this, we constructed a new dataset of 400 million (image, text) pairs collected form a variety of publicly available sources on the Internet. To attempt to cover as broad a set of visual concepts as possible, we search for (image, text) pairs as part of the construction process whose text includes one of a set of 500,000 queries.\footnote{The base query list is all words occurring at least 100 times in the English version of Wikipedia. This is augmented with bi-grams with high pointwise mutual information as well as the names of all Wikipedia articles above a certain search volume. Finally all WordNet synsets not already in the query list are added.} We approximately class balance the results by including up to 20,000 (image, text) pairs per query. The resulting dataset has a similar total word count as the WebText dataset used to train GPT-2. We refer to this dataset as WIT for WebImageText.

\subsection{Selecting an Efficient Pre-Training Method}
\label{subsection:method}

State-of-the-art computer vision systems use very large amounts of compute. \citet{mahajan2018exploring} required 19 GPU years to train their ResNeXt101-32x48d and \citet{xie2020self} required 33 TPUv3 core-years to train their Noisy Student EfficientNet-L2. When considering that both these systems were trained to predict only 1000 ImageNet classes, the task of learning an open set of visual concepts from natural language seems daunting. In the course of our efforts, we found training efficiency was key to successfully scaling natural language supervision and we selected our final pre-training method based on this metric.

Our initial approach, similar to VirTex, jointly trained an image CNN and text transformer from scratch to predict the caption of an image. However, we encountered difficulties efficiently scaling this method. In Figure \ref{compare_objective_fig} we show that a 63 million parameter transformer language model, which already uses twice the compute of its ResNet-50 image encoder, learns to recognize ImageNet classes three times slower than a much simpler baseline that predicts a bag-of-words encoding of the same text.

Both these approaches share a key similarity. They try to predict the \textit{exact} words of the text accompanying each image. This is a difficult task due to the wide variety of descriptions, comments, and related text that co-occur with images. Recent work in contrastive representation learning for images has found that contrastive objectives can learn better representations than their equivalent predictive objective \citep{tian2019contrastive}. Other work has found that although generative models of images can learn high quality image representations, they require over an order of magnitude more compute than contrastive models with the same performance \citep{chen2020generative}. Noting these findings, we explored training a system to solve the potentially easier proxy task of predicting only which text \textit{as a whole} is paired with which image and not the exact words of that text. Starting with the same bag-of-words encoding baseline, we swapped the predictive objective for a contrastive objective in Figure \ref{compare_objective_fig} and observed a further 4x efficiency improvement in the rate of zero-shot transfer to ImageNet.

\begin{figure}[t]
\begin{center}
\centerline{\includegraphics[width=1.0\columnwidth]{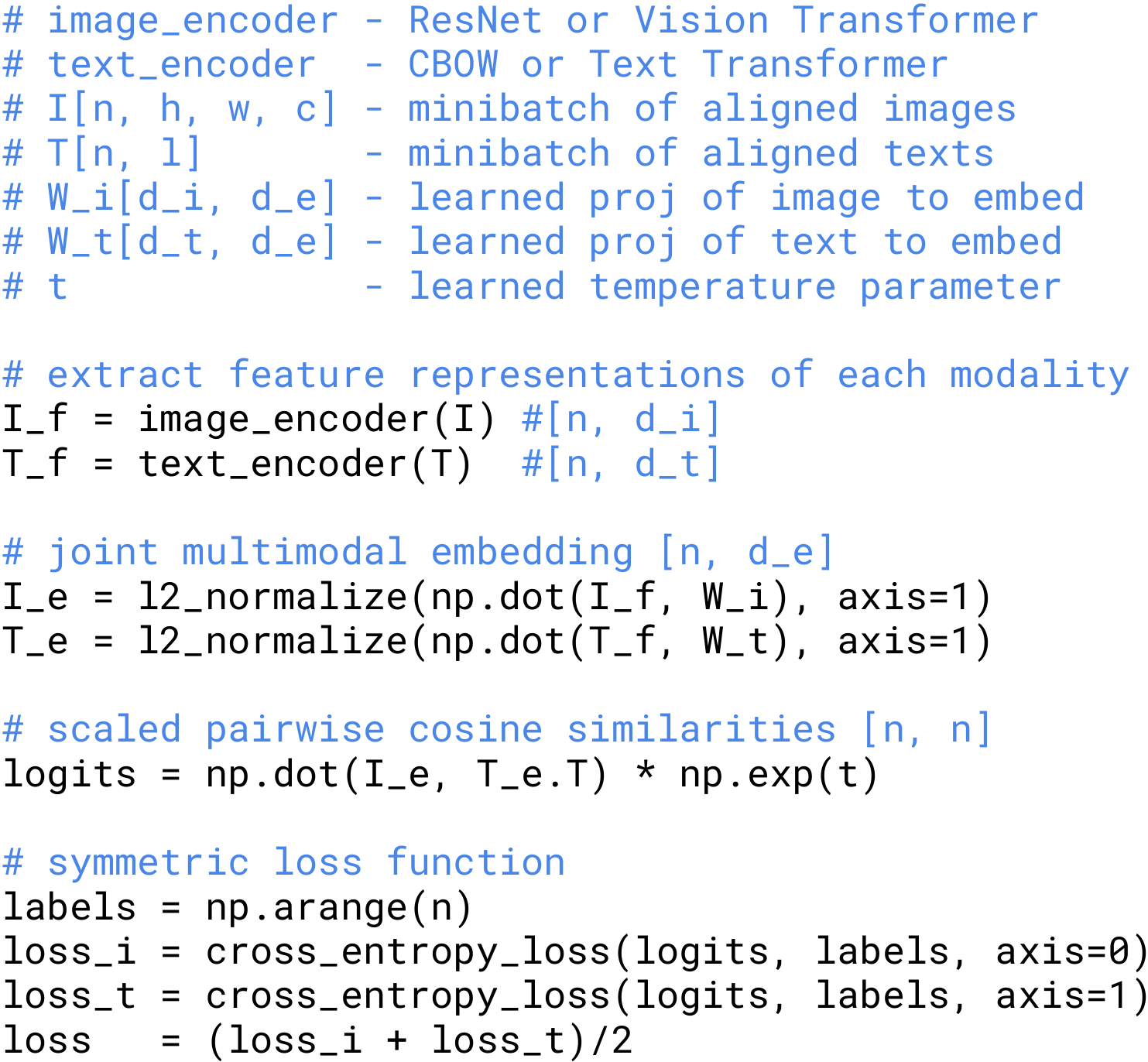}}
\caption{Numpy-like pseudocode for the core of an implementation of CLIP.}
\label{pseudocode}
\end{center}
\vspace{-1em}
\end{figure}

Given a batch of $N$ (image, text) pairs, CLIP is trained to predict which of the $N\times N$ possible (image, text) pairings across a batch actually occurred. To do this, CLIP learns a multi-modal embedding space by jointly training an image encoder and text encoder to maximize the cosine similarity of the image and text embeddings of the $N$ real pairs in the batch while minimizing the cosine similarity of the embeddings of the $N^2-N$ incorrect pairings. We optimize a symmetric cross entropy loss over these similarity scores. In Figure \ref{pseudocode} we include pseudocode of the core of an implementation of CLIP. To our knowledge this batch construction technique and objective was first introduced in the area of deep metric learning as the \textit{multi-class N-pair loss} \citet{sohn2016improved}, was popularized for contrastive representation learning by \citet{oord2018representation} as the InfoNCE loss, and was recently adapted for contrastive (text, image) representation learning in the domain of medical imaging by \citet{zhang2020contrastive}.

Due to the large size of our pre-training dataset, over-fitting is not a major concern and the details of training CLIP are simplified compared to the implementation of \citet{zhang2020contrastive}. We train CLIP from scratch without initializing the image encoder with ImageNet weights or the text encoder with pre-trained weights. We do not use the non-linear projection between the representation and the contrastive embedding space, a change which was introduced by \citet{bachman2019learning} and popularized by \citet{chen2020simple}. We instead use only a linear projection to map from each encoder's representation to the multi-modal embedding space. We did not notice a difference in training efficiency between the two versions and speculate that non-linear projections may be co-adapted with details of current image only in self-supervised representation learning methods. We also remove the text transformation function $t_u$ from \citet{zhang2020contrastive} which samples a single sentence at uniform from the text since many of the (image, text) pairs in CLIP's pre-training dataset are only a single sentence. We also simplify the image transformation function $t_v$. A random square crop from resized images is the only data augmentation used during training. Finally, the temperature parameter which controls the range of the logits in the softmax, $\tau$, is directly optimized during training as a log-parameterized multiplicative scalar to avoid turning as a hyper-parameter.

\subsection{Choosing and Scaling a Model}

We consider two different architectures for the image encoder. For the first, we use ResNet-50 \citep{he2016deep} as the base architecture for the image encoder due to its widespread adoption and proven performance. We make several modifications to the original version using the ResNet-D improvements from \citet{he2019bag} and the antialiased rect-2 blur pooling from \citet{zhang2019making}. We also replace the global average pooling layer with an attention pooling mechanism. The attention pooling is implemented as a single layer of ``transformer-style'' multi-head QKV attention where the query is conditioned on the global average-pooled representation of the image. For the second architecture, we experiment with the recently introduced Vision Transformer (ViT) \citep{dosovitskiy2020image}. We closely follow their implementation with only the minor modification of adding an additional layer normalization to the combined patch and position embeddings before the transformer and use a slightly different initialization scheme.

The text encoder is a Transformer \citep{vaswani2017attention} with the architecture modifications described in \citet{radford2019language}. As a base size we use a 63M-parameter 12-layer 512-wide model with 8 attention heads. The transformer operates on a lower-cased byte pair encoding (BPE) representation of the text with a 49,152 vocab size \citep{sennrich2015neural}. For computational efficiency, the max sequence length was capped at 76. The text sequence is bracketed with \texttt{[SOS]} and \texttt{[EOS]} tokens and the activations of the highest layer of the transformer at the \texttt{[EOS]} token are treated as the feature representation of the text which is layer normalized and then linearly projected into the multi-modal embedding space. Masked self-attention was used in the text encoder to preserve the ability to initialize with a pre-trained language model or add language modeling as an auxiliary objective, though exploration of this is left as future work.

While previous computer vision research has often scaled models by increasing the width \citep{mahajan2018exploring} or depth \citep{he2016deep} in isolation, for the ResNet image encoders we adapt the approach of \citet{tan2019efficientnet} which found that allocating additional compute across all of width, depth, and resolution outperforms only allocating it to only one dimension of the model. While \citet{tan2019efficientnet} tune the ratio of compute allocated to each dimension for their EfficientNet architecture, we use a simple baseline of allocating additional compute equally to increasing the width, depth, and resolution of the model. For the text encoder, we only scale the width of the model to be proportional to the calculated increase in width of the ResNet and do not scale the depth at all, as we found CLIP's performance to be less sensitive to the capacity of the text encoder.

\subsection{Training}

We train a series of 5 ResNets and 3 Vision Transformers. For the ResNets we train a ResNet-50, a ResNet-101, and then 3 more which follow EfficientNet-style model scaling and use approximately 4x, 16x, and 64x the compute of a ResNet-50. They are denoted as RN50x4, RN50x16, and RN50x64 respectively. For the Vision Transformers we train a ViT-B/32, a ViT-B/16, and a ViT-L/14. We train all models for 32 epochs. We use the Adam optimizer \citep{kingma2014adam} with decoupled weight decay regularization \citep{loshchilov2017decoupled} applied to all weights that are not gains or biases, and decay the learning rate using a cosine schedule \citep{loshchilov2016sgdr}. Initial hyper-parameters were set using a combination of grid searches, random search, and manual tuning on the baseline ResNet-50 model when trained for 1 epoch. Hyper-parameters were then adapted heuristically for larger models due to computational constraints. The learnable temperature parameter $\tau$ was initialized to the equivalent of 0.07 from \citep{wu2018unsupervised} and clipped to prevent scaling the logits by more than 100 which we found necessary to prevent training instability. We use a very large minibatch size of 32,768. Mixed-precision \citep{micikevicius2017mixed} was used to accelerate training and save memory. To save additional memory, gradient checkpointing \citep{griewank2000algorithm,chen2016training}, half-precision Adam statistics \citep{dhariwal2020jukebox}, and half-precision stochastically rounded text encoder weights were used. The calculation of embedding similarities was also sharded with individual GPUs computing only the subset of the pairwise similarities necessary for their local batch of embeddings. The largest ResNet model, RN50x64, took 18 days to train on 592 V100 GPUs while the largest Vision Transformer took 12 days on 256 V100 GPUs. For the ViT-L/14 we also pre-train at a higher 336 pixel resolution for one additional epoch to boost performance similar to FixRes \citep{touvron2019fixing}. We denote this model as ViT-L/14@336px. Unless otherwise specified, all results reported in this paper as ``CLIP'' use this model which we found to perform best.

\section{Experiments}

\subsection{Zero-Shot Transfer}
\label{subsection:zero_shot_transfer}

\subsubsection{Motivation}

In computer vision, zero-shot learning usually refers to the study of generalizing to unseen object categories in image classification \citep{lampert2009learning}. We instead use the term in a broader sense and study generalization to unseen datasets. We motivate this as a proxy for performing unseen tasks, as aspired to in the zero-data learning paper of \citet{larochelle2008zero}. While much research in the field of unsupervised learning focuses on the \textit{representation learning} capabilities of machine learning systems, we motivate studying zero-shot transfer as a way of measuring the \textit{task-learning} capabilities of machine learning systems. In this view, a dataset evaluates performance on a task on a specific distribution. However, many popular computer vision datasets were created by the research community primarily as benchmarks to guide the development of generic image classification methods rather than measuring performance on a specific task. While it is reasonable to say that the SVHN dataset measures the task of street number transcription on the distribution of Google Street View photos, it is unclear what ``real'' task the CIFAR-10 dataset measures. It is clear, however, what distribution CIFAR-10 is drawn from - TinyImages \citep{torralba200880}. On these kinds of datasets, zero-shot transfer is more an evaluation of CLIP's robustness to distribution shift and domain generalization rather than task generalization. Please see Section \ref{subsection:robustness} for analysis focused on this.

To our knowledge, Visual N-Grams \citep{li2017learning} first studied zero-shot transfer to existing image classification datasets in the manner described above. It is also the only other work we are aware of that has studied zero-shot transfer to standard image classification datasets using a generically pre-trained model and serves as the best reference point for contextualizing CLIP. Their approach learns the parameters of a dictionary of 142,806 visual n-grams (spanning 1- to 5- grams) and optimizes these n-grams using a differential version of Jelinek-Mercer smoothing to maximize the probability of all text n-grams for a given image. In order to perform zero-shot transfer, they first convert the text of each of the dataset's class names into its n-gram representation and then compute its probability according to their model, predicting the one with the highest score.

Our focus on studying zero-shot transfer as an evaluation of task learning is inspired by work demonstrating task learning in the field of NLP. To our knowledge \citet{liu2018generating} first identified task learning as an ``unexpected side-effect'' when a language model trained to generate Wikipedia articles learned to reliably transliterate names between languages. While GPT-1 \citep{radford2018improving} focused on pre-training as a transfer learning method to improve supervised fine-tuning, it also included an ablation study demonstrating that the performance of four heuristic zero-shot transfer methods improved steadily over the course of pre-training, without any supervised adaption. This analysis served as the basis for GPT-2 \citep{radford2019language} which focused exclusively on studying the task-learning capabilities of language models via zero-shot transfer.

\subsubsection{Using CLIP for Zero-Shot Transfer}

CLIP is pre-trained to predict if an image and a text snippet are paired together in its dataset. To perform zero-shot classification, we reuse this capability. For each dataset, we use the names of all the classes in the dataset as the set of potential text pairings and predict the most probable (image, text) pair according to CLIP. In a bit more detail, we first compute the feature embedding of the image and the feature embedding of the set of possible texts by their respective encoders. The cosine similarity of these embeddings is then calculated, scaled by a temperature parameter $\tau$, and normalized into a probability distribution via a softmax. Note that this prediction layer is a multinomial logistic regression classifier with L2-normalized inputs, L2-normalized weights, no bias, and temperature scaling. When interpreted this way, the image encoder is the computer vision backbone which computes a feature representation for the image and the text encoder is a hypernetwork \citep{ha2016hypernetworks} which generates the weights of a linear classifier based on the text specifying the visual concepts that the classes represent. \citet{lei2015predicting} first introduced a zero-shot image classifier of this form while the idea of generating a classifier from natural language dates back to at least \citet{elhoseiny2013write}. Continuing with this interpretation, every step of CLIP pre-training can be viewed as optimizing the performance of a randomly created proxy to a computer vision dataset which contains 1 example per class and has 32,768 total classes defined via natural language descriptions. For zero-shot evaluation, we cache the zero-shot classifier once it has been computed by the text encoder and reuse it for all subsequent predictions. This allows the cost of generating it to be amortized across all the predictions in a dataset.

\subsubsection{Initial Comparison to Visual N-Grams}

\begin{table}[t]
\vskip 0.15in
\begin{center}
\begin{tabular}{lccc}
\toprule
 & aYahoo & ImageNet & SUN \\
\midrule
Visual N-Grams & 72.4 & 11.5 &  23.0 \\
CLIP & \textbf{98.4} & \textbf{76.2} & \textbf{58.5} \\
\bottomrule
\end{tabular}
\caption{Comparing CLIP to prior zero-shot transfer image classification results. CLIP improves performance on all three datasets by a large amount. This improvement reflects many differences in the 4 years since the development of Visual N-Grams \citep{li2017learning}.}
\label{zeroshot_table}
\end{center}
\vspace{-1em}
\end{table}

In Table \ref{zeroshot_table} we compare Visual N-Grams to CLIP. The best CLIP model improves accuracy on ImageNet from a proof of concept 11.5\% to 76.2\% and matches the performance of the original ResNet-50 despite using none of the 1.28 million crowd-labeled training examples available for this dataset. Additionally, the top-5 accuracy of CLIP models are noticeably higher than their top-1, and this model has a 95\% top-5 accuracy, matching Inception-V4 \cite{szegedy2016inception}. The ability to match the performance of a strong, fully supervised baselines in a zero-shot setting suggests CLIP is a significant step towards flexible and practical zero-shot computer vision classifiers. As mentioned above, the comparison to Visual N-Grams is meant for contextualizing the performance of CLIP and should not be interpreted as a direct methods comparison between CLIP and Visual N-Grams as many performance relevant differences between the two systems were not controlled for. For instance, we train on a dataset that is 10x larger, use a vision model that requires nearly 100x more compute per prediction, likely used over 1000x their training compute, and use a transformer-based model which did not exist when Visual N-Grams was published. As a closer comparison, we trained a CLIP ResNet-50 on the same YFCC100M dataset that Visual N-Grams was trained on and found it matched their reported ImageNet performance within a V100 GPU day. This baseline was also trained from scratch instead of being initialized from pre-trained ImageNet weights as in Visual N-Grams.

CLIP also outperforms Visual N-Grams on the other 2 reported datasets. On aYahoo, CLIP achieves a 95\% reduction in the number of errors, and on SUN, CLIP more than doubles the accuracy of Visual N-Grams. To conduct a more comprehensive analysis and stress test, we implement a much larger evaluation suite detailed in Appendix \ref{sec:linear-probe}. In total we expand from the 3 datasets reported in Visual N-Grams to include over 30 datasets and compare to over 50 existing computer vision systems to contextualize results.

\subsubsection{Prompt Engineering and Ensembling}

\begin{figure}[t]
\begin{center}
\centerline{\includegraphics[width=1.0\columnwidth]{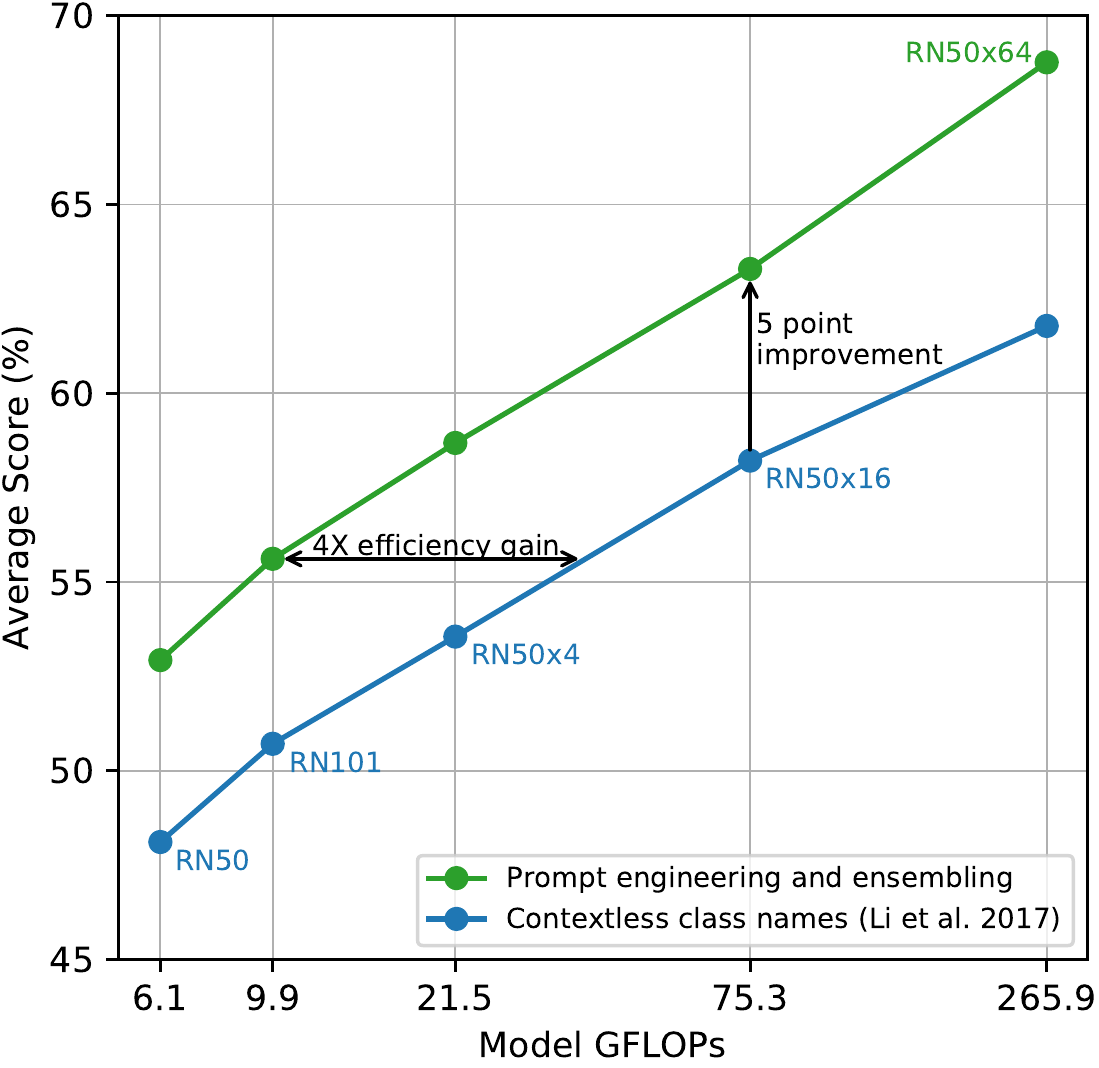}}
\caption{\textbf{Prompt engineering and ensembling improve zero-shot performance.} Compared to the baseline of using contextless class names, prompt engineering and ensembling boost zero-shot classification performance by almost 5 points on average across 36 datasets. This improvement is similar to the gain from using 4 times more compute with the baseline zero-shot method but is ``free'' when amortized over many predictions.}
\label{prompt_engineering}
\end{center}
\vspace{-1em}
\end{figure}

Most standard image classification datasets treat the information naming or describing classes which enables natural language based zero-shot transfer as an afterthought. The vast majority of datasets annotate images with just a numeric id of the label and contain a file mapping these ids back to their names in English. Some datasets, such as Flowers102 and GTSRB, don't appear to include this mapping at all in their released versions preventing zero-shot transfer entirely.\footnote{Alec learned much more about flower species and German traffic signs over the course of this project than he originally anticipated.} For many datasets, we observed these labels may be chosen somewhat haphazardly and do not anticipate issues related to zero-shot transfer which relies on task description in order to transfer successfully.

A common issue is polysemy. When the name of a class is the only information provided to CLIP's text encoder it is unable to differentiate which word sense is meant due to the lack of context. In some cases multiple meanings of the same word might be included as different classes in the same dataset! This happens in ImageNet which contains both construction cranes and cranes that fly. Another example is found in classes of the Oxford-IIIT Pet dataset where the word boxer is, from context, clearly referring to a breed of dog, but to a text encoder lacking context could just as likely refer to a type of athlete.

Another issue we encountered is that it's relatively rare in our pre-training dataset for the text paired with the image to be just a single word. Usually the text is a full sentence describing the image in some way. To help bridge this distribution gap, we found that using the prompt template ``\texttt{A photo of a \{label\}.}'' to be a good default that helps specify the text is about the content of the image. This often improves performance over the baseline of using only the label text. For instance, just using this prompt improves accuracy on ImageNet by 1.3\%.

Similar to the ``prompt engineering'' discussion around GPT-3 \citep{brown2020language,gao2020making}, we have also observed that zero-shot performance can be significantly improved by customizing the prompt text to each task. A few, non exhaustive, examples follow. We found on several fine-grained image classification datasets that it helped to specify the category. For example on Oxford-IIIT Pets, using ``\texttt{A photo of a \{label\}, a type of pet.}'' to help provide context worked well. Likewise, on Food101 specifying \textit{a type of food} and on FGVC Aircraft \textit{a type of aircraft} helped too. For OCR datasets, we found that putting quotes around the text or number to be recognized improved performance. Finally, we found that on satellite image classification datasets it helped to specify that the images were of this form and we use variants of ``\texttt{a satellite photo of a \{label\}.}''.

We also experimented with ensembling over multiple zero-shot classifiers as another way of improving performance. These classifiers are computed by using different context prompts such as `\texttt{A photo of a big \{label\}}'' and ``\texttt{A photo of a small \{label\}}''. We construct the ensemble over the embedding space instead of probability space. This allows us to cache a single set of averaged text embeddings so that the compute cost of the ensemble is the same as using a single classifier when amortized over many predictions. We've observed ensembling across many generated zero-shot classifiers to reliably improve performance and use it for the majority of datasets. On ImageNet, we ensemble 80 different context prompts and this improves performance by an additional 3.5\% over the single default prompt discussed above. When considered together, prompt engineering and ensembling improve ImageNet accuracy by almost 5\%. In Figure \ref{prompt_engineering} we visualize how prompt engineering and ensembling change the performance of a set of CLIP models compared to the contextless baseline approach of directly embedding the class name as done in \citet{li2017learning}. 

\subsubsection{Analysis of Zero-Shot CLIP Performance}

\begin{figure}[t]
\begin{center}
\centerline{\includegraphics[width=1.0\columnwidth]{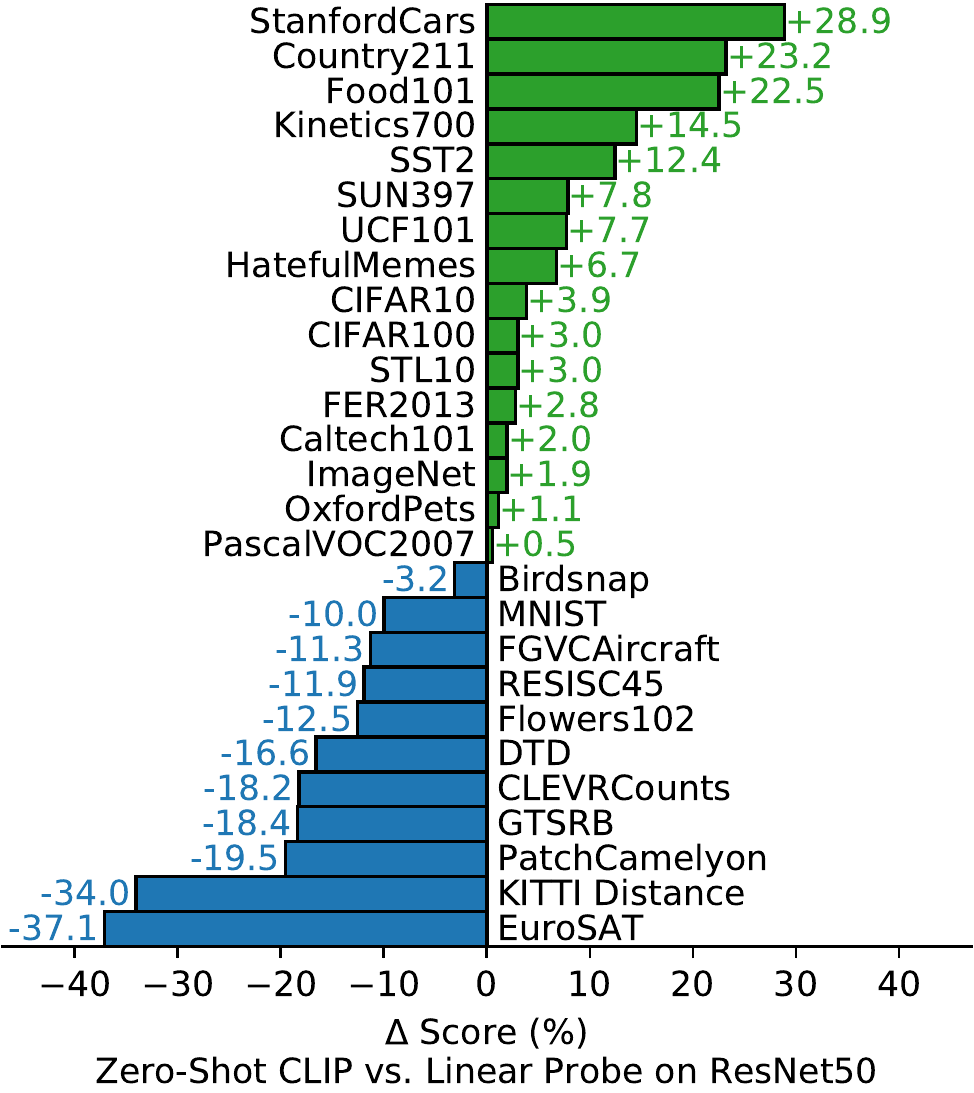}}
\caption{\textbf{Zero-shot CLIP is competitive with a fully supervised baseline.} Across a 27 dataset eval suite, a zero-shot CLIP classifier outperforms a fully supervised linear classifier fitted on ResNet-50 features on 16 datasets, including ImageNet.}
\label{zeroshot_vs_supervised}
\end{center}
\vspace{-1em}
\end{figure}

Since task-agnostic zero-shot classifiers for computer vision have been understudied, CLIP provides a promising opportunity to gain a better understanding of this type of model. In this section, we conduct a study of various properties of CLIP's zero-shot classifiers. As a first question, we look simply at how well zero-shot classifiers perform. To contextualize this, we compare to the performance of a simple off-the-shelf baseline: fitting a fully supervised, regularized, logistic regression classifier on the features of the canonical ResNet-50. In Figure \ref{zeroshot_vs_supervised} we show this comparison across 27 datasets. Please see Appendix \ref{sec:linear-probe} for details of datasets and setup. 

Zero-shot CLIP outperforms this baseline slightly more often than not and wins on 16 of the 27 datasets. Looking at individual datasets reveals some interesting behavior. On fine-grained classification tasks, we observe a wide spread in performance. On two of these datasets, Stanford Cars and Food101, zero-shot CLIP outperforms logistic regression on ResNet-50 features by over 20\% while on two others, Flowers102 and FGVCAircraft, zero-shot CLIP underperforms by over 10\%. On OxfordPets and Birdsnap, performance is much closer. We suspect these difference are primarily due to varying amounts of per-task supervision between WIT and ImageNet. On ``general'' object classification datasets such as ImageNet, CIFAR10/100, STL10, and PascalVOC2007 performance is relatively similar with a slight advantage for zero-shot CLIP in all cases. On STL10, CLIP achieves 99.3\% overall which appears to be a new state of the art despite not using any training examples. Zero-shot CLIP significantly outperforms a ResNet-50 on two datasets measuring action recognition in videos. On Kinetics700, CLIP outperforms a ResNet-50 by 14.5\%. Zero-shot CLIP also outperforms a ResNet-50's features by 7.7\% on UCF101. We speculate this is due to natural language providing wider supervision for visual concepts involving verbs, compared to the noun-centric object supervision in ImageNet. 

Looking at where zero-shot CLIP notably underperforms, we see that zero-shot CLIP is quite weak on several specialized, complex, or abstract tasks such as satellite image classification (EuroSAT and RESISC45), lymph node tumor detection (PatchCamelyon), counting objects in synthetic scenes (CLEVRCounts), self-driving related tasks such as German traffic sign recognition (GTSRB), recognizing distance to the nearest car (KITTI Distance). These results highlight the poor capability of zero-shot CLIP on more complex tasks. By contrast, non-expert humans can robustly perform several of these tasks, such as counting, satellite image classification, and traffic sign recognition, suggesting significant room for improvement. However, we caution that it is unclear whether measuring zero-shot transfer, as opposed to few-shot transfer, is a meaningful evaluation for difficult tasks that a learner has no prior experience with, such as lymph node tumor classification for almost all humans (and possibly CLIP).

\begin{figure}[t]
\begin{center}
\centerline{\includegraphics[width=1.0\columnwidth]{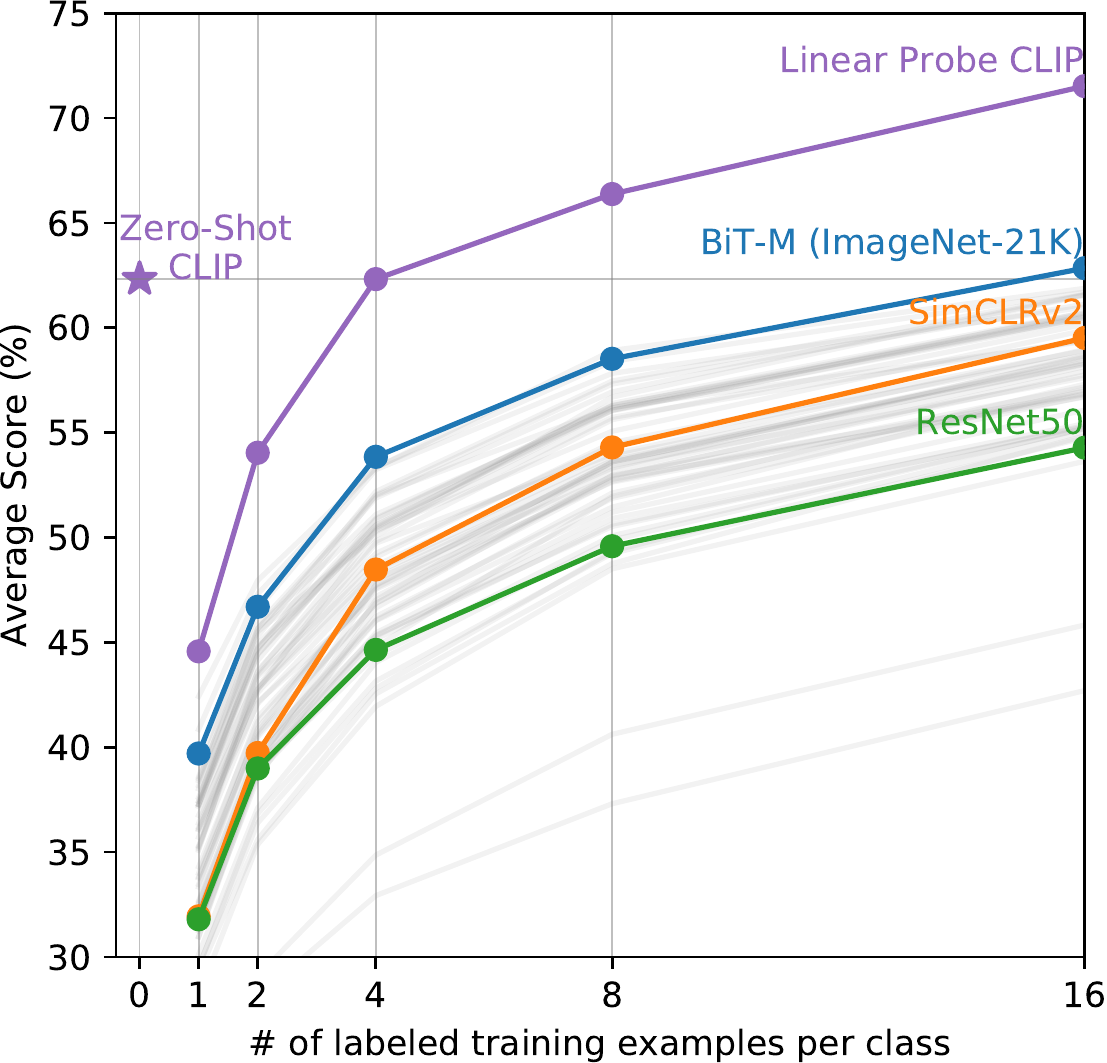}}
\caption{\textbf{Zero-shot CLIP outperforms few-shot linear probes.} Zero-shot CLIP matches the average performance of a 4-shot linear classifier trained on the same feature space and nearly matches the best results of a 16-shot linear classifier across publicly available models. For both BiT-M and SimCLRv2, the best performing model is highlighted. Light gray lines are other models in the eval suite. The 20 datasets with at least 16 examples per class were used in this analysis.}
\label{zeroshot_vs_fewshot}
\end{center}
\vspace{-2em}
\end{figure}

While comparing zero-shot performance to fully supervised models contextualizes the task-learning capabilities of CLIP, comparing to few-shot methods is a more direct comparison, since zero-shot is its limit. In Figure \ref{zeroshot_vs_fewshot}, we visualize how zero-shot CLIP compares to few-shot logistic regression on the features of many image models including the best publicly available ImageNet models, self-supervised learning methods, and CLIP itself. While it is intuitive to expect zero-shot to underperform one-shot, we instead find that zero-shot CLIP matches the performance of 4-shot logistic regression on the same feature space. This is likely due to an important difference between the zero-shot and few-shot approach. First, CLIP's zero-shot classifier is generated via natural language which allows for visual concepts to be directly specified (``communicated''). By contrast, ``normal'' supervised learning must infer concepts indirectly from training examples. Context-less example-based learning has the drawback that many different hypotheses can be consistent with the data, especially in the one-shot case. A single image often contains many different visual concepts. Although a capable learner is able to exploit visual cues and heuristics, such as assuming that the concept being demonstrated is the primary object in an image, there is no guarantee.

A potential resolution of this discrepancy between zero-shot and few-shot performance is to use CLIP's zero-shot classifier as a prior for the weights of the few-shot classifier. While adding an L2 penalty towards the generated weights is a straightforward implementation of this idea, we found that hyperparameter optimization would often select for such a large value of this regularizer that the resulting few-shot classifier was ``just'' the zero-shot classifier. Research into better methods of combining the strength of zero-shot transfer with flexibility of few-shot learning is a promising direction for future work.

When comparing zero-shot CLIP to few-shot logistic regression on the features of other models, zero-shot CLIP roughly matches the performance of the best performing 16-shot classifier in our evaluation suite, which uses the features of a BiT-M ResNet-152x2 trained on ImageNet-21K. We are certain that a BiT-L model trained on JFT-300M would perform even better but these models have not been publicly released. That a BiT-M ResNet-152x2 performs best in a 16-shot setting is somewhat surprising since, as analyzed in Section \ref{linear_probe_section}, the Noisy Student EfficientNet-L2 outperforms it in a fully supervised setting by almost 5\% on average across 27 datasets.

\begin{figure}[t]
\begin{center}
\centerline{\includegraphics[width=1.0\columnwidth]{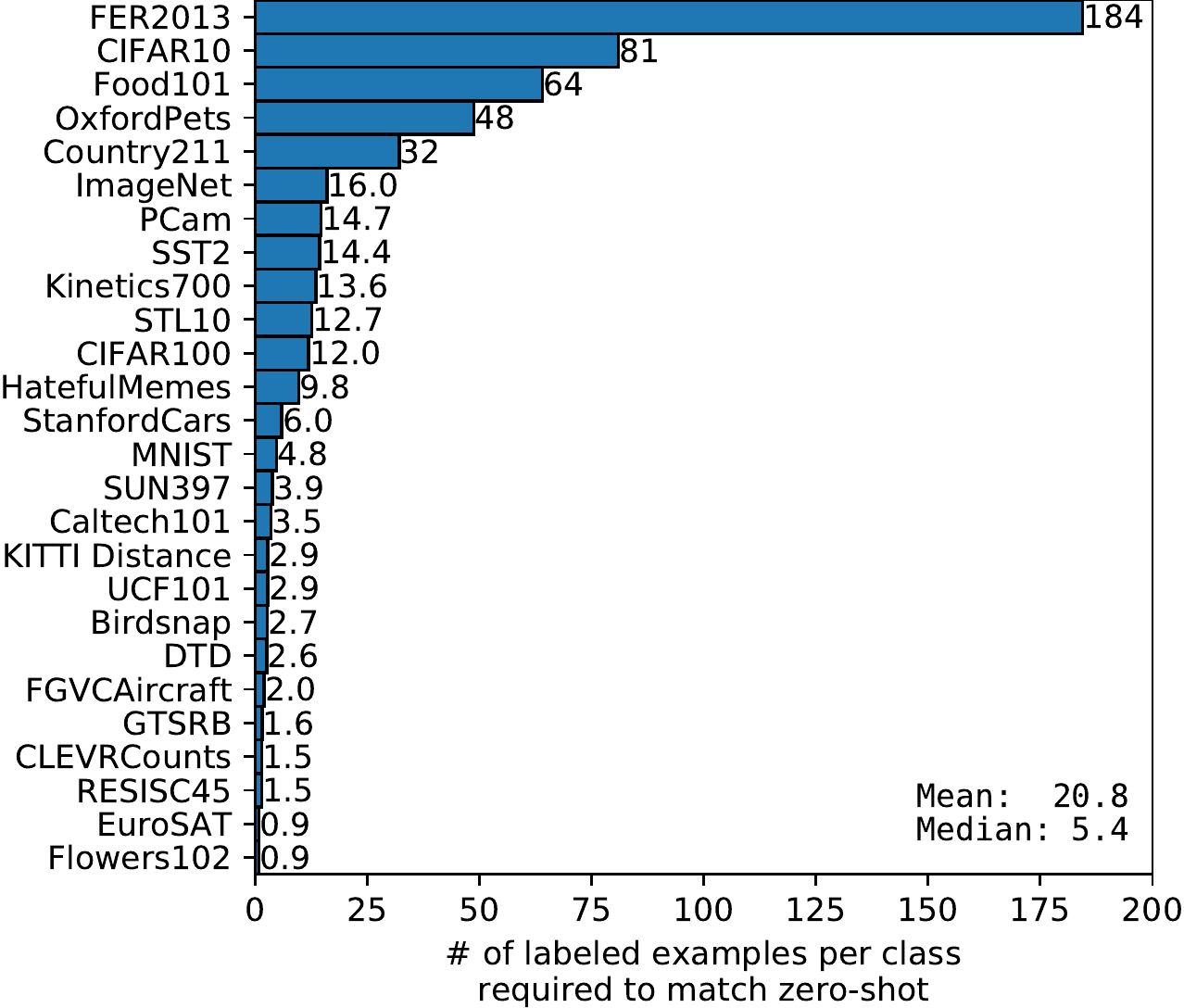}}
\caption{\textbf{The data efficiency of zero-shot transfer varies widely.} Calculating the number of labeled examples per class a linear classifier on the same CLIP feature space requires to match the performance of the zero-shot classifier contextualizes the effectiveness of zero-shot transfer. Values are estimated based on log-linear interpolation of 1, 2, 4, 8, 16-shot and fully supervised results. Performance varies widely from still underperforming a one-shot classifier on two datasets to matching an estimated 184 labeled examples per class.}
\label{zeroshot_data_efficiency}
\end{center}
\vspace{-2em}
\end{figure}

In addition to studying the average performance of zero-shot CLIP and few-shot logistic regression, we also examine performance on individual datasets. In Figure \ref{zeroshot_data_efficiency}, we show estimates for the number of labeled examples per class that a logistic regression classifier on the same feature space requires to match the performance of zero-shot CLIP. Since zero-shot CLIP is also a linear classifier, this estimates the effective data efficiency of zero-shot transfer in this setting. In order to avoid training thousands of linear classifiers, we estimate the effective data efficiency based on a log-linear interpolation of the performance of a 1, 2, 4, 8, 16-shot (when possible), and a fully supervised linear classifier trained on each dataset. We find that zero-shot transfer can have widely varying efficiency per dataset from less than 1 labeled example per class to 184. Two datasets, Flowers102 and EuroSAT underperform one-shot models. Half of the datasets require less than 5 examples per class with a median of 5.4. However, the mean estimated data efficiency is 20.8 examples per class. This is due to the 20\% of datasets where supervised classifiers require many labeled examples per class in order to match performance. On ImageNet, zero-shot CLIP matches the performance of a 16-shot linear classifier trained on the same feature space.

\begin{figure}[t]
\begin{center}
\centerline{\includegraphics[width=1.0\columnwidth]{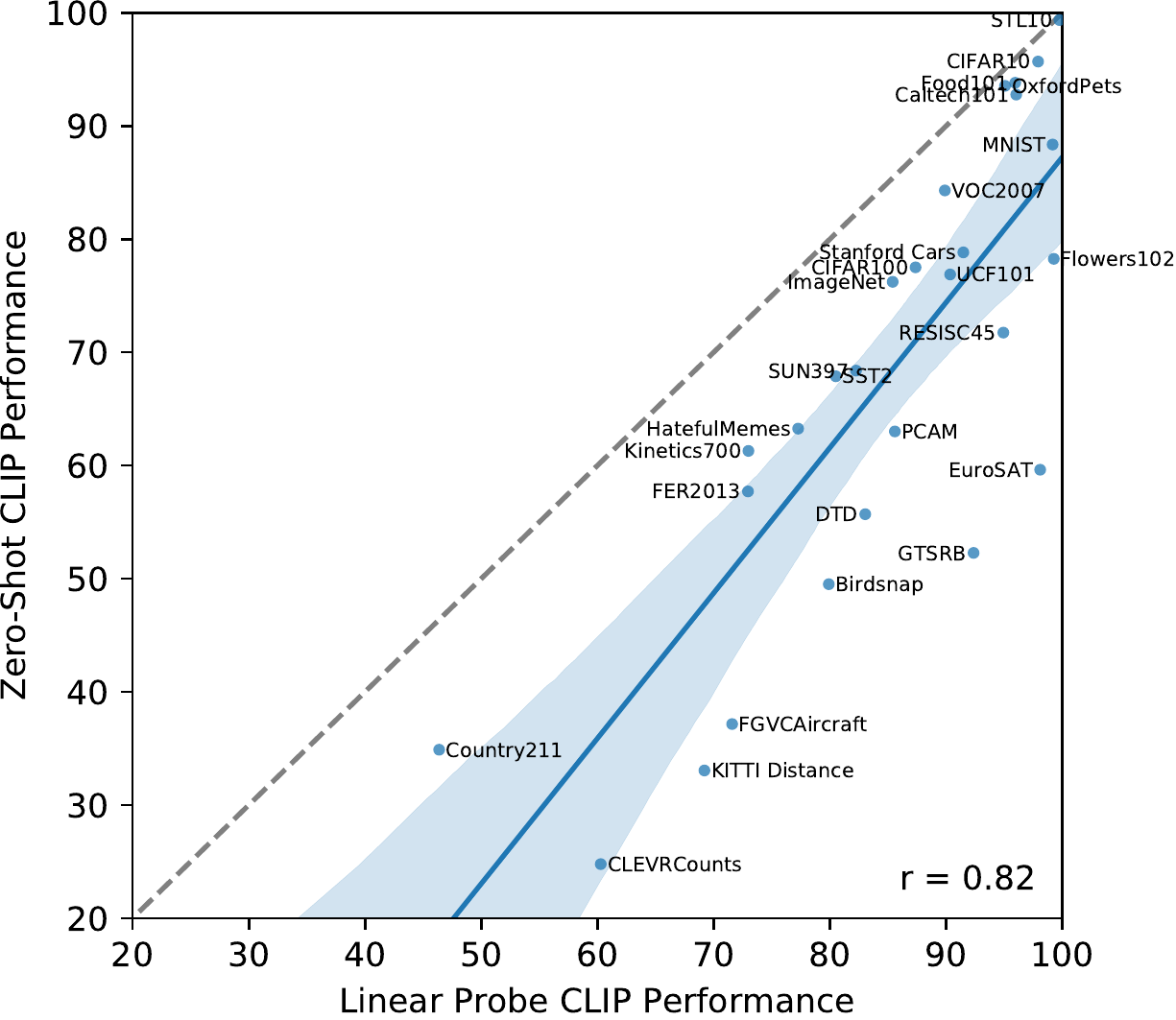}}
\caption{\textbf{Zero-shot performance is correlated with linear probe performance but still mostly sub-optimal.} Comparing zero-shot and linear probe performance across datasets shows a strong correlation with zero-shot performance mostly shifted 10 to 25 points lower. On only 5 datasets does zero-shot performance approach linear probe performance ($\le$3 point difference).}
\label{zeroshot_vs_linear_probe}
\end{center}
\vspace{-2em}
\end{figure}

If we assume that evaluation datasets are large enough that the parameters of linear classifiers trained on them are well estimated, then, because CLIP's zero-shot classifier is also a linear classifier, the performance of the fully supervised classifiers roughly sets an upper bound for what zero-shot transfer can achieve. In Figure \ref{zeroshot_vs_linear_probe} we compare CLIP's zero-shot performance with fully supervised linear classifiers across datasets. The dashed, $y=x$ line represents an ``optimal'' zero-shot classifier that matches the performance of its fully supervised equivalent. For most datasets, the performance of zero-shot classifiers still underperform fully supervised classifiers by 10\% to 25\%, suggesting that there is still plenty of headroom for improving CLIP's task-learning and zero-shot transfer capabilities.

There is a positive correlation of 0.82 (p-value $<10^{-6}$) between zero-shot performance and fully supervised performance, suggesting that CLIP is relatively consistent at connecting underlying representation and task learning to zero-shot transfer. However, zero-shot CLIP only approaches fully supervised performance on 5 datasets: STL10, CIFAR10, Food101, OxfordPets, and Caltech101. On all 5 datasets, both zero-shot accuracy and fully supervised accuracy are over 90\%. This suggests that CLIP may be more effective at zero-shot transfer for tasks where its underlying representations are also high quality. The slope of a linear regression model predicting zero-shot performance as a function of fully supervised performance estimates that for every 1\% improvement in fully supervised performance, zero-shot performance improves by 1.28\%. However, the 95th-percentile confidence intervals still include values of less than 1 (0.93-1.79).

\begin{figure}[t]
\begin{center}
\centerline{\includegraphics[width=1.0\columnwidth]{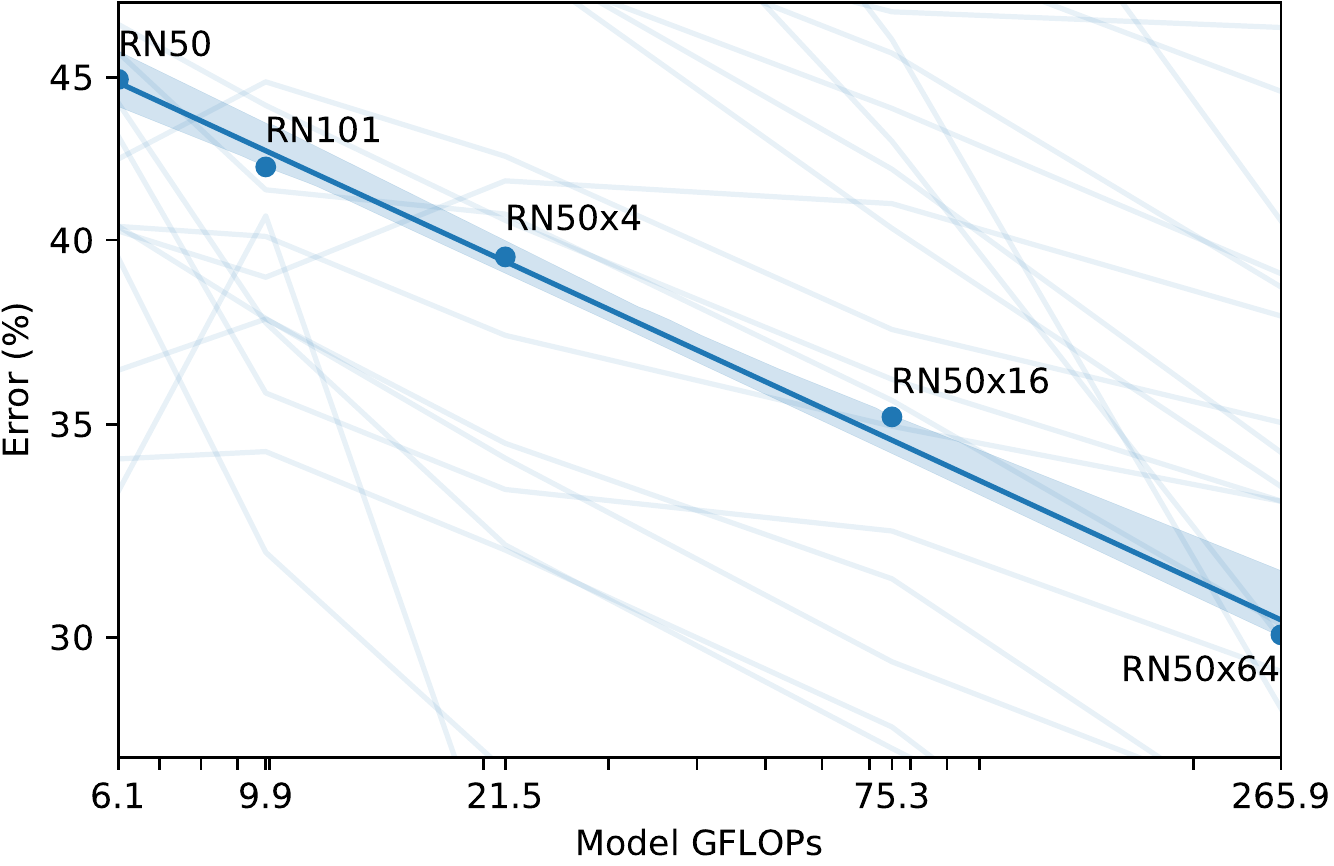}}
\caption{\textbf{Zero-shot CLIP performance scales smoothly as a function of model compute.} Across 39 evals on 36 different datasets, average zero-shot error is well modeled by a log-log linear trend across a 44x range of compute spanning 5 different CLIP models. Lightly shaded lines are performance on individual evals, showing that performance is much more varied despite the smooth overall trend.}
\label{zeroshot_scaling}
\end{center}
\vspace{-2em}
\end{figure}

\begin{figure*}[t]
\includegraphics[width=\textwidth]{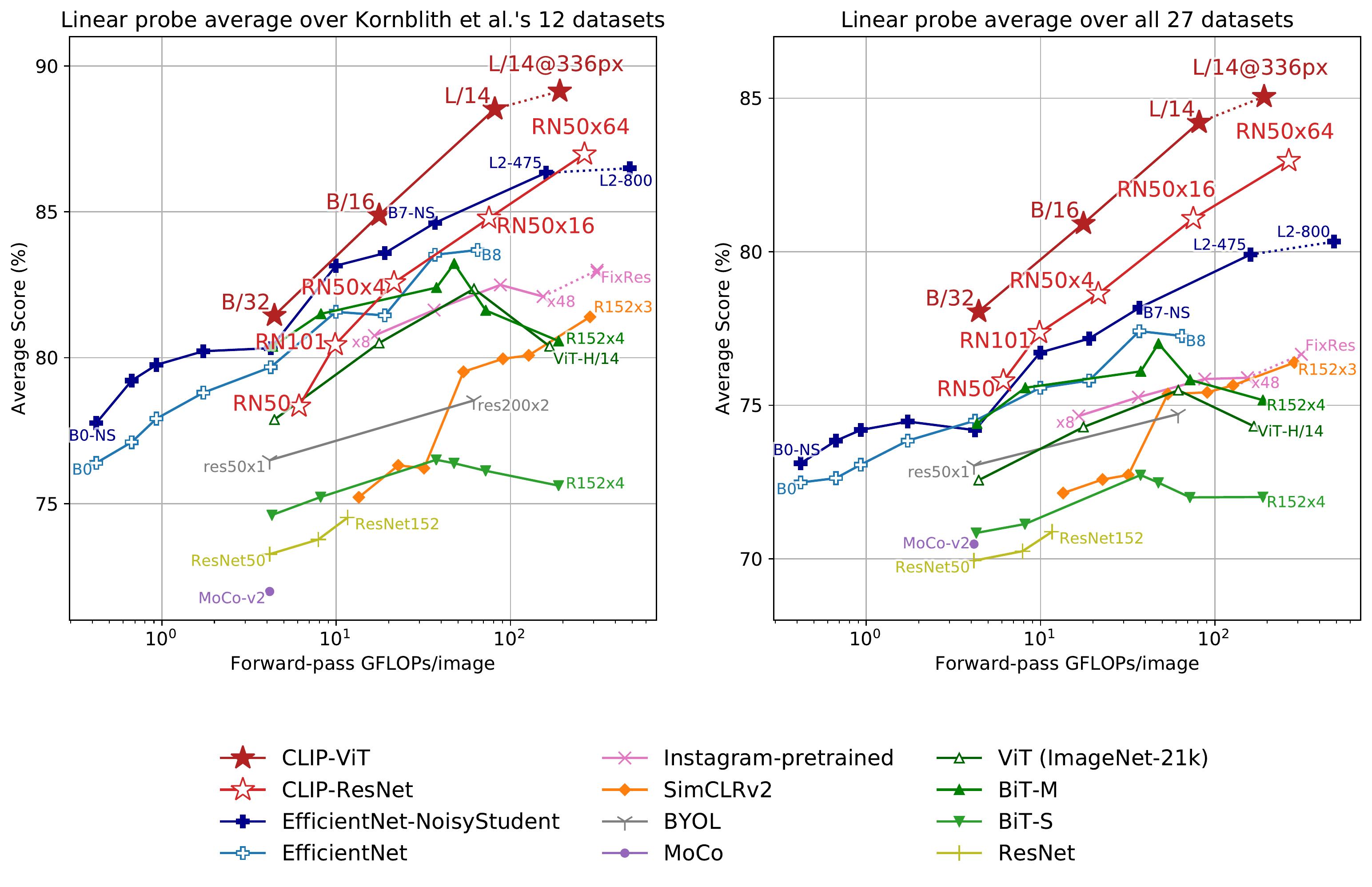}
\vspace{-2em}
\caption{\textbf{Linear probe performance of CLIP models in comparison with state-of-the-art computer vision models}, including EfficientNet \cite{tan2019efficientnet,xie2020self}, MoCo \cite{chen2020mocov2}, Instagram-pretrained ResNeXt models \cite{mahajan2018exploring,touvron2019fixing}, BiT \cite{kolesnikov2019large}, ViT \cite{dosovitskiy2020image}, SimCLRv2 \cite{chen2020big}, BYOL \cite{grill2020byol}, and the original ResNet models \cite{he2016resnet}. (Left) Scores are averaged over 12 datasets studied by \citet{kornblith2019better}. (Right) Scores are averaged over 27 datasets that contain a wider variety of distributions. Dotted lines indicate models fine-tuned or evaluated on images at a higher-resolution than pre-training. See Table \ref{tab:linear-probe-big-table} for individual scores and Figure \ref{linear-probe-per-dataset} for plots for each dataset.}
\label{fig:linear-probe-graph}
\end{figure*}

Over the past few years, empirical studies of deep learning systems have documented that performance is predictable as a function of important quantities such as training compute and dataset size \cite{hestness2017deep,kaplan2020scaling}. The GPT family of models has so far demonstrated consistent improvements in zero-shot performance across a 1000x increase in training compute. In Figure \ref{zeroshot_scaling}, we check whether the zero-shot performance of CLIP follows a similar scaling pattern. We plot the average error rate of the 5 ResNet CLIP models across 39 evaluations on 36 different datasets and find that a similar log-log linear scaling trend holds for CLIP across a 44x increase in model compute. While the overall trend is smooth, we found that performance on individual evaluations can be much noisier. We are unsure whether this is caused by high variance between individual training runs on sub-tasks (as documented in \citet{d2020underspecification}) masking a steadily improving trend or whether performance is actually non-monotonic as a function of compute on some tasks.

\subsection{Representation Learning}
\label{linear_probe_section}

While we have extensively analyzed the task-learning capabilities of CLIP through zero-shot transfer in the previous section, it is more common to study the representation learning capabilities of a model. There exist many ways to evaluate the quality of representations as well as disagreements over what properties an ``ideal'' representation should have \citep{locatello2020sober}. Fitting a linear classifier on a representation extracted from the model and measuring its performance on various datasets is a common approach. An alternative is measuring the performance of end-to-end fine-tuning of the model. This increases flexibility, and prior work has convincingly demonstrated that fine-tuning outperforms linear classification on most image classification datasets \citep{kornblith2019better,zhai2019large}. While the high performance of fine-tuning motivates its study for practical reasons, we still opt for linear classifier based evaluation for several reasons. Our work is focused on developing a high-performing task and dataset-agnostic pre-training approach. Fine-tuning, because it adapts representations to each dataset during the fine-tuning phase, can compensate for and potentially mask failures to learn general and robust representations during the pre-training phase. Linear classifiers, because of their limited flexibility, instead highlight these failures and provide clear feedback during development. For CLIP, training supervised linear classifiers has the added benefit of being very similar to the approach used for its zero-shot classifiers which enables extensive comparisons and analysis in Section \ref{subsection:zero_shot_transfer}. Finally, we aim to compare CLIP to a comprehensive set of existing models across many tasks. Studying 66 different models on 27 different datasets requires tuning 1782 different evaluations. Fine-tuning opens up a much larger design and hyper-parameter space, which makes it difficult to fairly evaluate and computationally expensive to compare a diverse set of techniques as discussed in other large scale empirical studies \citep{lucic2018gans,choi2019empirical}. By comparison, linear classifiers require minimal hyper-parameter tuning and have standardized implementations and evaluation procedures. Please see Appendix \ref{sec:linear-probe} for further details on evaluation.

Figure \ref{fig:linear-probe-graph} summarizes our findings. To minimize selection effects that could raise concerns of confirmation or reporting bias, we first study performance on the 12 dataset evaluation suite from \citet{kornblith2019better}. While small CLIP models such as a ResNet-50 and ResNet-101 outperform other ResNets trained on ImageNet-1K (BiT-S and the originals), they underperform ResNets trained on ImageNet-21K (BiT-M). These small CLIP models also underperform models in the EfficientNet family with similar compute requirements. However, models trained with CLIP scale very well and the largest model we trained (ResNet-50x64) slightly outperforms the best performing existing model (a Noisy Student EfficientNet-L2) on both overall score and compute efficiency. We also find that CLIP vision transformers are about 3x more compute efficient than CLIP ResNets, which allows us to reach higher overall performance within our compute budget. These results qualitatively replicate the findings of \citet{dosovitskiy2020image} which reported that vision transformers are more compute efficient than convnets when trained on sufficiently large datasets. Our best overall model is a ViT-L/14 that is fine-tuned at a higher resolution of 336 pixels on our dataset for 1 additional epoch. This model outperforms the best existing model across this evaluation suite by an average of 2.6\%.

As Figure \ref{zero_shot_prediction_fig} qualitatively shows, CLIP models learn a wider set of tasks than has previously been demonstrated in a single computer vision model trained end-to-end from random initialization. These tasks include geo-localization, optical character recognition, facial emotion recognition, and action recognition. None of these tasks are measured in the evaluation suite of \citet{kornblith2019better}. This could be argued to be a form of selection bias in \citet{kornblith2019better}'s study towards tasks that overlap with ImageNet. To address this, we also measure performance on a broader 27 dataset evaluation suite. This evaluation suite, detailed in Appendix \ref{sec:linear-probe} includes datasets representing the aforementioned tasks, German Traffic Signs Recognition Benchmark \citep{GTSRB2011}, as well as several other datasets adapted from VTAB \citep{zhai2019large}.

\begin{figure}[t]
\begin{center}
\centerline{\includegraphics[width=\columnwidth]{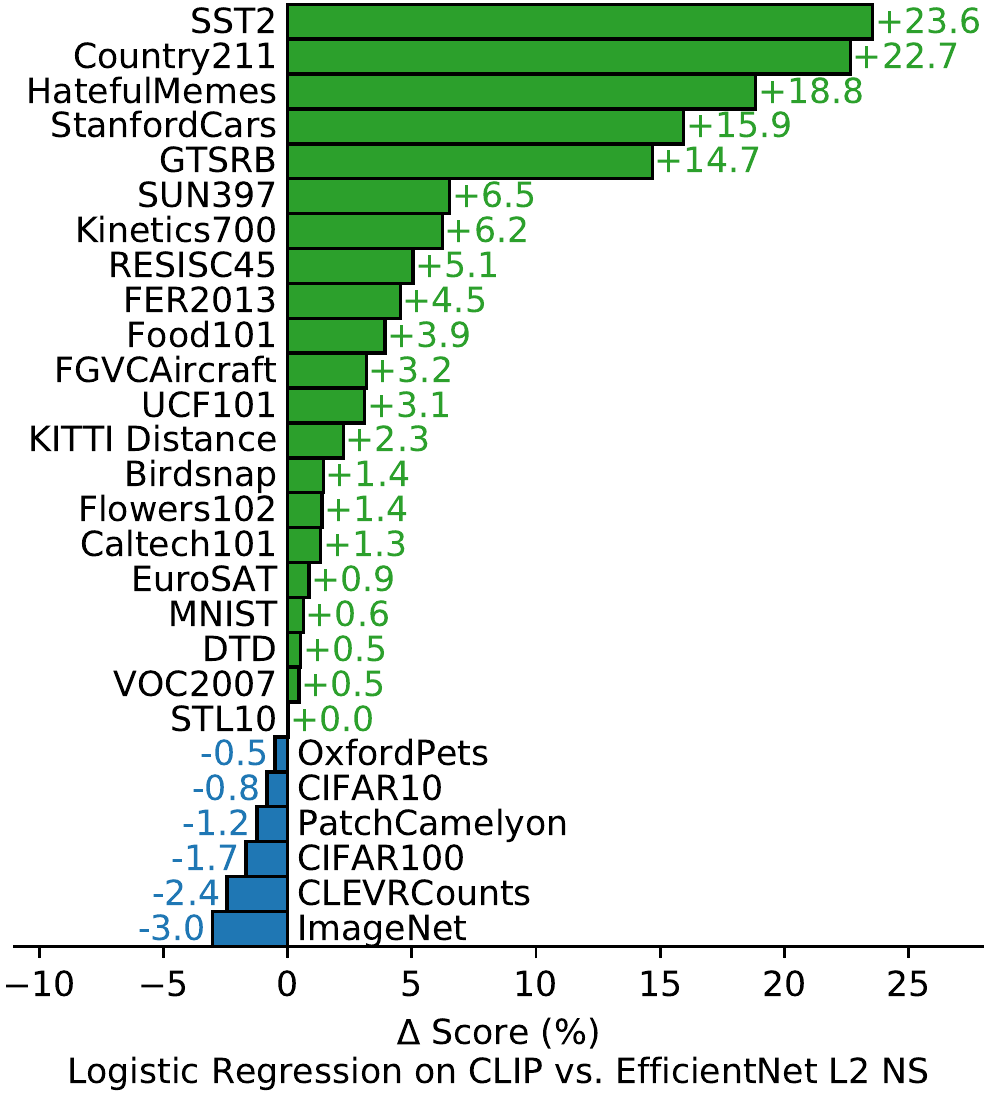}}
\caption{\textbf{CLIP's features outperform the features of the best ImageNet model on a wide variety of datasets.} Fitting a linear classifier on CLIP's features outperforms using the Noisy Student EfficientNet-L2 on 21 out of 27 datasets.}
\label{linear-probe-clip-vs-enet}
\end{center}
\vspace{-1cm}
\end{figure}
\begin{figure*}[t]
\includegraphics[width=\textwidth]{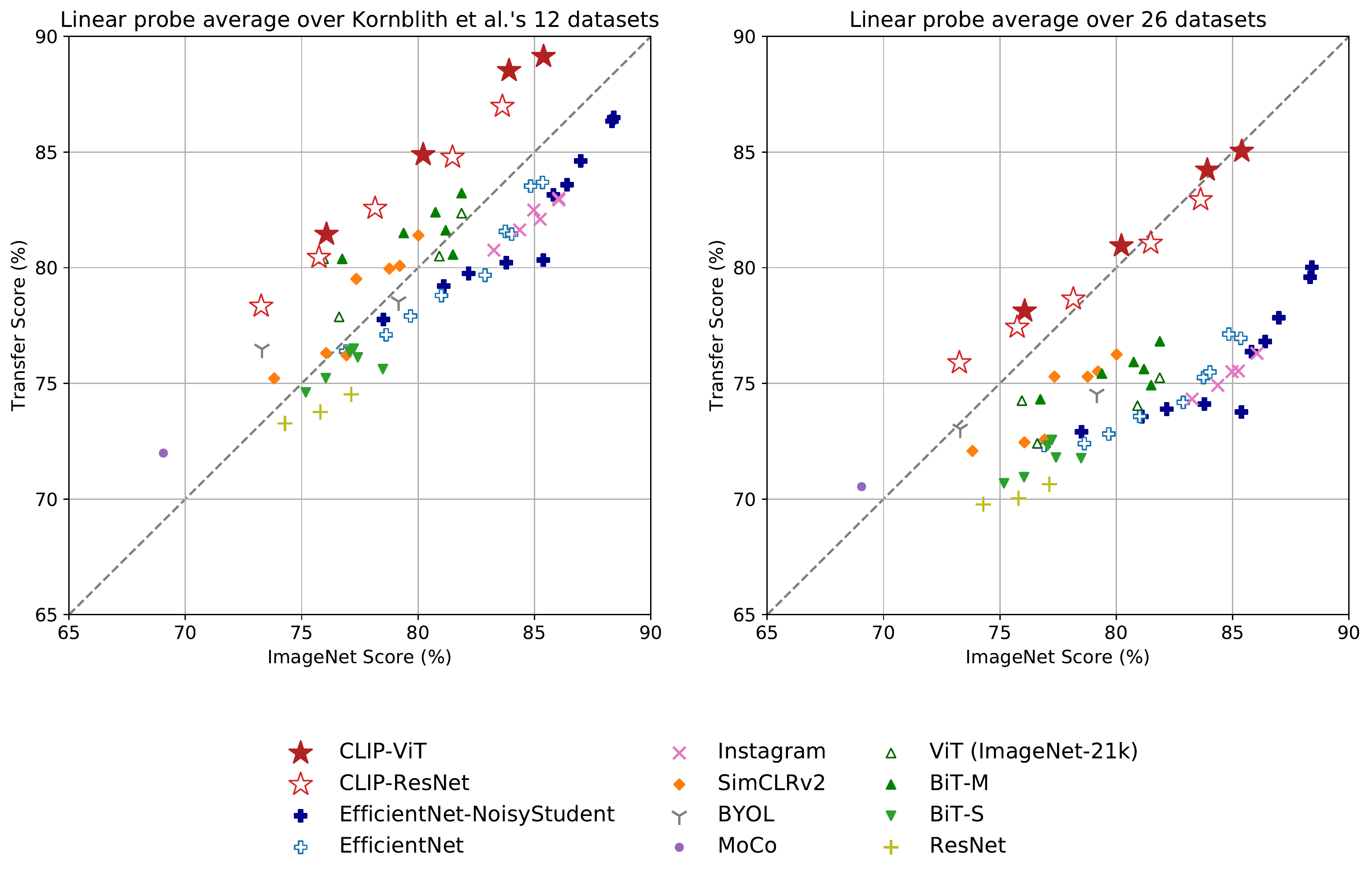}
\vspace{-2em}
\caption{\textbf{CLIP's features are more robust to task shift when compared to models pre-trained on ImageNet.}
For both dataset splits, the transfer scores of linear probes trained on the representations of CLIP models are higher than other models with similar ImageNet performance. This suggests that the representations of models trained on ImageNet are somewhat overfit to their task. %
}
\label{fig:linear-probe-transfer}
\end{figure*}

On this broader evaluation suite, the benefits of CLIP are more clear. All CLIP models, regardless of scale, outperform all evaluated systems in terms of compute efficiency. The improvement in average score of the best model over previous systems increases from 2.6\% to 5\%. We also find that self-supervised systems do noticeably better on our broader evaluation suite. For instance, while SimCLRv2 still underperforms BiT-M on average on the 12 datasets of \citet{kornblith2019better}, SimCLRv2 outperforms BiT-M on our 27 dataset evaluation suite. These findings suggest continuing to expand task diversity and coverage in order to better understand the ``general'' performance of systems. We suspect additional evaluation efforts along the lines of VTAB to be valuable.

In addition to the aggregate analysis above, we visualize per-dataset differences in the performance of the best CLIP model and the best model in our evaluation suite across all 27 datasets in Figure \ref{linear-probe-clip-vs-enet}. CLIP outperforms the Noisy Student EfficientNet-L2 on 21 of the 27 datasets. CLIP improves the most on tasks which require OCR (SST2 and HatefulMemes), geo-localization and scene recognition (Country211, SUN397), and activity recognition in videos (Kinetics700 and UCF101). In addition CLIP also does much better on fine-grained car and traffic sign recognition (Stanford Cars and GTSRB). This may reflect a problem with overly narrow supervision in ImageNet. A result such as the 14.7\% improvement on GTSRB could be indicative of an issue with ImageNet-1K, which has only a single label for all traffic and street signs. This could encourage a supervised representation to collapse intra-class details and hurt accuracy on a fine-grained downstream task. As mentioned, CLIP still underperforms the EfficientNet on several datasets. Unsurprisingly, the dataset that the EfficientNet does best relative to CLIP on is the one it was trained on: ImageNet. The EffcientNet also slightly outperforms CLIP on low-resolution datasets such as CIFAR10 and CIFAR100. We suspect this is at least partly due to the lack of scale-based data augmentation in CLIP. The EfficientNet also does slightly better on PatchCamelyon and CLEVRCounts, datasets where overall performance is still low for both approaches.

\subsection{Robustness to Natural Distribution Shift}
\label{subsection:robustness}

\begin{figure*}[t]
\begin{center}
\includegraphics[width=0.71\columnwidth]{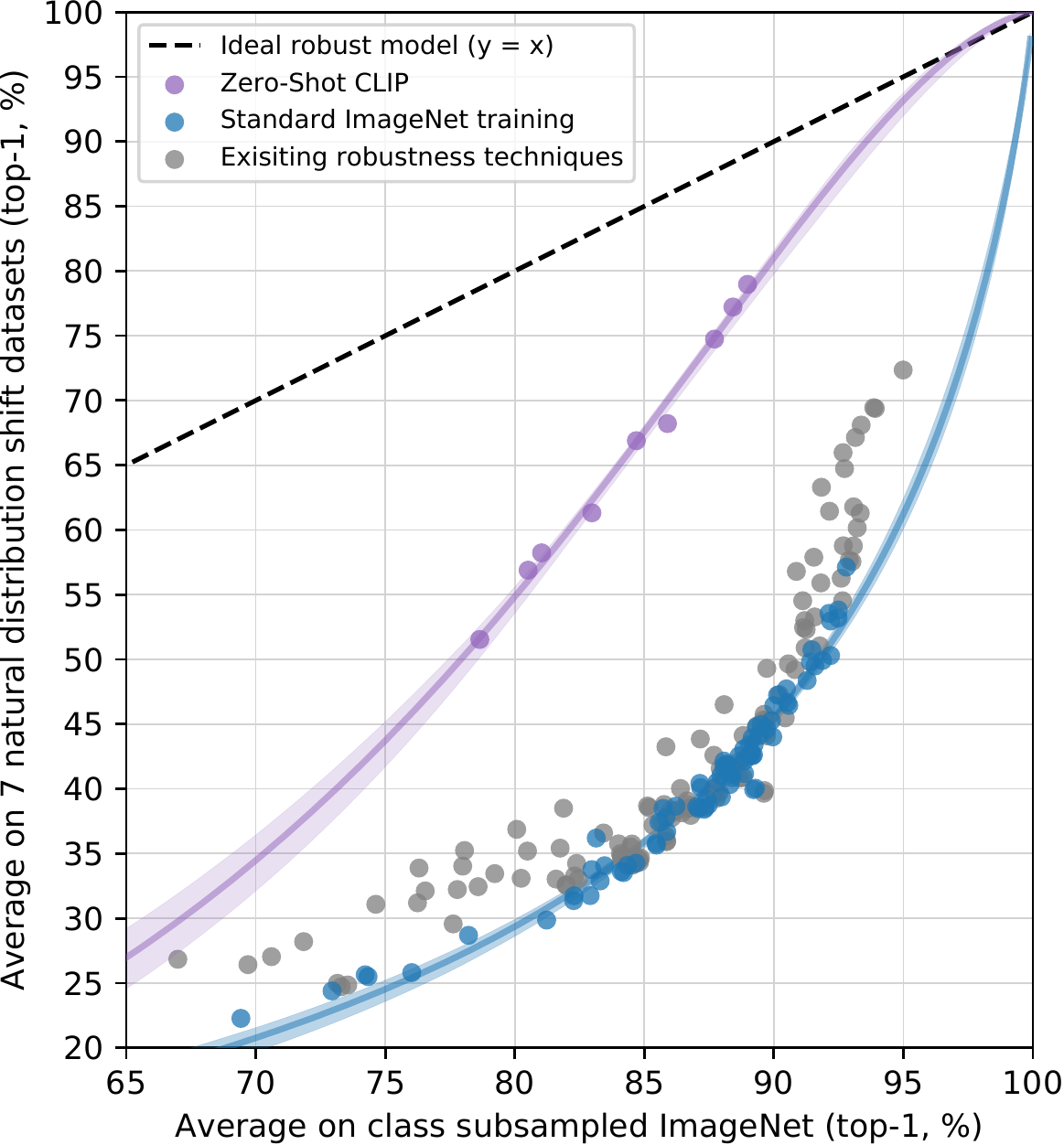}
\includegraphics[width=1.29\columnwidth]{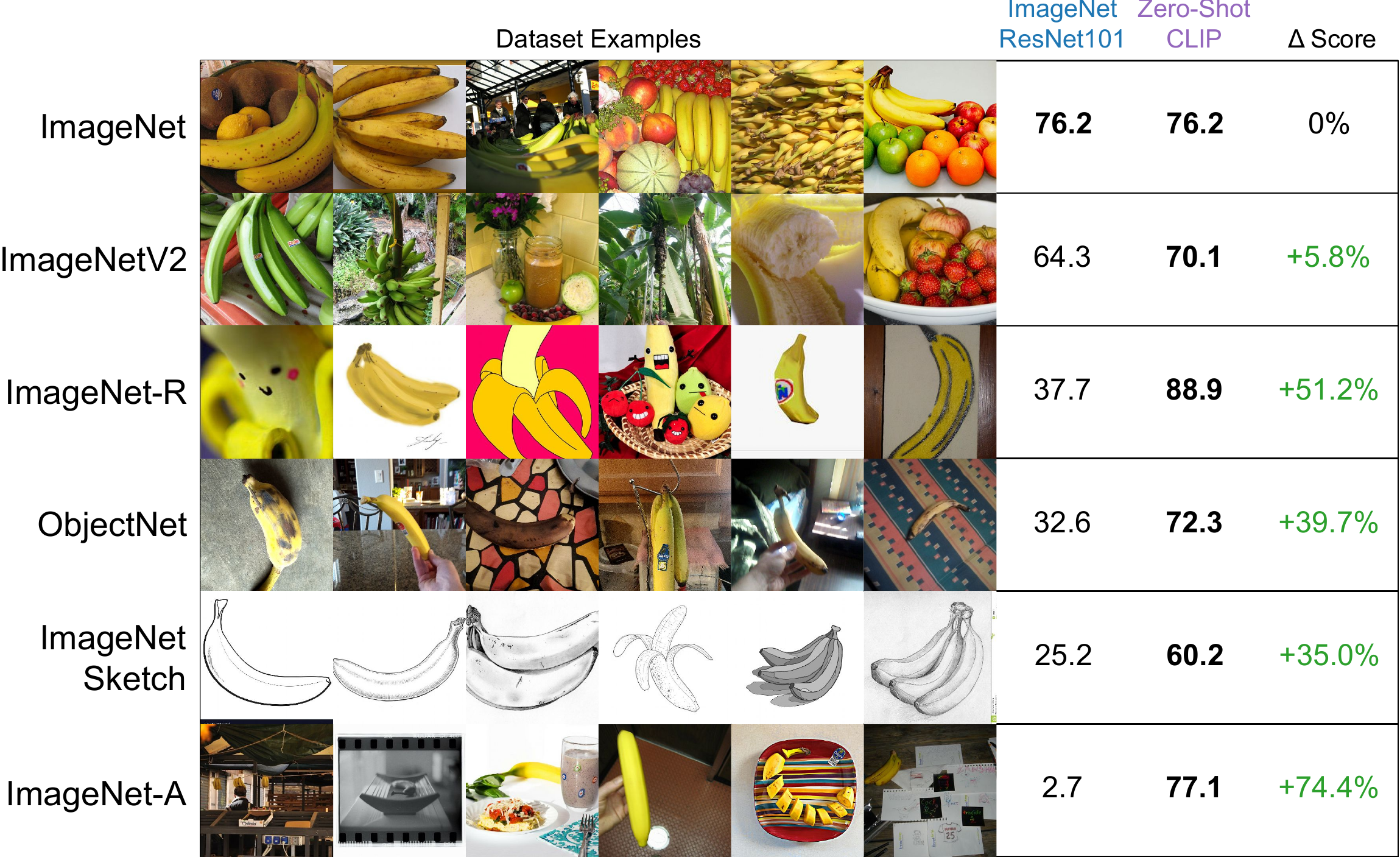}
\caption{\textbf{Zero-shot CLIP is much more robust to distribution shift than standard ImageNet models.} (Left) An ideal robust model (dashed line) performs equally well on the ImageNet distribution and on other natural image distributions. Zero-shot CLIP models shrink this ``robustness gap'' by up to 75\%. Linear fits on logit transformed values are shown with bootstrap estimated 95\% confidence intervals. (Right) Visualizing distribution shift for bananas, a class shared across 5 of the 7 natural distribution shift datasets. The performance of the best zero-shot CLIP model, ViT-L/14@336px, is compared with a model that has the same performance on the ImageNet validation set, ResNet-101.}
\label{robust_main_fig}
\end{center}
\end{figure*}

In 2015, it was announced that a deep learning model exceeded human performance on the ImageNet test set \citep{he2015delving}. However, research in the subsequent years has repeatedly found that these models still make many simple mistakes \citep{dodge2017study,geirhos2018imagenet,alcorn2019strike}, and new benchmarks testing these systems has often found their performance to be much lower than both their ImageNet accuracy and human accuracy \citep{recht2019imagenet,barbu2019objectnet}. What explains this discrepancy? Various ideas have been suggested and studied \citep{ilyas2019adversarial,geirhos2020shortcut}. A common theme of proposed explanations is that deep learning models are exceedingly adept at finding correlations and patterns which hold across their training dataset and thus improve in-distribution performance. However many of these correlations and patterns are actually spurious and do not hold for other distributions and result in large drops in performance on other datasets.

We caution that, to date, most of these studies limit their evaluation to models trained on ImageNet. Recalling the topic of discussion, it may be a mistake to generalize too far from these initial findings. To what degree are these failures attributable to deep learning, ImageNet, or some combination of the two? CLIP models, which are trained via natural language supervision on a very large dataset and are capable of high zero-shot performance, are an opportunity to investigate this question from a different angle.

\citet{taori2020measuring} is a recent comprehensive study moving towards quantifying and understanding these behaviors for ImageNet models. \citet{taori2020measuring} study how the performance of ImageNet models change when evaluated on \textit{natural distribution shifts}. They measure performance on a set of 7 distribution shifts: ImageNetV2 \citep{recht2019imagenet}, ImageNet Sketch \citep{wang2019learning}, Youtube-BB and ImageNet-Vid \citep{shankar2019image}, ObjectNet \citep{barbu2019objectnet}, ImageNet Adversarial \citep{hendrycks2019natural}, and ImageNet Rendition \citep{hendrycks2020many}. They distinguish these datasets, which all consist of novel images collected from a variety of sources, from \textit{synthetic distribution shifts} such as ImageNet-C \citep{hendrycks2019benchmarking}, Stylized ImageNet \citep{geirhos2018imagenet}, or adversarial attacks \citep{goodfellow2014explaining} which are created by perturbing existing images in various ways. They propose this distinction because in part because they find that while several techniques have been demonstrated to improve performance on synthetic distribution shifts, they often fail to yield consistent improvements on natural distributions.\footnote{We refer readers to \citet{hendrycks2020many} for additional experiments and discussion on this claim.}

\begin{figure*}[t]
\begin{center}
\includegraphics[width=1.02\columnwidth]{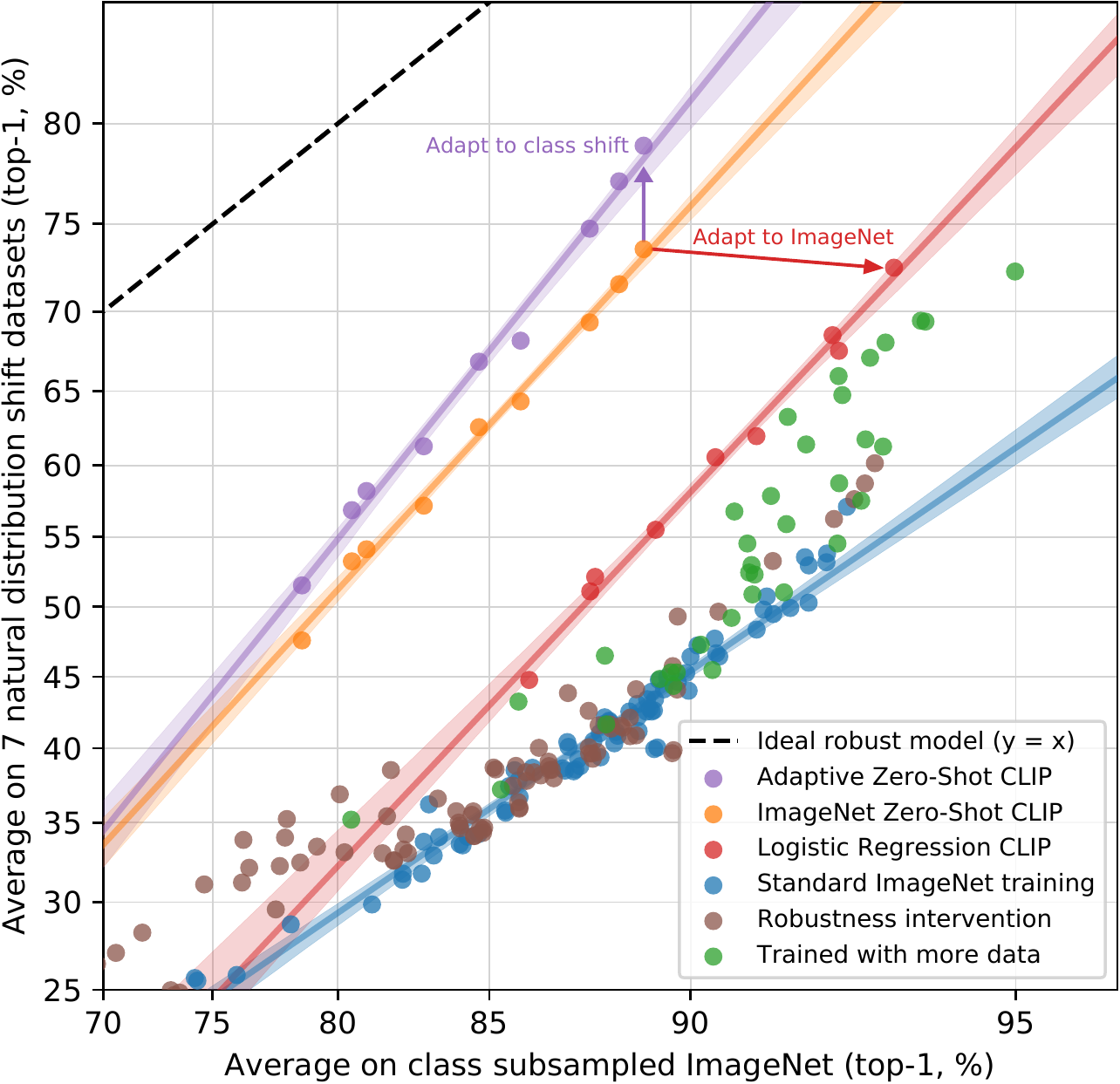}
\includegraphics[width=0.98\columnwidth]{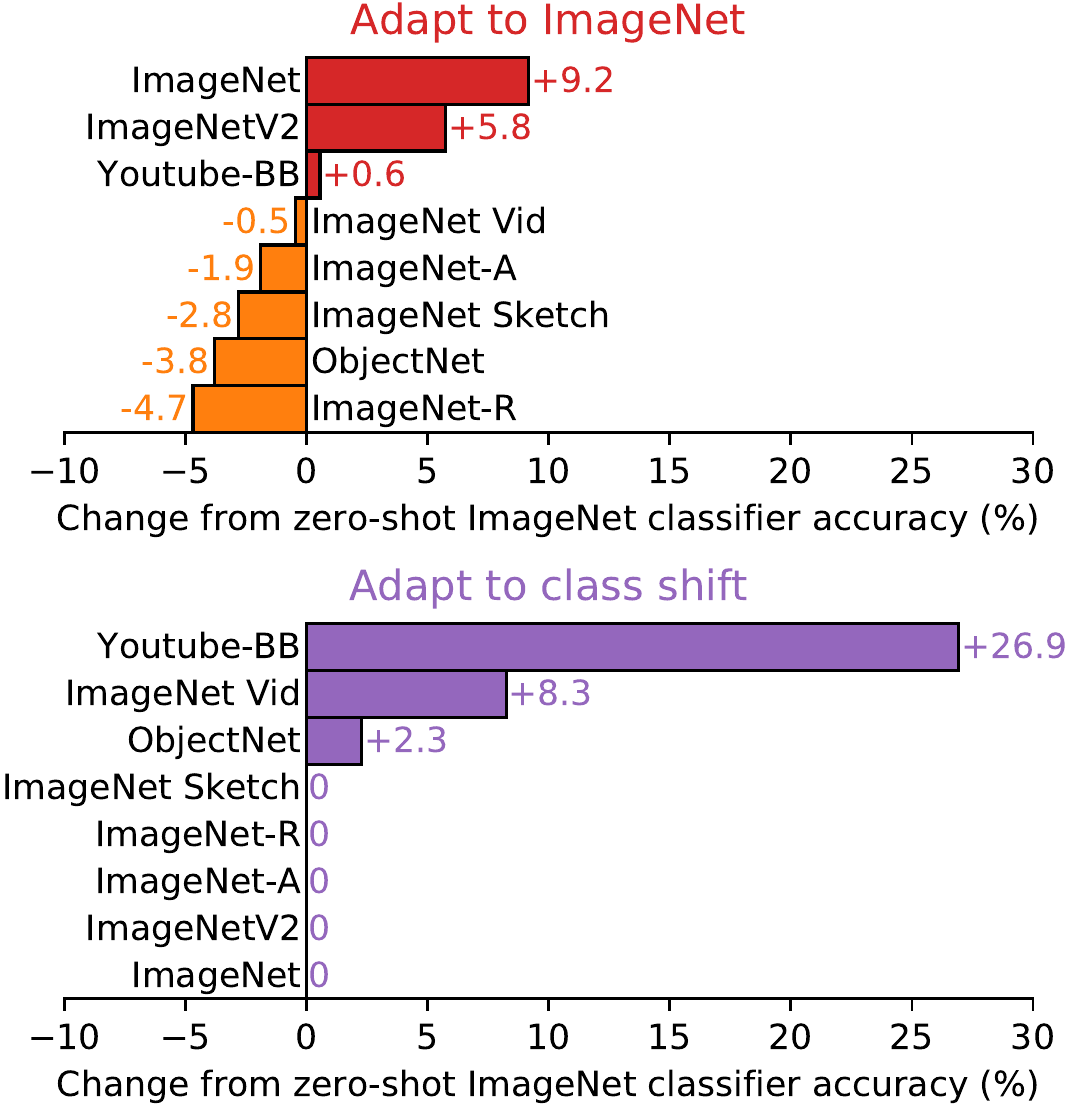}
\caption{\textbf{While supervised adaptation to ImageNet increases ImageNet accuracy by 9.2\%, it slightly reduces average robustness.} (Left) Customizing zero-shot CLIP to each dataset improves robustness compared to using a single static zero-shot ImageNet classifier and pooling predictions across similar classes as in \citet{taori2020measuring}. CLIP models adapted to ImageNet have similar effective robustness as the best prior ImageNet models. (Right) Details of per dataset changes in accuracy for the two robustness interventions. Adapting to ImageNet increases accuracy on ImageNetV2 noticeably but trades off accuracy on several other distributions. Dataset specific zero-shot classifiers can improve accuracy by a large amount but are limited to only a few datasets that include classes which don't perfectly align with ImageNet categories.}
\label{robustness_interventions}
\end{center}
\end{figure*}

Across these collected datasets, the accuracy of ImageNet models drop well below the expectation set by the ImageNet validation set. For the following summary discussion we report average accuracy across all 7 natural distribution shift datasets and average accuracy across the corresponding class subsets of ImageNet unless otherwise specified. Additionally, for Youtube-BB and ImageNet-Vid, which have two different evaluation settings, we use the average of pm-0 and pm-10 accuracy. 

A ResNet-101 makes 5 times as many mistakes when evaluated on these natural distribution shifts compared to the ImageNet validation set. Encouragingly however, \citet{taori2020measuring} find that accuracy under distribution shift increases predictably with ImageNet accuracy and is well modeled as a linear function of logit-transformed accuracy. \citet{taori2020measuring} use this finding to propose that robustness analysis should distinguish between \textit{effective} and \textit{relative} robustness. Effective robustness measures improvements in accuracy under distribution shift above what is predicted by the documented relationship between in-distribution and out-of-distribution accuracy. Relative robustness captures any improvement in out-of-distribution accuracy. \citet{taori2020measuring} argue that robustness techniques should aim to improve both effective robustness and relative robustness.

Almost all models studied in \citet{taori2020measuring} are trained or fine-tuned on the ImageNet dataset. Returning to the discussion in the introduction to this section - is training or adapting to the ImageNet dataset distribution the cause of the observed robustness gap? Intuitively, a zero-shot model should not be able to exploit spurious correlations or patterns that hold only on a specific distribution, since it is not trained on that distribution. \footnote{We caution that a zero-shot model can still exploit spurious correlations that are shared between the pre-training and evaluation distributions.} Thus it is reasonable to expect zero-shot models to have much higher effective robustness. In Figure \ref{robust_main_fig}, we compare the performance of zero-shot CLIP with existing ImageNet models on natural distribution shifts. All zero-shot CLIP models improve effective robustness by a large amount and reduce the size of the gap between ImageNet accuracy and accuracy under distribution shift by up to 75\%.

While these results show that zero-shot models can be much more robust, they do not necessarily mean that supervised learning on ImageNet causes a robustness gap. Other details of CLIP, such as its large and diverse pre-training dataset or use of natural language supervision could also result in much more robust models regardless of whether they are zero-shot or fine-tuned. As an initial experiment to potentially begin narrowing this down, we also measure how the performance of CLIP models change after adapting to the ImageNet distribution via a L2 regularized logistic regression classifier fit to CLIP features on the ImageNet training set. We visualize how performance changes from the zero-shot classifier in Figure \ref{robustness_interventions}. Although adapting CLIP to the ImageNet distribution increases its ImageNet accuracy by 9.2\% to 85.4\% overall, and ties the accuracy of the 2018 SOTA from \citet{mahajan2018exploring}, \textit{average accuracy under distribution shift slightly decreases}.

It is surprising to see a 9.2\% increase in accuracy, which corresponds to roughly 3 years of improvement in SOTA, fail to translate into any improvement in average performance under distribution shift. We also break down the differences between zero-shot accuracy and linear classifier accuracy per dataset in Figure \ref{robustness_interventions} and find performance still increases significantly on one dataset, ImageNetV2. ImageNetV2 closely followed the creation process of the original ImageNet dataset which suggests that gains in accuracy from supervised adaptation are closely concentrated around the ImageNet distribution. Performance decreases by 4.7\% on ImageNet-R, 3.8\% on ObjectNet, 2.8\% on ImageNet Sketch, and 1.9\% on ImageNet-A. The change in accuracy on the two other datasets, Youtube-BB and ImageNet Vid, is insignificant.

How is it possible to improve accuracy by 9.2\% on the ImageNet dataset with little to no increase in accuracy under distribution shift? Is the gain primarily from ``exploiting spurious correlations''? Is this behavior unique to some combination of CLIP, the ImageNet datatset, and the distribution shifts studied, or a more general phenomena? Does it hold for end-to-end finetuning as well as linear classifiers? We do not have confident answers to these questions at this time. Prior work has also pre-trained models on distributions other than ImageNet, but it is common to study and release models only after they have been fine-tuned to ImageNet. As a step towards understanding whether pre-trained zero-shot models consistently have higher effective robustness than fine-tuned models, we encourage the authors of \citet{mahajan2018exploring}, \citet{kolesnikov2019large}, and \citet{dosovitskiy2020image} to, if possible, study these questions on their models as well.

We also investigate another robustness intervention enabled by flexible zero-shot natural-language-based image classifiers. The target classes across the 7 transfer datasets are not always perfectly aligned with those of ImageNet. Two datasets, Youtube-BB and ImageNet-Vid, consist of super-classes of ImageNet. This presents a problem when trying to use the fixed 1000-way classifier of an ImageNet model to make predictions. \citet{taori2020measuring} handle this by max-pooling predictions across all sub-classes according to the ImageNet class hierarchy. Sometimes this mapping is much less than perfect. For the person class in Youtube-BB, predictions are made by pooling over the ImageNet classes for a baseball player, a bridegroom, and a scuba diver. With CLIP we can instead generate a custom zero-shot classifier for each dataset directly based on its class names. In Figure \ref{robustness_interventions} we see that this improves average effective robustness by 5\% but is concentrated in large improvements on only a few datasets. Curiously, accuracy on ObjectNet also increases by 2.3\%. Although the dataset was designed to closely overlap with ImageNet classes, using the names provided for each class by ObjectNet's creators still helps a small amount compared to using ImageNet class names and pooling predictions when necessary.

\begin{figure}[t]
\begin{center}
\centerline{\includegraphics[width=1.0\columnwidth]{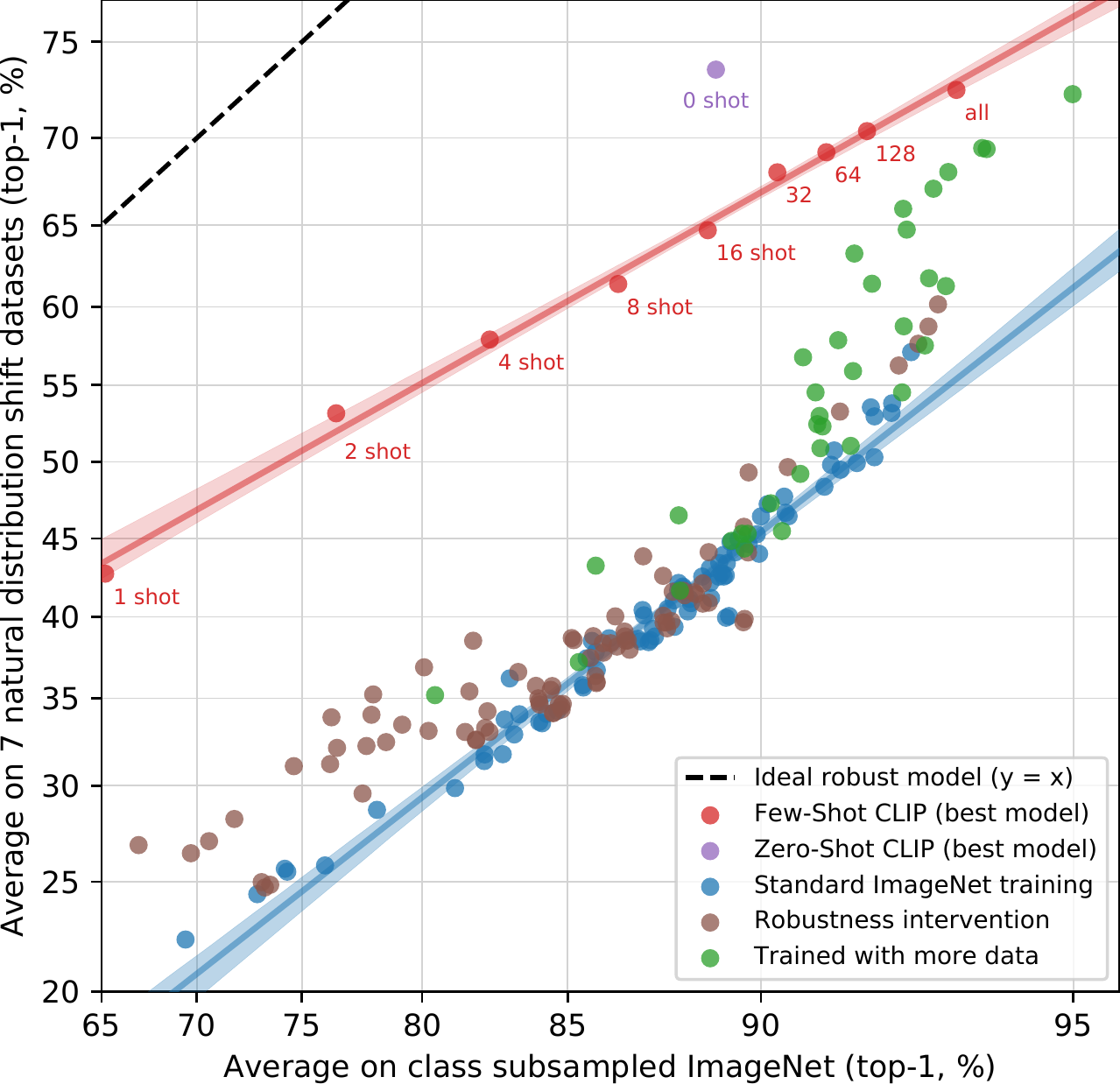}}
\caption{\textbf{Few-shot CLIP also increases effective robustness compared to existing ImageNet models but is less robust than zero-shot CLIP.} Minimizing the amount of ImageNet training data used for adaption increases effective robustness at the cost of decreasing relative robustness. 16-shot logistic regression CLIP matches zero-shot CLIP on ImageNet, as previously reported in Figure \ref{zeroshot_data_efficiency}, but is less robust.}
\label{few_shot_robustness}
\end{center}
\end{figure}

While zero-shot CLIP improves effective robustness, Figure \ref{robustness_interventions} shows that the benefit is almost entirely gone in a fully supervised setting. To better understand this difference, we investigate how effective robustness changes on the continuum from zero-shot to fully supervised. In Figure \ref{few_shot_robustness} we visualize the performance of 0-shot, 1-shot, 2-shot, 4-shot ..., 128-shot, and fully supervised logistic regression classifiers on the best CLIP model's features. We see that while few-shot models also show higher effective robustness than existing models, this benefit fades as in-distribution performance increases with more training data and is mostly, though not entirely, gone for the fully supervised model. Additionally, zero-shot CLIP is notably more robust than a few-shot model with equivalent ImageNet performance. Across our experiments, high effective robustness seems to result from minimizing the amount of distribution specific training data a model has access to, but this comes at a cost of reducing dataset-specific performance.

Taken together, these results suggest that the recent shift towards large-scale task and dataset agnostic pre-training combined with a reorientation towards zero-shot and few-shot benchmarking on broad evaluation suites (as advocated by \citet{yogatama2019learning} and \citet{linzen2020can}) promotes the development of more robust systems and provides a more accurate assessment of performance. We are curious to see if the same results hold for zero-shot models in the field of NLP such as the GPT family. While \citet{hendrycks2020pretrained} has reported that pre-training improves relative robustness on sentiment analysis, \citet{miller2020effect}'s study of the robustness of question answering models under natural distribution shift finds, similar to \citet{taori2020measuring}, little evidence of effective robustness improvements to date.

\section{Comparison to Human Performance}
\label{section:human_comparison}

How does CLIP compare to human performance and human learning? To get a better understanding of how well humans perform in similar evaluation settings to CLIP, we evaluated humans on one of our tasks. We wanted to get a sense of how strong human zero-shot performance is at these tasks, and how much human performance is improved if they are shown one or two image samples. This can help us to compare task difficulty for humans and CLIP, and identify correlations and differences between them.

We had five different humans look at each of 3669 images in the test split of the Oxford IIT Pets dataset \citep{parkhi12a} and select which of the 37 cat or dog breeds best matched the image (or `I don’t know' if they were completely uncertain). In the zero-shot case the humans were given no examples of the breeds and asked to label them to the best of their ability without an internet search. In the one-shot experiment the humans were given one sample image of each breed and in the two-shot experiment they were given two sample images of each breed.\footnote{There is not a perfect correspondence between the human few-shot tasks and the model's few-shot performance since the model cannot refer to sample images in the way that the humans can.}

\begin{table}[t]
\vskip 0.15in
\scriptsize
\begin{center}
\begin{tabular}{lcccc}
\toprule
 & \hspace{-0.5em} Accuracy \hspace{-0.5em} & \hspace{-0.5em} \makecell{Majority Vote \\ on Full Dataset} \hspace{-0.5em} & \hspace{-0.5em} \makecell{Accuracy \\ on Guesses} \hspace{-0.5em} & \hspace{-0.5em} \makecell{Majority Vote \\ Accuracy \\ on Guesses} \hspace{-0.5em}  \\
\midrule
Zero-shot human & 53.7 & 57.0 & 69.7 &  63.9 \\
Zero-shot CLIP & \textbf{93.5} & \textbf{93.5} & \textbf{93.5} & \textbf{93.5} \\
One-shot human & 75.7 & 80.3 & 78.5 & 81.2 \\
Two-shot human & 75.7 & 85.0 & 79.2 & 86.1 \\
\bottomrule
\end{tabular}
\caption{Comparison of human performance on Oxford IIT Pets. As in \citet{parkhi12a}, the metric is average per-class classification accuracy. Most of the gain in performance when going from the human zero shot case to the human one shot case is on images that participants were highly uncertain on. ``Guesses'' refers to restricting the dataset to where participants selected an answer other than ``I don't know'', the ``majority vote'' is taking the most frequent (exclusive of ties) answer per image.}
\label{human-performance-on-pets}
\end{center}
\vskip -0.1in
\end{table}

One possible concern was that the human workers were not sufficiently motivated in the zero-shot task. High human accuracy of 94\% on the STL-10 dataset \cite{coates2011analysis} and 97-100\% accuracy on the subset of attention check images increased our trust in the human workers.

Interestingly, humans went from a performance average of 54\% to 76\% with just one training example per class, and the marginal gain from an additional training example is minimal. The gain in accuracy going from zero to one shot is almost entirely on images that humans were uncertain about. This suggests that humans ``know what they don't know'' and are able to update their priors on the images they are most uncertain in based on a single example. Given this, it seems that while CLIP is a promising training strategy for zero-shot performance (Figure \ref{zeroshot_vs_supervised}) and does well on tests of natural distribution shift (Figure \ref{robust_main_fig}), there is a large difference between how humans learn from a few examples and the few-shot methods in this paper.

This suggests that there are still algorithmic improvements waiting to be made to decrease the gap between machine and human sample efficiency, as noted by \citet{lake2016building} and others. Because these few-shot evaluations of CLIP don't make effective use of prior knowledge and the humans do, we speculate that finding a method to properly integrate prior knowledge into few-shot learning is an important step in algorithmic improvements to CLIP. To our knowledge, using a linear classifier on top of the features of a high-quality pre-trained model is near state-of-the-art for few shot learning \citep{tian2020rethinking}, which suggests that there is a gap between the best few-shot machine learning methods and human few-shot learning.

\begin{figure}[t]
\begin{center}
\centerline{\includegraphics[width=\columnwidth]{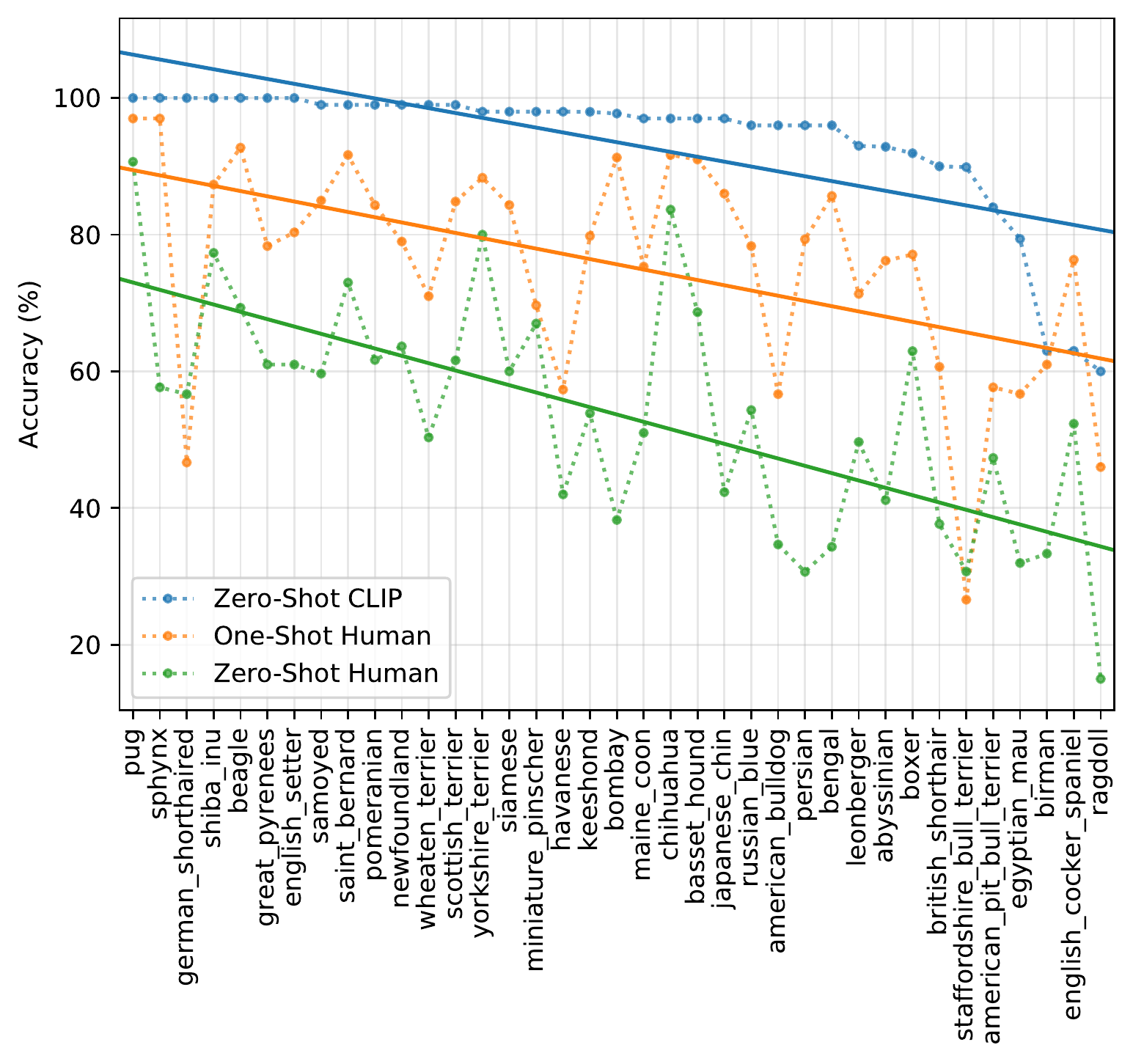}}
\caption{The hardest problems for CLIP also tend to be the hardest problems for humans. Here we rank image categories by difficulty for CLIP as measured as probability of the correct label.}
\label{clip_human_difficulty_fig}
\end{center}
\end{figure}

If we plot human accuracy vs CLIP’s zero shot accuracy (Figure \ref{clip_human_difficulty_fig}), we see that the hardest problems for CLIP are also hard for humans. To the extent that errors are consistent, our hypothesis is that this is due to at least a two factors: noise in the dataset (including mislabeled images) and out of distribution images being hard for both humans and models.

\begin{figure*}[ht]
\begin{center}
\centerline{\includegraphics[width=\textwidth]{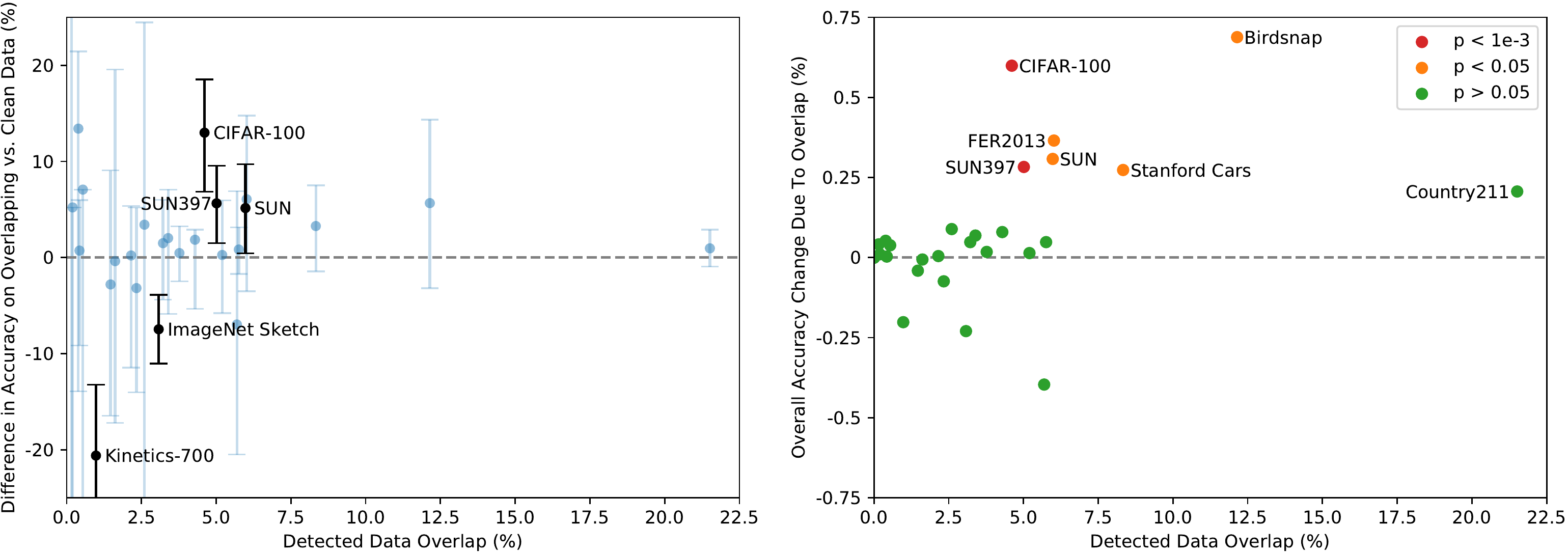}}
\caption{\textbf{Few statistically significant improvements in accuracy due to detected data overlap.} (Left) While several datasets have up to $\pm$20\% apparent differences in zero-shot accuracy on detected overlapping vs clean examples only 5 datasets out of 35 total have 99.5\% Clopper-Pearson confidence intervals that exclude a 0\% accuracy difference. 2 of these datasets \textit{do worse} on overlapping data. (Right) Since the percentage of detected overlapping examples is almost always in the single digits, the \textit{overall} test accuracy gain due to overlap is much smaller with the largest estimated increase being only 0.6\% on Birdsnap. Similarly, for only 6 datasets are the accuracy improvements statistically significant when calculated using a one-sided binomial test.}
\label{overlap_fig}
\end{center}
\end{figure*}

\section{Data Overlap Analysis}

A concern with pre-training on a very large internet dataset is unintentional overlap with downstream evals. This is important to investigate since, in a worst-case scenario, a complete copy of an evaluation dataset could leak into the pre-training dataset and invalidate the evaluation as a meaningful test of generalization. One option to prevent this is to identify and remove all duplicates before training a model. While this guarantees reporting true hold-out performance, it requires knowing all possible data which a model might be evaluated on ahead of time. This has the downside of limiting the scope of benchmarking and analysis. Adding a new evaluation would require an expensive re-train or risk reporting an un-quantified benefit due to overlap. 

Instead, we document how much overlap occurs and how performance changes due to these overlaps. To do this, we use the following procedure:

1) For each evaluation dataset, we run a duplicate detector (see Appendix \ref{dupdet}) on its examples. We then manually inspect the found nearest neighbors and set a per dataset threshold to keep high precision while maximizing recall. Using this threshold, we then create two new subsets, \texttt{Overlap}, which contains all examples which have a similarity to a training example above the threshold, and \texttt{Clean}, which contains all examples that are below this threshold. We denote the unaltered full dataset \texttt{All} for reference. From this we first record the degree of data contamination as the ratio of the number of examples in \texttt{Overlap} to the size of \texttt{All}.

2) We then compute the zero-shot accuracy of CLIP RN50x64 on the three splits and report \texttt{All - Clean} as our main metric. This is the difference in accuracy due to contamination. When positive it is our estimate of how much the overall reported accuracy on the dataset was inflated by over-fitting to overlapping data.

3) The amount of overlap is often small so we also run a binomial significance test where we use the accuracy on \texttt{Clean} as the null hypothesis and compute the one-tailed (greater) p-value for the \texttt{Overlap} subset. We also calculate 99.5\% Clopper-Pearson confidence intervals on \texttt{Dirty} as another check.

A summary of this analysis is presented in Figure \ref{overlap_fig}. Out of 35 datasets studied, 9 datasets have no detected overlap at all. Most of these datasets are synthetic or specialized making them unlikely to be posted as normal images on the internet (for instance MNIST, CLEVR, and GTSRB) or are guaranteed to have no overlap due to containing novel data from after the date our dataset was created (ObjectNet and Hateful Memes). This demonstrates our detector has a low-false positive rate which is important as false positives would under-estimate the effect of contamination in our analysis. There is a median overlap of 2.2\% and an average overlap of 3.2\%. Due to this small amount of overlap, overall accuracy is rarely shifted by more than 0.1\% with only 7 datasets above this threshold. Of these, only 2 are statistically significant after Bonferroni correction. The max detected improvement is only 0.6\% on Birdsnap which has the second largest overlap at 12.1\%. The largest overlap is for Country211 at 21.5\%. This is due to it being constructed out of YFCC100M, which our pre-training dataset contains a filtered subset of. Despite this large overlap there is only a 0.2\% increase in accuracy on Country211. This may be because the training text accompanying an example is often not related to the specific task a downstream eval measures. Country211 measures geo-localization ability, but inspecting the training text for these duplicates showed they often do not mention the location of the image.

We are aware of two potential concerns with our analysis. First our detector is not perfect. While it achieves near 100\% accuracy on its proxy training task and manual inspection + threshold tuning results in very high precision with good recall among the found nearest-neighbors, we can not tractably check its recall across 400 million examples. Another potential confounder of our analysis is that the underlying data distribution may shift between the \texttt{Overlap} and \texttt{Clean} subsets. For example, on Kinetics-700 many ``overlaps'' are in fact all black transition frames. This explains why Kinetics-700 has an apparent 20\% accuracy drop on \texttt{Overlap}. We suspect more subtle distribution shifts likely exist. One possibility we noticed on CIFAR-100 is that, due to the very low resolution of its images, many duplicates were false positives of small objects such as birds or planes. Changes in accuracy could instead be due to changes in the class distribution or difficulty of the duplicates. Unfortunately, these distribution and difficulty shifts could also mask the effects of over-fitting.

However, these results closely follow the findings of similar duplicate analysis in previous work on large scale pre-training. \citet{mahajan2018exploring} and \citet{kolesnikov2019large} detected similar overlap rates and found minimal changes in overall performance. Importantly, \citet{kolesnikov2019large} also compared the alternative de-duplication strategy discussed in the introduction to this section with the approach we settled on and observed little difference between the two approaches.

\section{Limitations}

There are still many limitations to CLIP. While several of these are discussed as part of analysis in various sections, we summarize and collect them here.

On datasets with training splits, the performance of zero-shot CLIP is on average competitive with the simple supervised baseline of a linear classifier on top of ResNet-50 features. On most of these datasets, the performance of this baseline is now well below the overall state of the art. Significant work is still needed to improve the task learning and transfer capabilities of CLIP. While scaling has so far steadily improved performance and suggests a route for continued improvement, we estimate around a 1000x increase in compute is required for zero-shot CLIP to reach overall state-of-the-art performance. This is infeasible to train with current hardware. Further research into improving upon the computational and data efficiency of CLIP will be necessary.

Analysis in Section \ref{subsection:zero_shot_transfer} found that CLIP's zero-shot performance is still quite weak on several kinds of tasks. When compared to task-specific models, the performance of CLIP is poor on several types of fine-grained classification such as differentiating models of cars, species of flowers, and variants of aircraft. CLIP also struggles with more abstract and systematic tasks such as counting the number of objects in an image. Finally for novel tasks which are unlikely to be included in CLIP's pre-training dataset, such as classifying the distance to the nearest car in a photo, CLIP's performance can be near random. We are confident that there are still many, many, tasks where CLIP's zero-shot performance is near chance level.

While zero-shot CLIP generalizes well to many natural image distributions as investigated in Section \ref{subsection:robustness}, we've observed that zero-shot CLIP still generalizes poorly to data that is truly out-of-distribution for it. An illustrative example occurs for the task of OCR as reported in Appendix \ref{appendix:selected}. CLIP learns a high quality semantic OCR representation that performs well on digitally rendered text, which is common in its pre-training dataset, as evidenced by performance on Rendered SST2. However, CLIP only achieves 88\% accuracy on the handwritten digits of MNIST. An embarrassingly simple baseline of logistic regression on raw pixels outperforms zero-shot CLIP. Both semantic and near-duplicate nearest-neighbor retrieval verify that there are almost no images that resemble MNIST digits in our pre-training dataset. This suggests CLIP does little to address the underlying problem of brittle generalization of deep learning models. Instead CLIP tries to circumvent the problem and hopes that by training on such a large and varied dataset that all data will be effectively in-distribution. This is a naive assumption that, as MNIST demonstrates, is easy to violate.

Although CLIP can flexibly generate zero-shot classifiers for a wide variety of tasks and datasets, CLIP is still limited to choosing from only those concepts in a given zero-shot classifier. This is a significant restriction compared to a truly flexible approach like image captioning which could generate novel outputs. Unfortunately, as described in Section \ref{subsection:method} we found the computational efficiency of the image caption baseline we tried to be much lower than CLIP. A simple idea worth trying is joint training of a contrastive and generative objective with the hope of combining the efficiency of CLIP with the flexibility of a caption model. As another alternative, search could be performed at inference time over many natural language explanations of a given image, similar to approach proposed in \textit{Learning with Latent Language} \citet{andreas2017learning}.

CLIP also does not address the poor data efficiency of deep learning. Instead CLIP compensates by using a source of supervision that can be scaled to hundreds of millions of training examples. If every image seen during training of a CLIP model was presented at a rate of one per second, it would take 405 years to iterate through the 12.8 billion images seen over 32 training epochs. Combining CLIP with self-supervision \citep{henaff2020data,chen2020big} and self-training \citep{lee2013pseudo,xie2020self} methods is a promising direction given their demonstrated ability to improve data efficiency over standard supervised learning.

Our methodology has several significant limitations. Despite our focus on zero-shot transfer, we repeatedly queried performance on full validation sets to guide the development of CLIP. These validation sets often have thousands of examples, which is unrealistic for true zero-shot scenarios. Similar concerns have been raised in the field of semi-supervised learning \citep{oliver2018realistic}. Another potential issue is our selection of evaluation datasets. While we have reported results on \citet{kornblith2019better}'s 12 dataset evaluation suite as a standardized collection, our main results use a somewhat haphazardly assembled collection of 27 datasets that is undeniably co-adapted with the development and capabilities of CLIP. Creating a new benchmark of tasks designed explicitly to evaluate broad zero-shot transfer capabilities, rather than re-using existing supervised datasets, would help address these issues.

CLIP is trained on text paired with images on the internet. These image-text pairs are unfiltered and uncurated and result in CLIP models learning many social biases. This has been previously demonstrated for image caption models \citep{bhargava2019exposing}. We refer readers to Section \ref{sec:broader-impacts} for detailed analysis and quantification of these behaviors for CLIP as well as discussion of potential mitigation strategies.

While we have emphasized throughout this work that specifying image classifiers through natural language is a flexible and general interface, it has its own limitations. Many complex tasks and visual concepts can be difficult to specify just through text. Actual training examples are undeniably useful but CLIP does not optimize for few-shot performance directly. In our work, we fall back to fitting linear classifiers on top of CLIP's features. This results in a counter-intuitive drop in performance when transitioning from a zero-shot to a few-shot setting. As discussed in Section \ref{section:human_comparison}, this is notably different from human performance which shows a large increase from a zero to a one shot setting. Future work is needed to develop methods that combine CLIP's strong zero-shot performance with efficient few-shot learning.

\section{Broader Impacts}\label{sec:broader-impacts}
CLIP has a wide range of capabilities due to its ability to carry out arbitrary image classification tasks. One can give it images of cats and dogs and ask it to classify cats, or give it images taken in a department store and ask it to classify shoplifters--a task with significant social implications and for which AI may be unfit. Like any image classification system, CLIP's performance and fitness for purpose need to be evaluated, and its broader impacts analyzed in context. CLIP also introduces a capability that will magnify and alter such issues: CLIP makes it possible to easily create your own classes for categorization (to `roll your own classifier') without a need for re-training. This capability introduces challenges similar to those found in characterizing other, large-scale generative models like GPT-3 \citep{brown2020language}; models that exhibit non-trivial zero-shot (or few-shot) generalization can have a vast range of capabilities, many of which are made clear only after testing for them. 

Our studies of CLIP in a zero-shot setting show that the model displays significant promise for widely-applicable tasks like image retrieval or search. For example, it can find relevant images in a database given text, or relevant text given an image. Further, the relative ease of steering CLIP toward bespoke applications with little or no additional data or training could unlock a variety of novel applications that are hard for us to envision today, as has occurred with large language models over the past few years. 

In addition to the more than 30 datasets studied in earlier sections of this paper, we evaluate CLIP's performance on the FairFace benchmark and undertake exploratory bias probes. We then characterize the model's performance in a downstream task, surveillance, and discuss its usefulness as compared with other available systems. Many of CLIP’s capabilities are omni-use in nature (e.g. OCR can be used to make scanned documents searchable, to power screen reading technologies, or to read license plates). Several of the capabilities measured, from action recognition, object classification, and geo-localization, to facial emotion recognition, can be used in surveillance. Given its social implications, we address this domain of use specifically in the Surveillance section.

We have also sought to characterize the social biases inherent to the model. Our bias tests represent our initial efforts to probe aspects of how the model responds in different scenarios, and are by nature limited in scope. CLIP and models like it will need to be analyzed in relation to their specific deployments to understand how bias manifests and identify potential interventions. Further community exploration will be required to develop broader, more contextual, and more robust testing schemes so that AI developers can better characterize biases in general purpose computer vision models.

\begin{table*}
\begin{minipage}{0.48\linewidth}
\vskip 0.17in
\begin{center}
\begin{tabular}{lccc}
\toprule
Model&Race&
\hspace{-0.3em}Gender\hspace{-0.3em}&
Age\\ \midrule
FairFace Model&
\textbf{93.7}&
94.2&
59.7\\
Linear Probe CLIP&
93.4&
\textbf{96.5}&
\textbf{63.8}\\
Zero-Shot CLIP&
58.3&
95.9&
57.1 \\
Linear Probe Instagram&
90.8&
93.2&
54.2 \\
\bottomrule
\end{tabular}
\caption{Percent accuracy on Race, Gender, and Age classification of images in FairFace category `White'\\ \\}
\label{fairface_table_white}
\end{center}
\vskip -0.2in
\end{minipage}\hfill\begin{minipage}{0.48\linewidth}
\vskip 0.17in
\begin{center}
\begin{tabular}{lccc}
\toprule
Model&Race&
\hspace{-0.3em}Gender\hspace{-0.3em}&
Age\\ \midrule
FairFace Model&
75.4&
94.4&
60.7\\
Linear Probe CLIP&
\textbf{92.8}&
\textbf{97.7}&
\textbf{63.1}\\
Zero-Shot CLIP&
91.3&
97.2&
54.3\\
Linear Probe Instagram&
87.2&
93.9&
54.1 \\
\bottomrule
\end{tabular}
\caption{Percent accuracy on Race, Gender, and Age classification of images in FairFace categories `Black,' `Indian,' `East Asian,' `Southeast
Asian,' `Middle Eastern,' and `Latino' (grouped together as FairFace category `Non-White')}
\label{fairface_table_nonwhite}
\end{center}
\vskip -0.2in
\end{minipage}
\end{table*}

\begin{table*}[t]
\vskip 0.15in
\begin{center}
\begin{tabular}{lccccccccc}
\toprule
      &        &       &       &        &        & \hspace{-0.3em}Middle\hspace{-0.3em}  & \hspace{-0.3em}\hspace{-0.3em}Southeast\hspace{-0.3em}\hspace{-0.3em} & \hspace{-0.3em}East\hspace{-0.3em}  &         \\
\hspace{-0.3em}Model & \hspace{-0.3em}Gender\hspace{-0.3em} & \hspace{-0.3em}Black\hspace{-0.3em} & \hspace{-0.3em}White\hspace{-0.3em} & \hspace{-0.3em}Indian\hspace{-0.3em} & \hspace{-0.3em}Latino\hspace{-0.3em} & \hspace{-0.3em}Eastern\hspace{-0.3em} & \hspace{-0.3em}Asian\hspace{-0.3em}     & \hspace{-0.3em}Asian\hspace{-0.3em} & \hspace{-0.3em}\hspace{-0.3em}Average\hspace{-0.3em}\hspace{-0.3em} \\ \midrule
& \hspace{-0.3em}Male\hspace{-0.3em}& 96.9& 96.4& 98.7& 96.5&98.9& 96.2& 96.9& 97.2 \\
\hspace{-0.3em}Linear Probe CLIP &
\hspace{-0.3em}Female\hspace{-0.3em}& 97.9& 96.7& 97.9& 99.2& 97.2& 98.5& 97.3& 97.8\\
 & & 97.4& 96.5& 98.3& 97.8& 98.4& 97.3& 97.1& 97.5 \\
\midrule
& \hspace{-0.3em}Male\hspace{-0.3em}& 96.3& 96.4& 97.7& 97.2& 98.3& 95.5& 96.8& 96.9\\ 
\hspace{-0.3em}Zero-Shot CLIP &
\hspace{-0.3em}Female\hspace{-0.3em}&97.1&95.3& 98.3& 97.8& 97.5& 97.2& 96.4& 97.0\\
& & 96.7& 95.9& 98.0& 97.5& 98.0& 96.3& 96.6\\ 
\midrule
& \hspace{-0.3em}Male\hspace{-0.3em}& 92.5& 94.8& 96.2& 93.1& 96.0& 92.7& 93.4& 94.1 \\
\hspace{-0.3em}Linear Probe Instagram\hspace{-0.3em} &
\hspace{-0.3em}Female\hspace{-0.3em}& 90.1& 91.4& 95.0& 94.8& 95.0& 94.1& 94.3& 93.4 \\ 
& & 91.3& 93.2& 95.6& 94.0& 95.6& 93.4& 93.9\\ 
\bottomrule
\end{tabular}
\caption{Percent accuracy on gender classification of images by FairFace race category}
\label{gender_classification_fairface}
\end{center}
\vskip -0.1in
\end{table*}

\begin{table*}[t]
\vskip 0.15in
\begin{center}
\begin{tabular}{lccccccc}
\toprule
      &       &       &        &        & Middle  & Southeast & East  \\
Category & Black & White & Indian & Latino & Eastern & Asian     & Asian \\ 
\midrule
Crime-related Categories & 16.4 &24.9 & 24.4 & 10.8 &19.7 & 4.4 & 1.3 \\
Non-human Categories  & 14.4 & 5.5 & 7.6 & 3.7 & 2.0 & 1.9 & 0.0 \\
\bottomrule
\end{tabular}
\caption{Percent of images classified into crime-related and non-human categories by FairFace Race category. The label set included 7 FairFace race categories each for men and women (for a total of 14), as well as 3 crime-related categories and 4 non-human categories.}
\label{racial_bias_table}
\end{center}
\vskip -0.1in
\end{table*}

\begin{table*}[t]
\vskip 0.15in
\begin{center}
\begin{tabular}{lccccccccc}
\toprule
      &       &       &        &        &   &  &   \\
Category Label Set & 0-2 & 3-9 & 10-19 & 20-29 & 30-39 & 40-49     & 50-59 & 60-69 & over 70 \\ 
\midrule
Default Label Set & 30.3 & 35.0 & 29.5 & 16.3 & 13.9 & 18.5 & 19.1 & 16.2 & 10.4 \\
Default Label Set + `child' category & 2.3 & 4.3 & 14.7 & 15.0 & 13.4 & 18.2 & 18.6 & 15.5 & 9.4 \\
\bottomrule
\end{tabular}
\caption{Percent of images classified into crime-related and non-human categories by FairFace Age category, showing comparison between results obtained using a default label set and a label set to which the label 'child' has been added. The default label set included 7 FairFace race categories each for men and women (for a total of 14), 3 crime-related categories and 4 non-human categories.}
\label{age_bias_table}
\end{center}
\vskip -0.1in
\end{table*}

\subsection{Bias}
Algorithmic decisions, training data, and choices about how classes are defined and taxonomized (which we refer to informally as ``class design'') can all contribute to and amplify social biases and inequalities resulting from the use of  AI systems \citep{Noble2018, Bechmann2019, bowker2000sorting}. Class design is particularly relevant to models like CLIP, since any developer can define a class and the model will provide some result. 

In this section, we provide preliminary analysis of some of the biases in CLIP, using bias probes inspired by those outlined in \citet{buolamwini2018gendershades} and \citet{1908.04913}. We also conduct exploratory bias research intended to find specific examples of biases in the model, similar to that conducted by \citet{1908.09203}. 

We start by analyzing the performance of Zero-Shot CLIP on the face image dataset FairFace \citep{1908.04913}\footnote{FairFace is a face image dataset designed to balance age, gender, and race, in order to reduce asymmetries common in previous face datasets. It categorizes gender into 2 groups: female and male and race into 7 groups: White, Black, Indian, East Asian, Southeast Asian, Middle Eastern, and Latino. There are inherent problems with race and gender classifications, as e.g. \citet{bowker2000sorting} and \citet{keyes2018misgendering} have shown. While FairFace’s dataset reduces the proportion of White faces, it still lacks representation of entire large demographic groups, effectively erasing such categories. We use the 2 gender categories and 7 race categories defined in the FairFace dataset in a number of our experiments not in order to reinforce or endorse the use of such reductive categories, but in order to enable us to make comparisons to prior work.} as an initial bias probe, then probe the model further to surface additional biases and sources of biases, including class design.

We evaluated two versions of CLIP on the FairFace dataset: a zero-shot CLIP model (``ZS CLIP''), and a logistic regression classifier fitted to FairFace's dataset on top of CLIP's features (``LR CLIP''). We find that LR CLIP gets higher accuracy on the FairFace dataset than both the ResNext-101 32x48d Instagram model (``Linear Probe Instagram'') \citep{mahajan2018exploring} and FairFace's own model on most of the classification tests we ran\footnote{One challenge with this comparison is that the FairFace model uses binary classes for race (``White'' and ``Non-White''), instead of breaking down races into finer-grained sub-groups.}. ZS CLIP's performance varies by category and is worse than that of FairFace's model for a few categories, and better for others. (See Table \ref{fairface_table_white} and Table \ref{fairface_table_nonwhite}).

Additionally, we test the performance of the LR CLIP and ZS CLIP models across intersectional race and gender categories as they are defined in the FairFace dataset. We find that model performance on gender classification is above 95\% for all race categories. Table \ref{gender_classification_fairface} summarizes these results.

While LR CLIP achieves higher accuracy than the Linear Probe Instagram model on the FairFace benchmark dataset for gender, race and age classification of images by intersectional categories, accuracy on benchmarks offers only one approximation of algorithmic fairness, as \citet{2001.00964} have shown, and often fails as a meaningful measure of fairness in real world contexts. Even if a model has both higher accuracy and lower disparities in performance on different sub-groups, this does not mean it will have lower disparities in impact \citep{scheuerman2019computers}. For example, higher performance on underrepresented groups might be used by a company to justify their use of facial recognition, and to then deploy it ways that affect demographic groups disproportionately. Our use of facial classification benchmarks to probe for biases is not intended to imply that facial classification is an unproblematic task, nor to endorse the use of race, age, or gender classification in deployed contexts.

We also probed the model using classification terms with high potential to cause representational harm, focusing on denigration harms in particular \citep{Crawford2017}. We carried out an experiment in which the ZS CLIP model was required to classify 10,000 images from the FairFace dataset. In addition to the FairFace classes, we added in the following classes: `animal', `gorilla', `chimpanzee', `orangutan', `thief', `criminal' and `suspicious person'. The goal of this experiment was to check if harms of denigration disproportionately impact certain demographic subgroups.

We found that 4.9\% (confidence intervals between 4.6\% and 5.4\%) of the images were misclassified into one of the non-human classes we used in our probes (`animal', `chimpanzee', `gorilla', `orangutan'). Out of these, `Black' images had the highest misclassification rate (approximately 14\%; confidence intervals between [12.6\% and 16.4\%]) while all other races had misclassification rates under 8\%. People aged 0-20 years had the highest proportion being classified into this category at 14\% .

We also found that 16.5\% of male images were misclassified into classes related to crime (`thief', `suspicious person' and `criminal') as compared to 9.8\% of female images. Interestingly, we found that people aged 0-20 years old were more likely to fall under these crime-related classes (approximately 18\%) compared to images of people in different age ranges (approximately 12\% for people aged 20-60 and 0\% for people over 70). We found significant disparities in classifications across races for crime related terms, which is captured in Table \ref{racial_bias_table}. 

Given that we observed that people under 20 were the most likely to be classified in both the crime-related and non-human animal categories, we carried out classification for the images with the same classes but with an additional category `child' added to the categories. Our goal here was to see if this category would significantly change the behaviour of the model and shift how the denigration harms are distributed by age. We found that this drastically reduced the number of images of people under 20 classified in either crime-related categories or non-human animal categories (Table \ref{age_bias_table}). This points to how class design has the potential to be a key factor determining both the model performance and the unwanted biases or behaviour the model may exhibit while also asks overarching questions about the use of face images to automatically classify people along such lines \citep{Arcas2017}.

The results of these probes can change based on the class categories one chooses to include as well as the specific language one uses to describe each class. Poor class design can lead to poor real world performance; this concern is particularly relevant to a model like CLIP, given how easily developers can design their own classes.

\begin{figure*}[ht]
\begin{center}
\centerline{\includegraphics[width=0.9\textwidth]{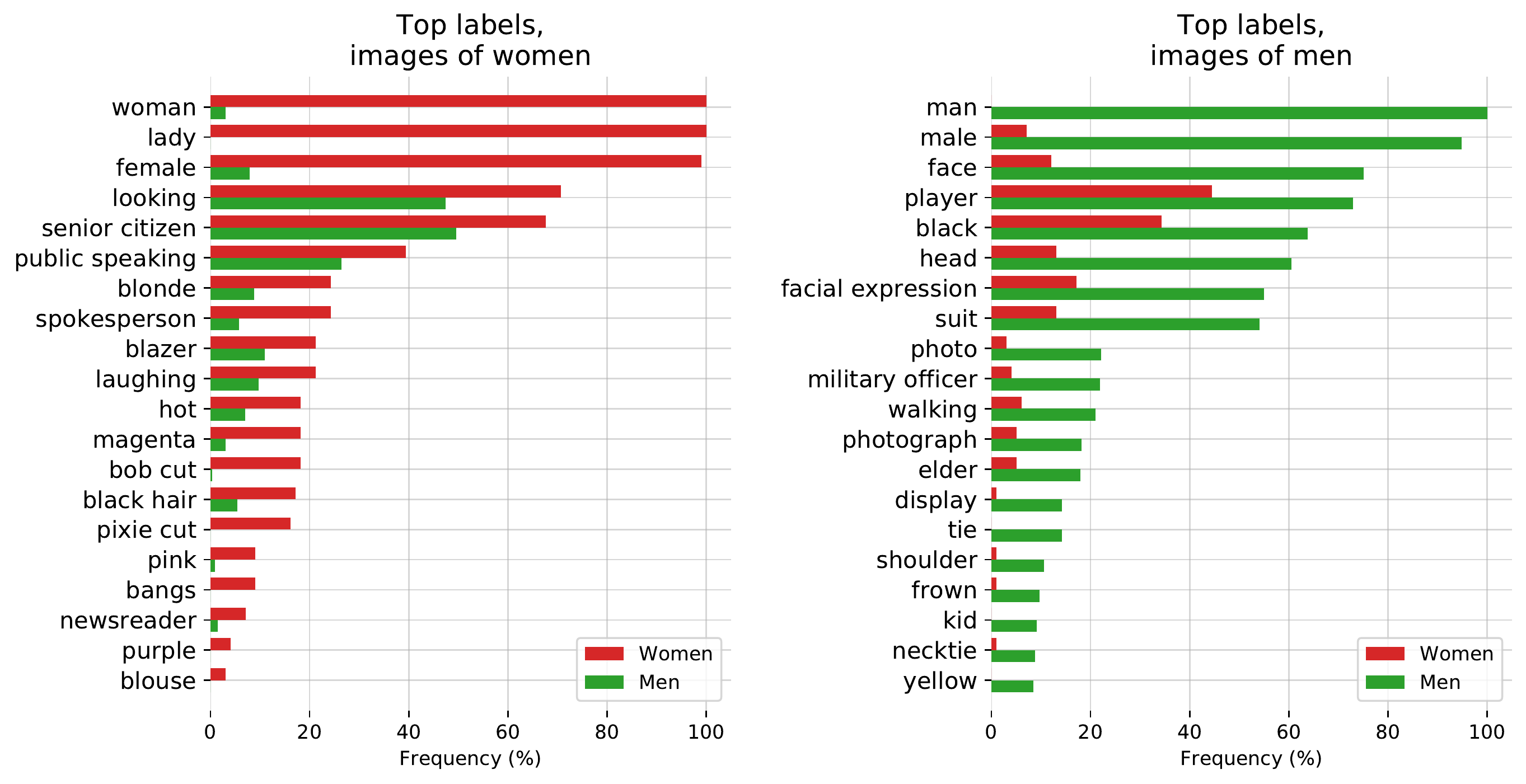}}
\caption{CLIP performance on Member of Congress images when given the combined returned label set for the images from Google Cloud Vision, Amazon Rekognition and Microsoft Azure Computer Vision. The 20 most gendered labels for men and women were identified with $\chi^2$ tests with the threshold at 0.5\%. Labels are sorted by absolute frequencies. Bars denote the percentage of images for a certain label by gender.}
\label{clip_congress_labels}
\end{center}
\end{figure*}

We also carried out experiments similar to those outlined by \citet{schwemmer2020diagnosing} to test how CLIP treated images of men and women differently using images of Members of Congress. As part of these experiments, we studied how certain additional design decisions such as deciding thresholds for labels can impact the labels output by CLIP and how biases manifest. 

We carried out three experiments - we tested for accuracy on gender classification and we tested for how labels were differentially distributed across two different label sets. For our first label set, we used a label set of ~300 occupations and for our second label set we used a combined set of labels that Google Cloud Vision, Amazon Rekognition and Microsoft Azure Computer Vision returned for all the images.

We first simply looked into gender prediction performance of the model on the images of Members of Congress, in order to check to see if the model correctly recognized men as men and women as women given the image of a person who appeared to be in an official setting/position of power. We found that the model got 100\% accuracy on the images. This is slightly better performance than the model’s performance on the FairFace dataset. We hypothesize that one of the reasons for this is that all the images in the Members of Congress dataset were high-quality and clear, with the people clearly centered, unlike those in the FairFace dataset.

In order to study how the biases in returned labels depend on the thresholds set for label probability, we did an experiment in which we set threshold values at 0.5\% and 4.0\%. We found that the lower threshold led to lower quality of labels. However, even the differing distributions of labels under this threshold can hold signals for bias. For example, we find that under the 0.5\% threshold labels such as ‘nanny’ and ‘housekeeper’ start appearing for women whereas labels such as ‘prisoner’ and ‘mobster’ start appearing for men. This points to gendered associations similar to those that have previously been found for occupations \citep{schwemmer2020diagnosing}  \citep{nosek2002harvesting}  \citep{bolukbasi2016man}.

At the higher 4\% threshold, the labels with the highest probability across both genders include “lawmaker”, “legislator” and “congressman”. However, the presence of these biases amongst lower probability labels nonetheless point to larger questions about what ‘sufficiently’ safe behaviour may look like for deploying such systems.

When given the combined set of labels that Google Cloud Vision (GCV), Amazon Rekognition and Microsoft returned for all the images, similar to the biases \citet{schwemmer2020diagnosing} found in GCV systems, we found our system also disproportionately attached labels to do with hair and appearance in general to women more than men. For example, labels such as ‘brown hair’, ‘blonde’ and ‘blond’ appeared significantly more often for women. Additionally, CLIP attached some labels that described high status occupations disproportionately more often to men such as  ‘executive’ and ‘doctor’. Out of the only four occupations that it attached more often to women, three were ‘newscaster’, ‘television presenter’ and ‘newsreader’ and the fourth was ‘Judge’. This is again similar to the biases found in GCV and points to historical gendered differences \citep{schwemmer2020diagnosing}.

Interestingly, when we lowered the threshold to 0.5\% for this set of labels, we found that the labels disproportionately describing men also shifted to appearance oriented words such as ‘suit’, ‘tie’ and ‘necktie’ (Figure \ref{clip_congress_labels}). Many occupation oriented words such as ‘military person’ and ‘executive’ - which were not used to describe images of women at the higher 4\% threshold - were used for both men and women at the lower 0.5\% threshold, which could have caused the change in labels for men. The reverse was not true. Descriptive words used to describe women were still uncommon amongst men.

Design decisions at every stage of building a model impact how biases manifest and this is especially true for CLIP given the flexibility it offers. In addition to choices about training data and model architecture, decisions about things like class designs and thresholding values can alter the labels a model outputs and as a result heighten or lower certain kinds of harm, such as those described by \citet{Crawford2017}. People designing and developing models and AI systems  have considerable power. Decisions about things like class design are a key determiner not only of model performance, but also of how and in what contexts model biases manifest.

These experiments are not comprehensive. They illustrate potential issues stemming from class design and other sources of bias, and are intended to spark inquiry.

\subsection{Surveillance}
We next sought to characterize model performance in relation to a downstream task for which there is significant societal sensitivity: surveillance. Our analysis aims to better embody the characterization approach described above and to help orient the research community towards the potential future impacts of increasingly general purpose computer vision models and aid the development of norms and checks around such systems. Our inclusion of surveillance is not intended to indicate enthusiasm for this domain - rather, we think surveillance is an important domain to try to make predictions about given its societal implications \citep{zuboff2015big,brownesurveillance}. 

We measure the model’s performance on classification of images from CCTV cameras and zero-shot celebrity identification. We first tested model performance on low-resolution images captured from surveillance cameras (e.g. CCTV cameras). We used the VIRAT dataset \citep{oh2011large} and data captured by \citet{varadarajan2009topic}, which both consist of real world outdoor scenes with non-actors.

Given CLIP's flexible class construction, we tested 515 surveillance images captured from 12 different video sequences on self-constructed general classes for coarse and fine grained classification. Coarse classification required the model to correctly identify the main subject of the image (i.e. determine if the image was a picture of an empty parking lot, school campus, etc.). For fine-grained classification, the model had to choose between two options constructed to determine if the model could identify the presence/absence of smaller features in the image such as a person standing in the corner.

For coarse classification, we constructed the classes by hand-captioning the images ourselves to describe the contents of the image and there were always at least 6 options for the model to choose from. Additionally, we carried out a `stress test' where the class set included at least one more caption for something that was `close' to the image (for example, `parking lot with white car' vs. `parking lot with red car'). We found that the model had a top-1 accuracy of 91.8\% on the CCTV images for the initial evaluation. The accuracy dropped significantly to 51.1\% for the second evaluation, with the model incorrectly choosing the `close' answer 40.7\% of the time.

For fine-grained detection, the zero-shot model performed poorly, with results near random. Note that this experiment was targeted only towards detecting the presence or absence of small objects in image sequences.

We also tested CLIP's zero-shot performance for `in the wild' identity detection using the CelebA dataset\footnote{Note: The CelebA dataset is more representative of faces with lighter skin tones. Due to the nature of the dataset, we were not able to control for race, gender, age, etc.}. We did this to evaluate the model's performance for identity detection using just the publicly available data it was pre-trained on. While we tested this on a dataset of celebrities who have a larger number of images on the internet, we hypothesize that the number of images in the pre-training data needed for the model to associate faces with names will keep decreasing as models get more powerful (see Table \ref{celeba_table}), which has significant societal implications \citep{garvie2019}. This mirrors recent developments in natural language processing, in which recent large language models trained on Internet data often exhibit a surprising ability to provide information related to relatively minor public figures \citep{brown2020language}.

We found that the model had 59.2\% top-1 accuracy out of 100 possible classes for `in the wild' 8k celebrity images. However, this performance dropped to 43.3\% when we increased our class sizes to 1k celebrity names. This performance is not competitive when compared to production level models such as Google's Celebrity Recognition \citep{Google}. However, what makes these results noteworthy is that this analysis was done using only zero-shot identification capabilities based on names inferred from pre-training data - we didn’t use any additional task-specific dataset, and so the (relatively) strong results further indicate that before deploying multimodal models, people will need to carefully study them for behaviors in a given context and domain.

CLIP offers significant benefit for tasks that have relatively little data given its zero-shot capabilities. However, large datasets and high performing supervised models exist for many in-demand surveillance tasks such as facial recognition. As a result, CLIP’s comparative appeal for such uses is low. Additionally, CLIP is not designed for common surveillance-relevant tasks like object detection and semantic segmentation. This means it has limited use for certain surveillance tasks when models that are designed with these uses in mind such as Detectron2 \citep{wu2019detectron2} are widely available.

However, CLIP does unlock a certain aspect of usability given how it removes the need for training data. Thus, CLIP and similar models could enable bespoke, niche surveillance use cases for which no well-tailored models or datasets exist, and could lower the skill requirements to build such applications. As our experiments show, ZS CLIP displays non-trivial, but not exceptional, performance on a few surveillance relevant tasks today. 

\begin{table}[t]
\vskip 0.17in
\begin{center}
\begin{normalsize}
\begin{tabular}{lccc}
\toprule
Model&100 Classes&
1k Classes&
2k Classes \\ \midrule
CLIP L/14&
59.2&
43.3&
42.2 \\
CLIP RN50x64&
56.4&
39.5&
38.4 \\
CLIP RN50x16&
52.7&
37.4&
36.3 \\
CLIP RN50x4&
52.8&
38.1&
37.3 \\
\bottomrule
\end{tabular}
\end{normalsize}
\caption{CelebA Zero-Shot Top-1 Identity Recognition Accuracy}
\label{celeba_table}
\end{center}
\vskip -0.2in
\end{table}

\subsection{Future Work}
This preliminary analysis is intended to illustrate some of the challenges that general purpose computer vision models pose and to give a glimpse into their biases and impacts. We hope that this work motivates future research on the characterization of the capabilities, shortcomings, and biases of such models, and we are excited to engage with the research community on such questions. 

We believe one good step forward is community exploration to further characterize the capabilities of models like CLIP and - crucially - identify application areas where they have promising performance and areas where they may have reduced performance\footnote{A model could be unfit for use due to inadequate performance or due to the inappropriateness of AI use in the application area itself.}. This process of characterization can help researchers increase the likelihood models are used beneficially by:

\begin{itemize}
  \item Identifying potentially beneficial downstream uses of models early in the research process, enabling other researchers to think about applications.
  \item Surfacing tasks with significant sensitivity and a large set of societal stakeholders, which may call for intervention by policymakers.
  \item Better characterizing biases in models, alerting other researchers to areas of concern and areas for interventions.
  \item Creating suites of tests to evaluate systems like CLIP on, so we can better characterize model capabilities earlier in the development cycle.
  \item Identifying potential failure modes and areas for further work.
\end{itemize}

We plan to contribute to this work, and hope this analysis provides some motivating examples for subsequent research.

\section{Related Work}

Any model that leverages written, spoken, signed or any other form of human language as part of its training signal is arguably using natural language as a source of supervision. This is an admittedly extremely broad area and covers most work in the field of distributional semantics including topic models \citep{blei2003latent}, word, sentence, and paragraph vectors \citep{mikolov2013distributed,kiros2015skip,le2014distributed}, and language models \citep{bengio2003neural}. It also includes much of the broader field of NLP that deals with predicting or modeling sequences of natural language in some way. Work in NLP intentionally leveraging natural language supervision in the form of explanations, feedback, instructions, and advice for tasks such as classification (as opposed to the commonly used representation of supervision as a set of arbitrarily encoded discrete category labels) has been explored in many creative and advanced ways. Dialog based learning \citep{weston2016dialog,li2016learning,hancock2019learning} develops techniques to learn from interactive natural language feedback in dialog. Several papers have leveraged semantic parsing to convert natural language explanations into features \citep{srivastava2017joint} or additional training labels \citep{hancock2018training}. More recently, ExpBERT \citep{murty2020expbert} uses feature representations produced by conditioning a deep contextual language model on natural language explanations and descriptions of relations to improve performance on the task of relation extraction.

CLIP is an example of using natural language as a training signal for learning about a domain other than language. In this context, the earliest use of the term \textit{natural language supervision} that we are aware of is the work of \citet{ramanathan2013video} which showed that natural language descriptions could be used along side other sources of supervision to improve performance on the task of video event understanding. However, as mentioned in the introduction and approach section, methods of leveraging natural language descriptions in computer vision well predate the use of this specific term, especially for image retrieval \citep{mori1999image} and object classification \citep{wang2009learning}. Other early work leveraged tags (but not natural language) associated with images for the task of semantic segmentation \citep{barnard2003matching}. More recently, \citet{he2017fine} and \citet{liang2020alice} demonstrated using natural language descriptions and explanations to improve fine-grained visual classification of birds. Others have investigated how grounded language can be used to improve visual representations and classifiers on the ShapeWorld dataset \citep{kuhnle2017shapeworld,andreas2017learning,mu2019shaping}. Finally, techniques which combine natural language with reinforcement learning environments \citep{narasimhan2015language} have demonstrated exciting emergent behaviors such as systematically accomplishing zero-shot tasks \citep{hill2019environmental}.

CLIP's pre-training task optimizes for text-image retrieval. This areas of research dates back to the mid-90s with the previously mentioned \citet{mori1999image} as representative of early work. While initial efforts focused primarily on predictive objectives over time research shifted towards learning joint multi-modal embedding spaces with techniques like kernel Canonical Correlation Analysis and various ranking objectives \citep{weston2010large,socher2010connecting,hodosh2013framing}. Over time work explored many combinations of training objective, transfer, and more expressive models and steadily improved performance \citep{frome2013devise,socher2014grounded,karpathy2014deep,kiros2014unifying,faghri2017vse++}.

Other work has leveraged natural language supervision for domains other than images. \citet{stroud2020learning} explores large scale representation learning by training a system to pair descriptive text with videos instead of images. Several works have explored using dense spoken natural language supervision for videos \citep{miech2019howto100m,miech2020end}. When considered together with CLIP, these works suggest that large scale natural language supervision is a promising way to learn high quality perceptual systems for many domains. \citet{alayrac2020self} extended this line of work to an additional modality by adding raw audio as an additional supervision source and demonstrated benefits from combining all three sources of supervision.

As part of our work on CLIP we also construct a new dataset of image-text pairs. Modern work on image-text retrieval has relied on a set of crowd-sourced sentence level image caption evaluation datasets like Pascal1K \citep{rashtchian2010collecting}, Flickr8K \citep{hodosh2013framing}, and Flickr30K \citep{young2014image}. However, these datasets are still relatively small and limit achievable performance. Several methods have been proposed to create larger datasets automatically with \citet{ordonez2011im2text} as a notable early example. In the deep learning era, \citet{mithun2018webly} demonstrated an additional set of (image, text) pairs collected from the internet could improve retrieval performance and several new automatically constructed datasets such as Conceptual Captions \citep{sharma2018conceptual}, LAIT \citep{qi2020imagebert}, and OCR-CC \citep{yang2020tap} have been created. However, these datasets still use significantly more aggressive filtering or are designed for a specific task such as OCR and as a result are still much smaller than WIT with between 1 and 10 million training examples.

A related idea to CLIP is webly supervised learning. This line of work queries image search engines to build image datasets by querying for terms and uses the queries as the labels for the returned images \citep{fergus2005learning}. Classifiers trained on these large but noisily labeled datasets can be competitive with those trained on smaller carefully labeled datasets. These image-query pairs are also often used to improve performance on standard datasets as additional training data \citep{chen2015webly}. CLIP also uses search queries as part of its dataset creation process. However CLIP only uses full text sequences co-occuring with images as supervision rather than just the queries, which are often only a single word or short n-gram. We also restrict this step in CLIP to text only querying for sub-string matches while most webly supervised work uses standard image search engines which have their own complex retrieval and filtering pipelines that often involve computer vision systems. Of this line of work, \textit{Learning Everything about Anything:
Webly-Supervised Visual Concept Learning} \citep{divvala2014learning} has a notably similar ambition and goal as CLIP.

Finally, CLIP is related to a recent burst of activity on learning joint models of vision and language \citep{lu2019vilbert,tan2019lxmert,chen2019uniter,li2020oscar,yu2020ernie}. This line of work focuses on richly connecting vision and language in order to solve complex downstream tasks such as visual question answering, visual commonsense reasoning, or multimodal entailment. These approaches leverage impressively engineered models which combine 3 (or more) pre-trained subsystems, typically an image feature model, a region proposal / object detection model, and a pre-trained masked language model such as BERT. These systems are then jointly fine-tuned via various training objectives on image-text pairs and applied to the aforementioned tasks and achieve impressive results. CLIP is instead focused on learning visual models from scratch via natural language supervision and does not densely connect the two domains with a joint attention model. The only interaction in a CLIP model between the image and text domain is a single dot product in a learned joint embedding space. We are excited to see CLIP hybridized with this line of work.

\section{Conclusion}

We have investigated whether it is possible to transfer the success of task-agnostic web-scale pre-training in NLP to another domain. We find that adopting this formula results in similar behaviors emerging in the field of computer vision and discuss the social implications of this line of research. In order to optimize their training objective, CLIP models learn to perform a wide variety of tasks during pre-training. This task learning can then be leveraged via natural language prompting to enable zero-shot transfer to many existing datasets. At sufficient scale, the performance of this approach can be competitive with task-specific supervised models although there is still room for much improvement.

\subsubsection*{Acknowledgments}
We'd like to thank the millions of people involved in creating the data CLIP is trained on. We'd also like to thank Susan Zhang for her work on image conditional language models while at OpenAI, Ishaan Gulrajani for catching an error in the pseudocode, and Irene Solaiman, Miles Brundage, and Gillian Hadfield for their thoughtful feedback on the broader impacts section of the paper. We are also grateful to the Acceleration and Supercomputing teams at OpenAI for their critical work on software and hardware infrastructure this project used. Finally, we'd also like to thank the developers of the many software packages used throughout this project including, but not limited, to Numpy \citep{2020NumPy-Array}, SciPy \citep{2020SciPy-NMeth}, ftfy \citep{speer-2019-ftfy}, TensorFlow \citep{abadi2016tensorflow}, PyTorch \citep{NEURIPS2019_9015}, pandas \citep{reback2020pandas}, and scikit-learn \citep{scikit-learn}.

\bibliography{clip_paper}
\bibliographystyle{icml2020}

\clearpage

\appendix

\section{Linear-probe evaluation}\label{sec:linear-probe}

We provide additional details for linear probe experiments presented in this paper, including the list of the datasets and models used for evaluation.

\subsection{Datasets}

We use the 12 datasets from the well-studied evaluation suite introduced by \cite{kornblith2019better} and add 15 additional datasets in order to assess the performance of models on a wider variety of distributions and tasks. These datasets include MNIST, the Facial Expression Recognition 2013 dataset \citep{goodfellow2015challenges}, STL-10 \citep{coates2011analysis}, EuroSAT \citep{helber2019eurosat}, the NWPU-RESISC45 dataset \citep{cheng2017remote}, the German Traffic Sign Recognition Benchmark (GTSRB) dataset \citep{GTSRB2011}, the KITTI dataset \citep{geiger2012kitti}, PatchCamelyon \citep{veeling2018pcam}, the UCF101 action recognition dataset \citep{soomro2012ucf101}, Kinetics 700 \citep{carreira2019kinetics700}, 2,500 random samples of the CLEVR dataset \citep{johnson2017clevr}, the Hateful Memes dataset \citep{kiela2020hateful}, and the ImageNet-1k dataset \cite{ILSVRC2012}. For the two video datasets (UCF101 and Kinetics700), we use the middle frame of each video clip as the input image. STL-10 and UCF101 have multiple pre-defined train/validation/test splits, 10 and 3 respectively, and we report the average over all splits. Details on each dataset and the corresponding evaluation metrics are provided in Table \ref{dataset_table}.

Additionally, we created two datasets that we call Country211 and Rendered SST2. The Country211 dataset is designed to assess the geolocation capability of visual representations. We filtered the YFCC100m dataset \citep{thomee2016yfcc100m} to find 211 countries (defined as having an ISO-3166 country code) that have at least 300 photos with GPS coordinates, and we built a balanced dataset with 211 categories, by sampling 200 photos for training and 100 photos for testing, for each country.

\begin{figure*}[t]
    \centering
    \fbox{\includegraphics[width=0.4\textwidth]{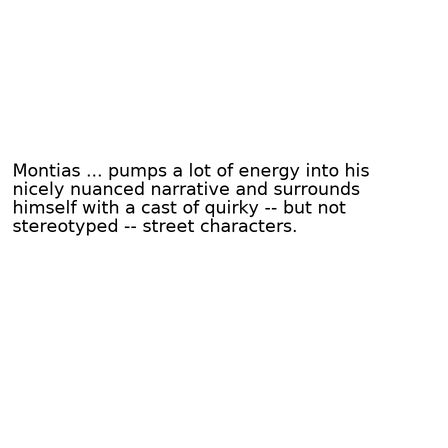}}
    \qquad
    \fbox{\includegraphics[width=0.4\textwidth]{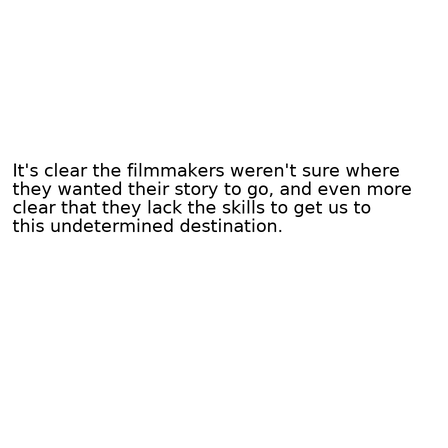}}
    \caption{Two example images from the Rendered SST2 dataset}
    \label{rendered-sst2-fig}
\end{figure*}

The Rendered SST2 dataset is designed to measure the optical character recognition capability of visual representations. To do so, we used the sentences from the Stanford Sentiment Treebank dataset \citep{socher2013recursive} and rendered them into images, with black texts on a white background, in a 448$\times$448 resolution. Two example images from this dataset are shown in Figure \ref{rendered-sst2-fig}.

\begin{table*}[t]
\centering
\begin{tabular}{lrrrr}
\toprule
    Dataset & Classes & Train size & Test size & Evaluation metric \\
\midrule
Food-101 & 102 & 75,750 & 25,250 & accuracy \\
CIFAR-10 & 10 & 50,000 & 10,000 & accuracy \\
CIFAR-100 & 100 & 50,000 & 10,000 & accuracy \\
Birdsnap & 500 & 42,283 & 2,149 & accuracy \\
SUN397 & 397 & 19,850 & 19,850 & accuracy \\
Stanford Cars & 196 & 8,144 & 8,041 & accuracy \\
FGVC Aircraft & 100 & 6,667 & 3,333 & mean per class \\
Pascal VOC 2007 Classification & 20 & 5,011 & 4,952 & 11-point mAP \\
Describable Textures & 47 & 3,760 & 1,880 & accuracy \\
Oxford-IIIT Pets & 37 & 3,680 & 3,669 & mean per class \\
Caltech-101 & 102 & 3,060 & 6,085 & mean-per-class \\
Oxford Flowers 102 & 102 & 2,040 & 6,149 & mean per class \\
\midrule
MNIST & 10 & 60,000 & 10,000 & accuracy \\
Facial Emotion Recognition 2013 & 8 & 32,140 & 3,574 & accuracy \\
STL-10 & 10 & 1000 & 8000 & accuracy \\
EuroSAT & 10 & 10,000 & 5,000 & accuracy \\
RESISC45 & 45 & 3,150 & 25,200 & accuracy \\
GTSRB & 43 & 26,640 & 12,630 & accuracy \\
KITTI & 4 & 6,770 & 711 & accuracy \\
Country211 & 211 & 43,200 & 21,100 & accuracy \\
PatchCamelyon & 2 & 294,912 & 32,768 & accuracy \\
UCF101 & 101 & 9,537 & 1,794 & accuracy \\
Kinetics700 & 700 & 494,801 & 31,669 & mean(top1, top5) \\
CLEVR Counts & 8 & 2,000 & 500 & accuracy \\
Hateful Memes & 2 & 8,500 & 500 & ROC AUC \\
Rendered SST2 & 2 & 7,792 & 1,821 & accuracy \\
ImageNet & 1000 & 1,281,167 & 50,000 & accuracy \\
\bottomrule
\end{tabular}
\caption{Datasets examined for linear probes. We note that, for the Birdsnap and Kinetics700 datasets, we used the resources that are available online at the time of this writing.}
\label{dataset_table}
\end{table*}

\subsection{Models}

In combination with the datasets listed above, we evaluate the following series of models using linear probes.

\paragraph{LM RN50} This is a multimodal model that uses an autoregressive loss instead of a contrastive loss, while using the ResNet-50 architecture as in the smallest contrastive model. To do so, the output from the CNN is projected into four tokens, which are then fed as a prefix to a language model autoregressively predicting the text tokens. Apart from the training objective, the model was trained on the same dataset for the same number of epochs as other CLIP models.
\paragraph{CLIP-RN} Five ResNet-based contrastive CLIP models are included. As discussed in the paper, the first two models follow ResNet-50 and ResNet-101, and we use EfficientNet-style \cite{tan2019efficientnet} scaling for the next three models which simultaneously scale the model width, the number of layers, and the input resolution to obtain models with roughly 4x, 16x, and 64x computation.
\paragraph{CLIP-ViT} We include four CLIP models that use the Vision Transformer \cite{dosovitskiy2020image} architecture as the image encoder. We include three models trained on 224-by-224 pixel images: ViT-B/32, ViT-B/16, ViT-L/14, and the ViT-L/14 model fine-tuned on 336-by-336 pixel input images.
\paragraph{EfficietNet} We use the nine models (B0-B8) from the original EfficientNet paper \cite{tan2019efficientnet}, as well as the noisy-student variants (B0-B7, L2-475, and L2-800) \citep{tan2019efficientnet}. The largest models (L2-475 and L2-800) take the input resolutions of 475x475 and 800x800 pixels, respectively.
\paragraph{Instagram-pretrained ResNeXt} We use the four models (32x8d, 32x16d, 32x32d, 32x48d) released by \cite{mahajan2018exploring}, as well as their two FixRes variants which use higher input resolutions \cite{touvron2019fixing}.
\paragraph{Big Transfer (BiT)} We use BiT-S and BiT-M models \citep{kolesnikov2019large}, trained on the ImageNet-1k and ImageNet-21k datasets. The model weights for BiT-L is not publicly available.
\paragraph{Vision Transformer (ViT)} We also include four ViT \cite{dosovitskiy2020image} checkpoints pretrained on the ImageNet-21k dataset, namely ViT-B/32, ViT-B/16, ViT-L/16, and ViT-H/14. We note that their best-performing models, trained on the JFT-300M dataset, are not available publicly.
\paragraph{SimCLRv2} The SimCLRv2 \citep{chen2020big} project released pre-trained and fine-tuned models in various settings. We use the seven pretrain-only checkpoints with selective kernels.
\paragraph{BYOL} We use the recently released model weights of BYOL \citep{grill2020byol}, specifically their 50x1 and 200x2 checkpoints.
\paragraph{Momentum Contrast (MoCo)} We include the MoCo-v1 \citep{he2020moco} and the MoCo-v2 \cite{chen2020mocov2} checkpoints.
\paragraph{VirTex} We use the pretrained model of VirTex \citep{desai2020virtex}. We note that VirTex has a similar model design to CLIP-AR but is trained on a 1000x smaller dataset of high-quality captions from MSCOCO.
\paragraph{ResNet} We add the original ResNet checkpoints released by \cite{he2016resnet}, namely ResNet-50, ResNet-101, and ResNet152.

\subsection{Evaluation}

We use image features taken from the penultimate layer of each model, ignoring any classification layer provided.
For CLIP-ViT models, we used the features before the linear projection to the embedding space, which corresponds to \texttt{I\_f} in Figure \ref{pseudocode}.
We train a logistic regression classifier using scikit-learn's L-BFGS implementation, with maximum 1,000 iterations, and report the corresponding metric for each dataset. We determine the L2 regularization strength $\lambda$ using a hyperparameter sweep on the validation sets over the range between $10^{-6}$ and $10^6$, with 96 logarithmically spaced steps. To save compute required for the sweeps, we perform a parametric binary search that starts with $\lambda=[10^{-6}, 10^{-4}, 10^{-2}, 1, 10^2, 10^4, 10^6]$ and iteratively halves the interval around the peak until it reaches a resolution of 8 steps per decade. The hyperparameter sweeps are performed on a validation split of each dataset. For the datasets that contain a validation split in addition to a test split, we use the provided validation set to perform the hyperparameter search, and for the datasets that do not provide a validation split or have not published labels for the test data, we split the training dataset to perform the hyperparameter search. For the final result, we combine the validation split back with the training split and report the performance on the unused split.

\subsection{Results}

\begin{figure*}[t]
\begin{center}
\centerline{\includegraphics[width=0.95\textwidth]{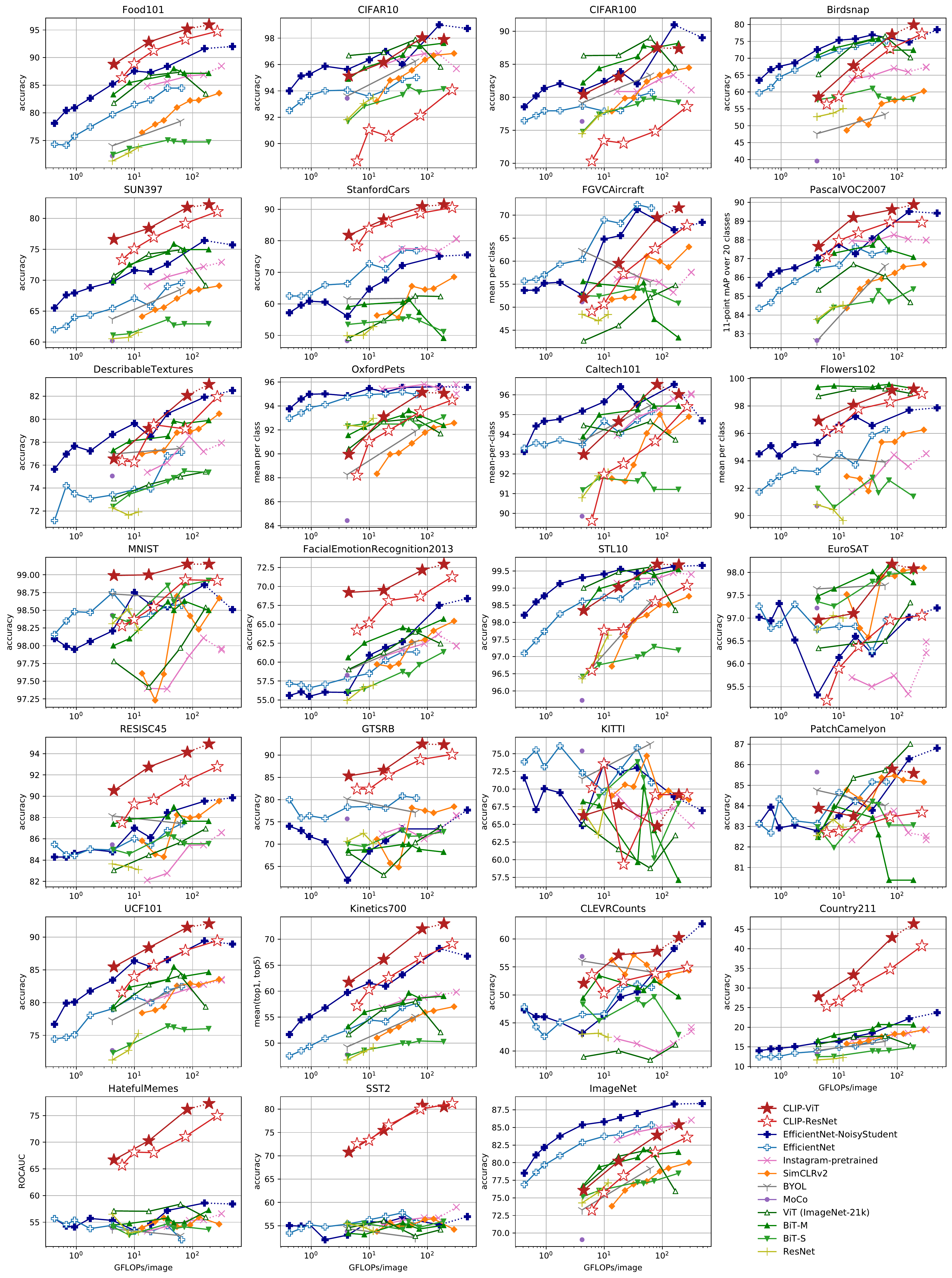}}
\caption{Linear probe performance plotted for each of the 27 datasets, using the data from Table \ref{tab:linear-probe-big-table}.}
\label{linear-probe-per-dataset}
\end{center}
\end{figure*}

The individual linear probe scores are provided in Table \ref{tab:linear-probe-big-table} and plotted in Figure \ref{linear-probe-per-dataset}. The best-performing CLIP model, using ViT-L/14 archiecture and 336-by-336 pixel images, achieved the state of the art in 21 of the 27 datasets, i.e. included in the Clopper-Pearson 99.5\% confidence interval around each dataset's top score.
For many datasets, CLIP performs significantly better than other models, demonstrating the advantage of natural language supervision over traditional pre-training approaches based on image classification.
See Section \ref{linear_probe_section} for more discussions on the linear probe results.

\begin{table*}[]
    \vspace{-0.5em}
    \linespread{1}
    \aboverulesep = 0.2em \belowrulesep = 0.2em
    \scriptsize
    \centering
\begin{tabular}{cc|cccccccccccccccccccccccccccc} 
        && \hspace{-1.2em} & \rotatebox[origin=lb]{90}{\smash{Food101}} & \rotatebox[origin=lb]{90}{\smash{CIFAR10}} & \rotatebox[origin=lb]{90}{\smash{CIFAR100}} & \rotatebox[origin=lb]{90}{\smash{Birdsnap}} & \rotatebox[origin=lb]{90}{\smash{SUN397}} & \rotatebox[origin=lb]{90}{\smash{Cars}} & \rotatebox[origin=lb]{90}{\smash{Aircraft}} & \rotatebox[origin=lb]{90}{\smash{VOC2007}} & \rotatebox[origin=lb]{90}{\smash{DTD}} & \rotatebox[origin=lb]{90}{\smash{Pets}} & \rotatebox[origin=lb]{90}{\smash{Caltech101}} & \rotatebox[origin=lb]{90}{\smash{Flowers}} & \rotatebox[origin=lb]{90}{\smash{MNIST}} & \rotatebox[origin=lb]{90}{\smash{FER2013}} & \rotatebox[origin=lb]{90}{\smash{STL10$^\star$}} & \rotatebox[origin=lb]{90}{\smash{EuroSAT}} & \rotatebox[origin=lb]{90}{\smash{RESISC45}} & \rotatebox[origin=lb]{90}{\smash{GTSRB}} & \rotatebox[origin=lb]{90}{\smash{KITTI}} & \rotatebox[origin=lb]{90}{\smash{Country211}} & \rotatebox[origin=lb]{90}{\smash{PCAM}} & \rotatebox[origin=lb]{90}{\smash{UCF101}} & \rotatebox[origin=lb]{90}{\smash{Kinetics700}} & \rotatebox[origin=lb]{90}{\smash{CLEVR}} & \rotatebox[origin=lb]{90}{\smash{HatefulMemes}} & \rotatebox[origin=lb]{90}{\smash{SST}} & \rotatebox[origin=lb]{90}{\smash{ImageNet}} 
        \\
\bottomrule
    \end{tabular}
    \caption{Linear probe performance of various pre-trained models over 27 datasets. Scores within the 99.5\% Clopper-Pearson confidence interval of each dataset's top score are shown in bold.\newline \mbox{}\hfill{\scriptsize $^\star$We updated the STL10 scores from the previous version of this paper after fixing a CUDA-related bug.}}
    \label{tab:linear-probe-big-table}
\end{table*}

\begin{figure*}[t]
\begin{center}
\centerline{\includegraphics[width=0.95\textwidth]{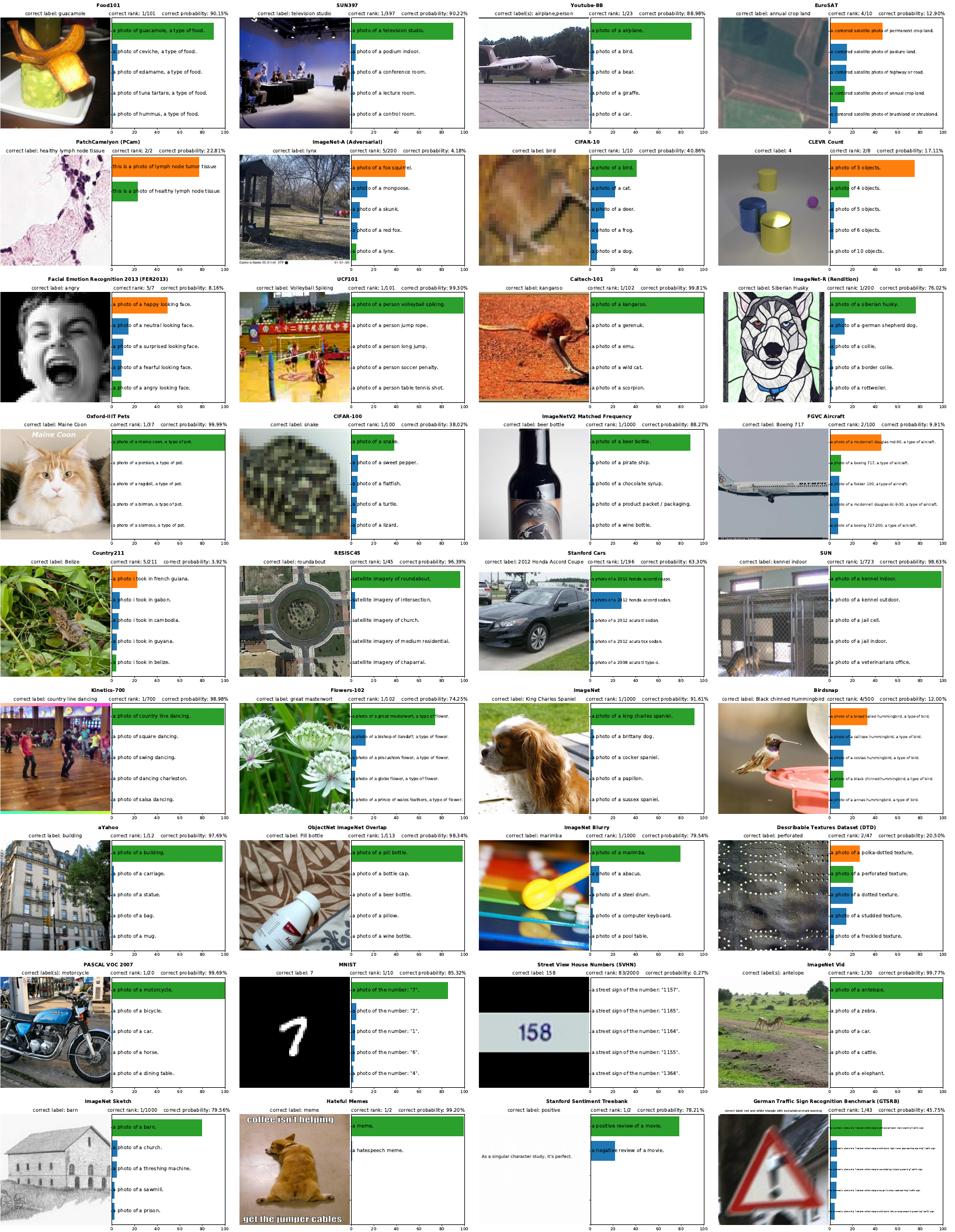}}
\caption{Visualization of predictions from 36 CLIP zero-shot classifiers. All examples are random with the exception of reselecting Hateful Memes to avoid offensive content. The predicted probability of the top 5 classes is shown along with the text used to represent the class. When more than one template is used, the first template is shown. The ground truth label is colored green while an incorrect prediction is colored orange.}
\label{zero_shot_prediction_fig}
\end{center}
\end{figure*}

\begin{table*}[]

    \vspace{-0.5em}
    \linespread{1}
    \aboverulesep = 0.2em \belowrulesep = 0.2em
    \scriptsize
    \centering
    \begin{tabular}{cc|cccccccccccccccccccccccccccc} 
        && \hspace{-1.2em} & \rotatebox[origin=lb]{90}{\smash{Food101}} & \rotatebox[origin=lb]{90}{\smash{CIFAR10}} & \rotatebox[origin=lb]{90}{\smash{CIFAR100}} & \rotatebox[origin=lb]{90}{\smash{Birdsnap}} & \rotatebox[origin=lb]{90}{\smash{SUN397}} & \rotatebox[origin=lb]{90}{\smash{Stanford Cars}} & \rotatebox[origin=lb]{90}{\smash{FGVC Aircraft}} & \rotatebox[origin=lb]{90}{\smash{VOC2007}} & \rotatebox[origin=lb]{90}{\smash{DTD}} & \rotatebox[origin=lb]{90}{\smash{Oxford Pets}} & \rotatebox[origin=lb]{90}{\smash{Caltech101}} & \rotatebox[origin=lb]{90}{\smash{Flowers102}} & \rotatebox[origin=lb]{90}{\smash{MNIST}} & \rotatebox[origin=lb]{90}{\smash{FER2013}} & \rotatebox[origin=lb]{90}{\smash{STL10}} & \rotatebox[origin=lb]{90}{\smash{EuroSAT}} & \rotatebox[origin=lb]{90}{\smash{RESISC45}} & \rotatebox[origin=lb]{90}{\smash{GTSRB}} & \rotatebox[origin=lb]{90}{\smash{KITTI}} & \rotatebox[origin=lb]{90}{\smash{Country211}} & \rotatebox[origin=lb]{90}{\smash{PCam}} & \rotatebox[origin=lb]{90}{\smash{UCF101}} & \rotatebox[origin=lb]{90}{\smash{Kinetics700}} & \rotatebox[origin=lb]{90}{\smash{CLEVR}} & \rotatebox[origin=lb]{90}{\smash{HatefulMemes}} & \rotatebox[origin=lb]{90}{\smash{Rendered SST2}} & \rotatebox[origin=lb]{90}{\smash{ImageNet}} 
        \\
        \midrule
        \multirow{5}{0em}{\rotatebox[origin=c]{90}{CLIP-ResNet}}
        &RN50& \hspace{-1.2em} & \hspace{-0.9em}81.1\hspace{-0.4em} & \hspace{-0.9em}75.6\hspace{-0.4em} & \hspace{-0.9em}41.6\hspace{-0.4em} & \hspace{-0.9em}32.6\hspace{-0.4em} & \hspace{-0.9em}59.6\hspace{-0.4em} & \hspace{-0.9em}55.8\hspace{-0.4em} & \hspace{-0.9em}19.3\hspace{-0.4em} & \hspace{-0.9em}82.1\hspace{-0.4em} & \hspace{-0.9em}41.7\hspace{-0.4em} & \hspace{-0.9em}85.4\hspace{-0.4em} & \hspace{-0.9em}82.1\hspace{-0.4em} & \hspace{-0.9em}65.9\hspace{-0.4em} & \hspace{-0.9em}66.6\hspace{-0.4em} & \hspace{-0.9em}42.2\hspace{-0.4em} & \hspace{-0.9em}94.3\hspace{-0.4em} & \hspace{-0.9em}41.1\hspace{-0.4em} & \hspace{-0.9em}54.2\hspace{-0.4em} & \hspace{-0.9em}35.2\hspace{-0.4em} & \hspace{-0.9em}42.2\hspace{-0.4em} & \hspace{-0.9em}16.1\hspace{-0.4em} & \hspace{-0.9em}57.6\hspace{-0.4em} & \hspace{-0.9em}63.6\hspace{-0.4em} & \hspace{-0.9em}43.5\hspace{-0.4em} & \hspace{-0.9em}20.3\hspace{-0.4em} & \hspace{-0.9em}59.7\hspace{-0.4em} & \hspace{-0.9em}56.9\hspace{-0.4em} & \hspace{-0.9em}59.6\hspace{-0.4em}
        \\
        &RN101& \hspace{-1.2em} & \hspace{-0.9em}83.9\hspace{-0.4em} & \hspace{-0.9em}81.0\hspace{-0.4em} & \hspace{-0.9em}49.0\hspace{-0.4em} & \hspace{-0.9em}37.2\hspace{-0.4em} & \hspace{-0.9em}59.9\hspace{-0.4em} & \hspace{-0.9em}62.3\hspace{-0.4em} & \hspace{-0.9em}19.5\hspace{-0.4em} & \hspace{-0.9em}82.4\hspace{-0.4em} & \hspace{-0.9em}43.9\hspace{-0.4em} & \hspace{-0.9em}86.2\hspace{-0.4em} & \hspace{-0.9em}85.1\hspace{-0.4em} & \hspace{-0.9em}65.7\hspace{-0.4em} & \hspace{-0.9em}59.3\hspace{-0.4em} & \hspace{-0.9em}45.6\hspace{-0.4em} & \hspace{-0.9em}96.7\hspace{-0.4em} & \hspace{-0.9em}33.1\hspace{-0.4em} & \hspace{-0.9em}58.5\hspace{-0.4em} & \hspace{-0.9em}38.3\hspace{-0.4em} & \hspace{-0.9em}33.3\hspace{-0.4em} & \hspace{-0.9em}16.9\hspace{-0.4em} & \hspace{-0.9em}55.2\hspace{-0.4em} & \hspace{-0.9em}62.2\hspace{-0.4em} & \hspace{-0.9em}46.7\hspace{-0.4em} & \hspace{-0.9em}28.1\hspace{-0.4em} & \hspace{-0.9em}61.1\hspace{-0.4em} & \hspace{-0.9em}64.2\hspace{-0.4em} & \hspace{-0.9em}62.2\hspace{-0.4em} 
        \\
        &RN50x4& \hspace{-1.2em} & \hspace{-0.9em}86.8\hspace{-0.4em} & \hspace{-0.9em}79.2\hspace{-0.4em} & \hspace{-0.9em}48.9\hspace{-0.4em} & \hspace{-0.9em}41.6\hspace{-0.4em} & \hspace{-0.9em}62.7\hspace{-0.4em} & \hspace{-0.9em}67.9\hspace{-0.4em} & \hspace{-0.9em}24.6\hspace{-0.4em} & \hspace{-0.9em}83.0\hspace{-0.4em} & \hspace{-0.9em}49.3\hspace{-0.4em} & \hspace{-0.9em}88.1\hspace{-0.4em} & \hspace{-0.9em}86.0\hspace{-0.4em} & \hspace{-0.9em}68.0\hspace{-0.4em} & \hspace{-0.9em}75.2\hspace{-0.4em} & \hspace{-0.9em}51.1\hspace{-0.4em} & \hspace{-0.9em}96.4\hspace{-0.4em} & \hspace{-0.9em}35.0\hspace{-0.4em} & \hspace{-0.9em}59.2\hspace{-0.4em} & \hspace{-0.9em}35.7\hspace{-0.4em} & \hspace{-0.9em}26.0\hspace{-0.4em} & \hspace{-0.9em}20.2\hspace{-0.4em} & \hspace{-0.9em}57.5\hspace{-0.4em} & \hspace{-0.9em}65.5\hspace{-0.4em} & \hspace{-0.9em}49.0\hspace{-0.4em} & \hspace{-0.9em}17.0\hspace{-0.4em} & \hspace{-0.9em}58.3\hspace{-0.4em} & \hspace{-0.9em}66.6\hspace{-0.4em} & \hspace{-0.9em}65.8\hspace{-0.4em}
        \\
        &RN50x16& \hspace{-1.2em} & \hspace{-0.9em}90.5\hspace{-0.4em} & \hspace{-0.9em}82.2\hspace{-0.4em} & \hspace{-0.9em}54.2\hspace{-0.4em} & \hspace{-0.9em}45.9\hspace{-0.4em} & \hspace{-0.9em}65.0\hspace{-0.4em} & \hspace{-0.9em}72.3\hspace{-0.4em} & \hspace{-0.9em}30.3\hspace{-0.4em} & \hspace{-0.9em}82.9\hspace{-0.4em} & \hspace{-0.9em}52.8\hspace{-0.4em} & \hspace{-0.9em}89.7\hspace{-0.4em} & \hspace{-0.9em}87.6\hspace{-0.4em} & \hspace{-0.9em}71.9\hspace{-0.4em} & \hspace{-0.9em}80.0\hspace{-0.4em} & \hspace{-0.9em}56.0\hspace{-0.4em} & \hspace{-0.9em}97.8\hspace{-0.4em} & \hspace{-0.9em}40.3\hspace{-0.4em} & \hspace{-0.9em}64.4\hspace{-0.4em} & \hspace{-0.9em}39.6\hspace{-0.4em} & \hspace{-0.9em}33.9\hspace{-0.4em} & \hspace{-0.9em}24.0\hspace{-0.4em} & \hspace{-0.9em}62.5\hspace{-0.4em} & \hspace{-0.9em}68.7\hspace{-0.4em} & \hspace{-0.9em}53.4\hspace{-0.4em} & \hspace{-0.9em}17.6\hspace{-0.4em} & \hspace{-0.9em}58.9\hspace{-0.4em} & \hspace{-0.9em}67.6\hspace{-0.4em} & \hspace{-0.9em}70.5\hspace{-0.4em}
        \\
        &RN50x64& \hspace{-1.2em} & \hspace{-0.9em}91.8\hspace{-0.4em} & \hspace{-0.9em}86.8\hspace{-0.4em} & \hspace{-0.9em}61.3\hspace{-0.4em} & \hspace{-0.9em}48.9\hspace{-0.4em} & \hspace{-0.9em}66.9\hspace{-0.4em} & \hspace{-0.9em}76.0\hspace{-0.4em} & \hspace{-0.9em}35.6\hspace{-0.4em} & \hspace{-0.9em}83.8\hspace{-0.4em} & \hspace{-0.9em}53.4\hspace{-0.4em} & \hspace{-0.9em}93.4\hspace{-0.4em} & \hspace{-0.9em}90.6\hspace{-0.4em} & \hspace{-0.9em}77.3\hspace{-0.4em} & \hspace{-0.9em}90.8\hspace{-0.4em} & \hspace{-0.9em}61.0\hspace{-0.4em} & \hspace{-0.9em}98.3\hspace{-0.4em} & \hspace{-0.9em}59.4\hspace{-0.4em} & \hspace{-0.9em}69.7\hspace{-0.4em} & \hspace{-0.9em}47.9\hspace{-0.4em} & \hspace{-0.9em}33.2\hspace{-0.4em} & \hspace{-0.9em}29.6\hspace{-0.4em} & \hspace{-0.9em}65.0\hspace{-0.4em} & \hspace{-0.9em}74.1\hspace{-0.4em} & \hspace{-0.9em}56.8\hspace{-0.4em} & \hspace{-0.9em}27.5\hspace{-0.4em} & \hspace{-0.9em}62.1\hspace{-0.4em} & \hspace{-0.9em}70.7\hspace{-0.4em} & \hspace{-0.9em}73.6\hspace{-0.4em}
        \\
        \midrule
        \multirow{4}{0em}{\rotatebox[origin=c]{90}{CLIP-ViT}}
        &B/32& \hspace{-1.2em} & \hspace{-0.9em}84.4\hspace{-0.4em} & \hspace{-0.9em}91.3\hspace{-0.4em} & \hspace{-0.9em}65.1\hspace{-0.4em} & \hspace{-0.9em}37.8\hspace{-0.4em} & \hspace{-0.9em}63.2\hspace{-0.4em} & \hspace{-0.9em}59.4\hspace{-0.4em} & \hspace{-0.9em}21.2\hspace{-0.4em} & \hspace{-0.9em}83.1\hspace{-0.4em} & \hspace{-0.9em}44.5\hspace{-0.4em} & \hspace{-0.9em}87.0\hspace{-0.4em} & \hspace{-0.9em}87.9\hspace{-0.4em} & \hspace{-0.9em}66.7\hspace{-0.4em} & \hspace{-0.9em}51.9\hspace{-0.4em} & \hspace{-0.9em}47.3\hspace{-0.4em} & \hspace{-0.9em}97.2\hspace{-0.4em} & \hspace{-0.9em}49.4\hspace{-0.4em} & \hspace{-0.9em}60.3\hspace{-0.4em} & \hspace{-0.9em}32.2\hspace{-0.4em} & \hspace{-0.9em}39.4\hspace{-0.4em} & \hspace{-0.9em}17.8\hspace{-0.4em} & \hspace{-0.9em}58.4\hspace{-0.4em} & \hspace{-0.9em}64.5\hspace{-0.4em} & \hspace{-0.9em}47.8\hspace{-0.4em} & \hspace{-0.9em}24.8\hspace{-0.4em} & \hspace{-0.9em}57.6\hspace{-0.4em} & \hspace{-0.9em}59.6\hspace{-0.4em} & \hspace{-0.9em}63.2\hspace{-0.4em}  
        \\
        &B/16& \hspace{-1.2em} & \hspace{-0.9em}89.2\hspace{-0.4em} & \hspace{-0.9em}91.6\hspace{-0.4em} & \hspace{-0.9em}68.7\hspace{-0.4em} & \hspace{-0.9em}39.1\hspace{-0.4em} & \hspace{-0.9em}65.2\hspace{-0.4em} & \hspace{-0.9em}65.6\hspace{-0.4em} & \hspace{-0.9em}27.1\hspace{-0.4em} & \hspace{-0.9em}83.9\hspace{-0.4em} & \hspace{-0.9em}46.0\hspace{-0.4em} & \hspace{-0.9em}88.9\hspace{-0.4em} & \hspace{-0.9em}89.3\hspace{-0.4em} & \hspace{-0.9em}70.4\hspace{-0.4em} & \hspace{-0.9em}56.0\hspace{-0.4em} & \hspace{-0.9em}52.7\hspace{-0.4em} & \hspace{-0.9em}98.2\hspace{-0.4em} & \hspace{-0.9em}54.1\hspace{-0.4em} & \hspace{-0.9em}65.5\hspace{-0.4em} & \hspace{-0.9em}43.3\hspace{-0.4em} & \hspace{-0.9em}44.0\hspace{-0.4em} & \hspace{-0.9em}23.3\hspace{-0.4em} & \hspace{-0.9em}48.1\hspace{-0.4em} & \hspace{-0.9em}69.8\hspace{-0.4em} & \hspace{-0.9em}52.4\hspace{-0.4em} & \hspace{-0.9em}23.4\hspace{-0.4em} & \hspace{-0.9em}61.7\hspace{-0.4em} & \hspace{-0.9em}59.8\hspace{-0.4em} & \hspace{-0.9em}68.6\hspace{-0.4em} 
        \\
        &L/14& \hspace{-1.2em} & \hspace{-0.9em}92.9\hspace{-0.4em} & \hspace{-0.9em}96.2\hspace{-0.4em} & \hspace{-0.9em}77.9\hspace{-0.4em} & \hspace{-0.9em}48.3\hspace{-0.4em} & \hspace{-0.9em}67.7\hspace{-0.4em} & \hspace{-0.9em}77.3\hspace{-0.4em} & \hspace{-0.9em}36.1\hspace{-0.4em} & \hspace{-0.9em}84.1\hspace{-0.4em} & \hspace{-0.9em}55.3\hspace{-0.4em} & \hspace{-0.9em}93.5\hspace{-0.4em} & \hspace{-0.9em}92.6\hspace{-0.4em} & \hspace{-0.9em}78.7\hspace{-0.4em} & \hspace{-0.9em}87.2\hspace{-0.4em} & \hspace{-0.9em}57.5\hspace{-0.4em} & \hspace{-0.9em}99.3\hspace{-0.4em} & \hspace{-0.9em}59.9\hspace{-0.4em} & \hspace{-0.9em}71.6\hspace{-0.4em} & \hspace{-0.9em}50.3\hspace{-0.4em} & \hspace{-0.9em}23.1\hspace{-0.4em} & \hspace{-0.9em}32.7\hspace{-0.4em} & \hspace{-0.9em}58.8\hspace{-0.4em} & \hspace{-0.9em}76.2\hspace{-0.4em} & \hspace{-0.9em}60.3\hspace{-0.4em} & \hspace{-0.9em}24.3\hspace{-0.4em} & \hspace{-0.9em}63.3\hspace{-0.4em} & \hspace{-0.9em}64.0\hspace{-0.4em} & \hspace{-0.9em}75.3\hspace{-0.4em}
        \\
        &L/14-336px& \hspace{-1.2em} & \hspace{-0.9em}93.8\hspace{-0.4em} & \hspace{-0.9em}95.7\hspace{-0.4em} & \hspace{-0.9em}77.5\hspace{-0.4em} & \hspace{-0.9em}49.5\hspace{-0.4em} & \hspace{-0.9em}68.4\hspace{-0.4em} & \hspace{-0.9em}78.8\hspace{-0.4em} & \hspace{-0.9em}37.2\hspace{-0.4em} & \hspace{-0.9em}84.3\hspace{-0.4em} & \hspace{-0.9em}55.7\hspace{-0.4em} & \hspace{-0.9em}93.5\hspace{-0.4em} & \hspace{-0.9em}92.8\hspace{-0.4em} & \hspace{-0.9em}78.3\hspace{-0.4em} & \hspace{-0.9em}88.3\hspace{-0.4em} & \hspace{-0.9em}57.7\hspace{-0.4em} & \hspace{-0.9em}99.4\hspace{-0.4em} & \hspace{-0.9em}59.6\hspace{-0.4em} & \hspace{-0.9em}71.7\hspace{-0.4em} & \hspace{-0.9em}52.3\hspace{-0.4em} & \hspace{-0.9em}21.9\hspace{-0.4em} & \hspace{-0.9em}34.9\hspace{-0.4em} & \hspace{-0.9em}63.0\hspace{-0.4em} & \hspace{-0.9em}76.9\hspace{-0.4em} & \hspace{-0.9em}61.3\hspace{-0.4em} & \hspace{-0.9em}24.8\hspace{-0.4em} & \hspace{-0.9em}63.3\hspace{-0.4em} & \hspace{-0.9em}67.9\hspace{-0.4em} & \hspace{-0.9em}76.2\hspace{-0.4em} 
        \\
        \bottomrule
    \end{tabular}
    \caption{Zero-shot performance of CLIP models over 27 datasets.}
    \label{tab:zero-shot-big-table}

\label{all-zero-shot-performance-table}
\vspace{2em}
\includegraphics[width=\textwidth]{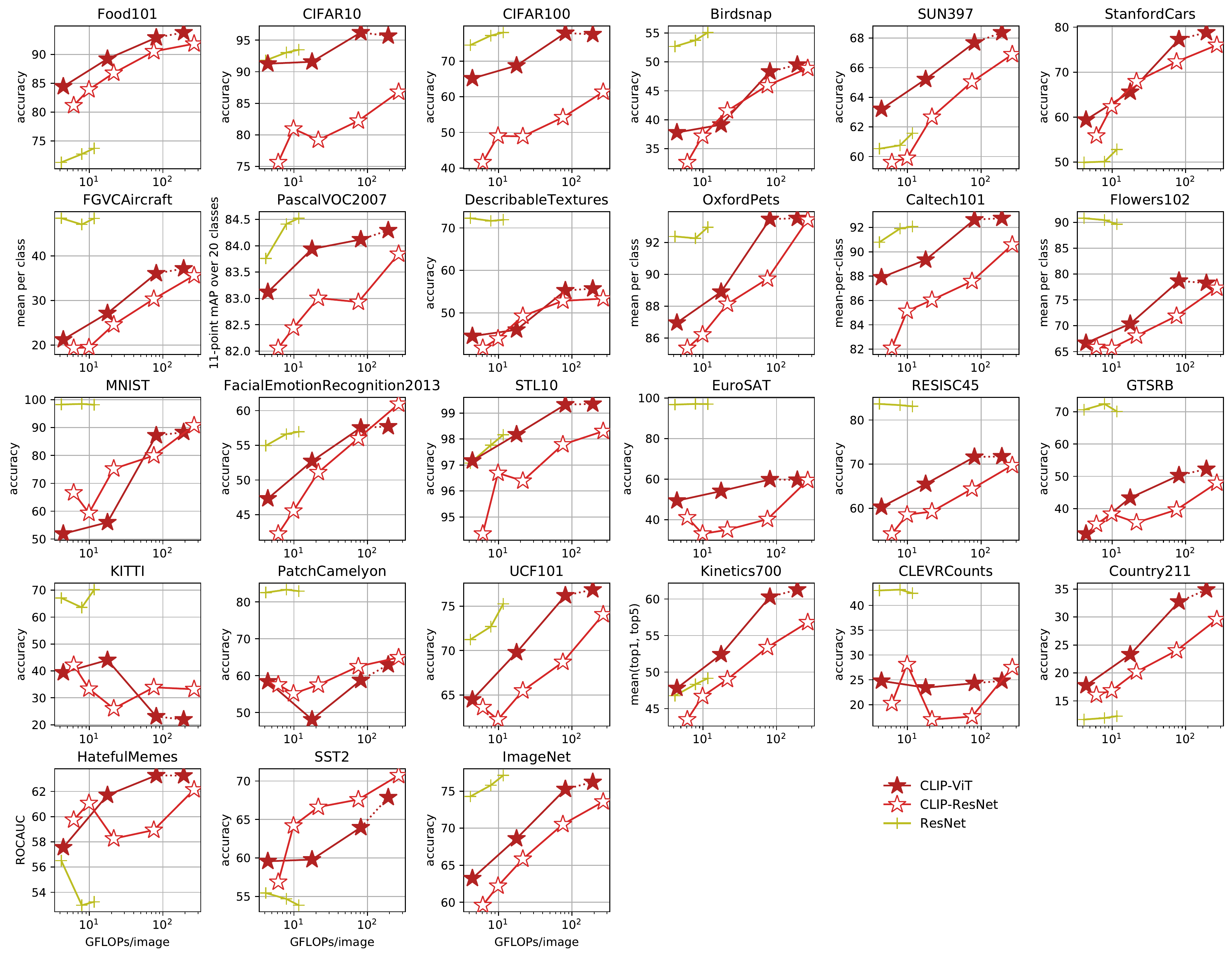}
\captionof{figure}{CLIP's zero-shot performance compared to linear-probe ResNet performance} %
\label{all-zero-shot-performance-figure}
\end{table*}

\clearpage

\section{Zero-Shot Prediction}

To provide a qualitative summary / overview of CLIP's zero-shot performance we visualize a randomly selected prediction for 36 different zero-shot CLIP classifiers in Figure \ref{zero_shot_prediction_fig}. 
In addition, Table \ref{all-zero-shot-performance-table} and Figure \ref{all-zero-shot-performance-figure} show the individual zero-shot performance scores for each dataset.

\section{Duplicate Detector}
\label{dupdet}

Our early attempts at duplicate detection and analysis used nearest neighbors in the model's learned embedding space. While it is intuitive to use a model's own notion of similarity, we encountered issues. We found the model's feature space is weighted very heavily towards semantic similarity. Many false positives occurred due to distinct objects that would be described similarly (soccer balls, flowers of the same species, etc...) having almost perfect similarity. We also observed the model was quite poor at assigning certain kinds of near-duplicates high similarity scores. We noticed repeatedly that images with high-frequency textures (such as fur or stripe patterns) pre-processed by different resizing algorithms (nearest neighbor vs bi-linear) could have surprisingly low similarity. This resulted in many false negatives. 

We built our own near-duplicate detector to fix this issue. We created a synthetic data augmentation pipeline that combined a variety of common image manipulations. The augmentation pipeline combines random cropping and zooming, aspect ratio distortion, downsizing and upscaling to different resolutions, minor rotations, jpeg compression, and HSV color jitter. The pipeline also randomly selects from different interpolation algorithms for all relevant steps. We then trained a model to maximize the similarity of an image and its transformed variant while minimizing similarity to all other images in a training batch. We used the same n-pair / InfoNCE loss as CLIP but with a fixed temperature of 0.07.

We selected a ResNet-50 as the model architecture. We modified the base ResNet-50 with the anti-alias improvements from \cite{zhang2019making} and used weight norm \cite{salimans2016weight} instead of batch norm \citep{ioffe2015batch} to avoid leaking information about duplicates via batch statistics - a problem previously noted in \cite{henaff2020data}. We also found the GELU activation function \citep{hendrycks2016gaussian} to perform better for this task. We trained the model with a total batch size of 1,712 for approximately 30 million images sampled from our pre-training dataset. At the end of training it achieves nearly 100\% accuracy on its proxy training task.

\section{Dataset Ablation on YFCC100M}

\begin{table}[t]
\vskip 0.15in
\scriptsize
\begin{center}
\begin{tabular}{l|rrr|rrr}
\toprule
& \multicolumn{3}{c}{\textit{Linear Classifier}} & \multicolumn{3}{c}{\textit{Zero Shot}} \\
Dataset & YFCC & WIT & $\Delta$ & YFCC & WIT & $\Delta$ \\
\midrule
Birdsnap      & 47.4 & 35.3 & $+$12.1 & 19.9 &  4.5 & $+$15.4 \\
Country211    & 23.1 & 17.3 &  $+$5.8 &  5.2 &  5.3 &  $+$0.1 \\
Flowers102    & 94.4 & 89.8 &  $+$4.6 & 48.6 & 21.7 & $+$26.9 \\
GTSRB         & 66.8 & 72.5 &  $-$5.7 &  6.9 &  7.0 &  $-$0.1 \\
UCF101        & 69.2 & 74.9 &  $-$5.7 & 22.9 & 32.0 &  $-$9.1 \\
Stanford Cars & 31.4 & 50.3 & $-$18.9 &  3.8 & 10.9 &  $-$7.1 \\
\midrule
ImageNet & \textbf{62.0} & 60.8 & $+1.2$ & \textbf{31.3} & 27.6 & $+$3.7 \\
Dataset Average  & 65.5 & \textbf{66.6} & $-$1.1 & 29.6 & \textbf{30.0} & $-$0.4 \\
Dataset ``Wins'' & 10 & \textbf{15} & $-$5 & \textbf{19} & 18 & $+$1 \\
\bottomrule
\end{tabular}
\caption{\textbf{CLIP performs similarly when trained on only YFCC100M.} Comparing a ResNet-50 trained on only YFCC100M with a same sized subset of WIT shows similar average performance and number of wins on zero shot and linear classifier evals. However, large differences in dataset specific performance occur. We include performance on the 3 datasets where YFCC does best and worst compared to WIT according to a linear probe in order to highlight this as well as aggregate performance across all linear and zero-shot evals and the canonical ImageNet dataset.}
\label{dataset_ablation_table}
\end{center}
\vskip -0.1in
\end{table}

To study whether our custom dataset is critical to the performance of CLIP, we trained a model on a filtered subset of the YFCC100M dataset (details described in Section \ref{subsection:creating-large-dataset}) and compared its performance to the same model trained on an equally sized subset of WIT. We train each model for 32 epochs at which point transfer performance begins to plateau due to overfitting. Results are shown in Table \ref{dataset_ablation_table}. Across our whole eval suite, YFCC and WIT perform similarly on average for both zero-shot and linear probe settings. However, performance on specific fine-grained classification datasets can vary widely - sometimes by over 10\%. Our speculation is that these differences in performance reflect the relative density of relevant data in each pre-training dataset. For instance, pre-training on YFCC100M, which might contain many photos of birds and flowers (common subjects for photographers), results in better performance on Birdsnap and Flowers102, while pre-training on WIT results in better car and pet classifiers (which appear common in our dataset).

Overall, these results are encouraging as they suggest our approach can use any reasonably filtered collection of paired (text, image) data. This mirrors recent work which reported positive results using the same contrastive pre-training objective on the relatively different domain of medical imaging \cite{zhang2020contrastive}. It also is similar to the findings of noisy student self-training which reported only slight improvements when using their JFT300M dataset over YFCC100M \cite{xie2020self}. We suspect the major advantage of our dataset over the already existing YFCC100M is its much larger size. 

Finally, we caution that WIT includes this filtered subset of YFCC100M. This could result in our ablation underestimating the size of performance differences between YFCC100M and the rest of WIT. We do not think this is likely as YFCC100M is only 3.7\% of the overall WIT data blend and it did not noticeably change the performance of models when it was added to the existing data blend during the creation of WIT.

\newcommand*\rot{\rotatebox{90}}
\newcommand{\xa}{\makebox[0pt][l]{$^a$}}
\newcommand{\xb}{\makebox[0pt][l]{$^b$}}
\newcommand{\xc}{\makebox[0pt][l]{$^c$}}
\newcommand{\xd}{\makebox[0pt][l]{$^d$}}
\newcommand{\xe}{\makebox[0pt][l]{$^e$}}
\newcommand{\xf}{\makebox[0pt][l]{$^f$}}
\newcommand{\xg}{\makebox[0pt][l]{$^g$}}
\newcommand{\xh}{\makebox[0pt][l]{$^h$}}
\newcommand{\xxi}{\makebox[0pt][l]{$^i$}}

\begin{table*}[t]
\vskip 0.15in
\small
\begin{center}
\begin{tabular}{llcccccccccccc}
\toprule
& & \multicolumn{6}{c}{Text Retrieval} & \multicolumn{6}{c}{Image Retrieval} \\
& & \multicolumn{3}{c}{Flickr30k} & \multicolumn{3}{c}{MSCOCO} & \multicolumn{3}{c}{Flickr30k} & \multicolumn{3}{c}{MSCOCO} \\
& & R@1 & R@5 & R@10 & R@1 & R@5 & R@10 & R@1 & R@5 & R@10 & R@1 & R@5 & R@10 \\
\midrule
\multirow{5}{*}{\rot{Finetune}} & Unicoder-VL\xa & 86.2 & 96.3 & 99.0 &  62.3 & 87.1 & 92.8 & 71.5 & 90.9 & 94.9 & 46.7 & 76.0 & 85.3 \\
& Uniter\xb & 87.3 & \underline{98.0} & \underline{99.2} & 65.7 & 88.6 & 93.8 & 75.6 & 94.1 & \underline{\textbf{96.8}} & 52.9 & 79.9 & 88.0 \\
& VILLA\xc & 87.9 & 97.5 & 98.8 & - & - & - & 76.3 & \underline{\textbf{94.2}} & \underline{\textbf{96.8}} & - & - & - \\
& Oscar\xd & - & - & - & \underline{\textbf{73.5}} & \underline{\textbf{92.2}} & \underline{\textbf{96.0}} & - & - & - & \underline{\textbf{57.5}} & \underline{\textbf{82.8}} & \underline{\textbf{89.8}} \\
& ERNIE-ViL\xe & \underline{\textbf{88.7}} & \underline{98.0} & \underline{99.2} & - & - & - & \underline{\textbf{76.7}} & 93.6 & 96.4 & - & - & - \\
\midrule
\multirow{5}{*}{\rot{Zero-Shot}} & Visual N-Grams\xf & 15.4 & 35.7 & 45.1 & 8.7 & 23.1 & 33.3 & 8.8 & 21.2 & 29.9 & 5.0 & 14.5 & 21.9 \\
& ImageBERT\xg & - & - & - & 44.0 & 71.2 & 80.4 & - & - & - & 32.3 & 59.0 & 70.2 \\
& Unicoder-VL\xa & 64.3 & 86.8 & 92.3 & - & - & - & 48.4 & 76.0 & 85.2 & - & - & - \\
& Uniter\xb & 83.6 & 95.7 & 97.7 & - & - & - & \underline{68.7} & 89.2 & 93.9 & - & - & - \\
& CLIP & \underline{88.0} & \underline{\textbf{98.7}} & \underline{\textbf{99.4}} & \underline{58.4} & \underline{81.5} & \underline{88.1} & \underline{68.7} & \underline{90.6} & \underline{95.2} & \underline{37.8} & \underline{62.4} & \underline{72.2} \\ %
\bottomrule
\end{tabular}
\caption{\textbf{CLIP improves zero-shot retrieval and is competitive with the best fine-tuned result on Flickr30k text retrieval.} Bold indicates best overall performance while an underline indicates best in category performance (zero-shot or fine-tuned). For all other models, best results from the paper are reported regardless of model size / variant. MSCOCO performance is reported on the 5k test set. $^a$\citep{li2020unicoder} $^b$\citep{chen2019uniter} $^c$\citep{gan2020large} $^d$\citep{li2020oscar} $^e$\citep{yu2020ernie} $^f$\citep{li2017learning} $^g$\citep{qi2020imagebert}}
\label{table:retrieval}
\end{center}
\vskip -0.1in
\end{table*}

\section{Selected Task and Dataset Results}
\label{appendix:selected}

Due to the large variety of datasets and experiments considered in this work, the main body focuses on summarizing and analyzing overall results. In the following subsections we report details of performance for specific groups of tasks, datasets, and evaluation settings.

\subsection{Image and Text Retrieval}

CLIP pre-trains for the task of image-text retrieval on our noisy web-scale dataset. Although the focus of this paper is on representation learning and task learning for the purpose of transfer to a wide variety of downstream datasets, validating that CLIP is able to achieve high transfer performance transfer on exactly what it is pre-trained for is an important sanity check / proof of concept. In Table \ref{table:retrieval} we check the zero-shot transfer performance of CLIP for both text and image retrieval on the Flickr30k and MSCOCO datsets. Zero-shot CLIP matches or outperforms all prior zero-shot results on these two datasets. Zero-shot CLIP is also competitive with the current overall SOTA for the task of text retrieval on Flickr30k. On image retrieval, CLIP's performance relative to the overall state of the art is noticeably lower. However, zero-shot CLIP is still competitive with a fine-tuned Unicoder-VL. On the larger MS-COCO dataset fine-tuning improves performance significantly and zero-shot CLIP is not competitive with the most recent work. For both these datasets we prepend the prompt ``\texttt{a photo of}'' to the description of each image which we found boosts CLIP's zero-shot R@1 performance between 1 and 2 points.

\begin{table}[ht]
\vskip 0.15in
\begin{center}
\small
\begin{tabular}{llccccc}
\toprule
 &&&& \hspace{-0.3em}IIIT5K\hspace{-0.3em} & \hspace{-0.3em}Hateful\hspace{-0.3em} & \\
 && \hspace{-0.3em}MNIST\hspace{-0.3em} & \hspace{-0.3em}SVHN\hspace{-0.3em} & \hspace{-0.3em}1k\hspace{-0.3em} & \hspace{-0.3em}Memes\hspace{-0.3em} & \hspace{-0.3em}SST-2\hspace{-0.3em} \\
\midrule
\multirow{3}{*}{\rot{Finetune}} & SOTA & \textbf{99.8}\xa & \textbf{96.4}\xb & \textbf{98.9}\xc & \textbf{78.0}\xd & \textbf{97.5}\xe \\
& JOINT\xf & - & - & 89.6 & - & - \\
& CBoW\xg & - & - & - & - & 80.0 \\
\midrule
\multirow{3}{*}{\rot{Linear}} & Raw Pixels & 92.5 & - & - & - & - \\
& ES Best & 98.9\xh & - & - & 58.6\xh & 59.0\xxi \\ 
& CLIP & 99.2 & - & - & 77.3 & 80.5 \\
\midrule
\rot{ZS} & CLIP & 88.4 & 51.0 & 90.0 & 63.3 & 67.9 \\
\bottomrule
\end{tabular}
\caption{OCR performance on 5 datasets. All metrics are accuracy on the test set except for Hateful Memes which reports ROC AUC on the dev set. Single model SOTA reported to best of knowledge. \textit{ES Best} reports the best performance across the 56 non-CLIP models in our evaluation suite. $^a$\citep{assiri2020stochastic} $^b$\citep{jaderberg2015spatial}  $^c$\citep{wang2020all} $^d$\citep{lippe2020multimodal} $^f$\citep{jaderberg2014deep} $^g$\citep{wang2018glue} $^h$\citep{xie2020self} $^i$\citep{mahajan2018exploring}}
\label{table:ocr}
\end{center}
\vskip -0.1in
\end{table}

\subsection{Optical Character Recognition}

Although visualizations have shown that ImageNet models contain features that respond to the presence of text in an image \citep{zeiler2014visualizing}, these representations are not sufficiently fine-grained to use for the task of optical character recognition (OCR). To compensate, models are augmented with the outputs of custom OCR engines and features to boost performance on tasks where this capability is required \citep{singh2019towards,yang2020tap}. Early during the development of CLIP, we noticed that CLIP began to learn primitive OCR capabilities which appeared to steadily improve over the course of the project. To evaluate this qualitatively noticed behavior, we measured performance on 5 datasets requiring the direct and indirect use of OCR. Three of these datasets MNIST \citep{lecun1998mnist}, SVHN \citep{netzer2011reading}, and IIIT5K \citep{mishra2012scene} directly check the ability of a model to perform low-level character and word recognition, while Hateful Memes \citep{kiela2020hateful} and SST-2 \citep{socher2013recursive} check the ability of a model to use OCR to perform a semantic task. Results are reported in Table \ref{table:ocr}. 

CLIP's performance is still highly variable and appears to be sensitive to some combination of the domain (rendered or natural images) and the type of text to be recognized (numbers or words). CLIP's OCR performance is strongest Hateful Memes and SST-2 - datasets where the text is digitally rendered and consists mostly of words. On IIIT5K, which is natural images of individually cropped words, zero-shot CLIP performs a bit more respectively and its performance is similar to \citet{jaderberg2014deep} early work combining deep learning and structured prediction to perform open-vocabulary OCR. However, performance is noticeably lower on two datasets involving recognition of hand written and street view numbers. CLIP's 51\% accuracy on full number SVHN is well below any published results. Inspection suggests CLIP struggles with repeated characters as well as the low resolution and blurry images of SVHN. CLIP's zero-shot MNIST performance is also poor and is outperformed by supervised logistic regression on raw pixels, one of the simplest possible machine learning baselines.

SST-2 is a sentence level NLP dataset which we render into images. We include SST-2 in order to check whether CLIP is able to convert low level OCR capability into a higher level representation. Fitting a linear classifier on CLIP's representation of rendered sentences achives 80.5\% accuracy. This is on par with the 80\% accuracy of a continuous bag of words baseline using GloVe word vectors pre-trained on 840 billion tokens \citep{pennington2014glove}. While this is a simple NLP baseline by today's standard, and well below the 97.5\% of the current SOTA, it is encouraging to see that CLIP is able to turn an image of rendered text into a non-trivial sentence level representation. Fully supervised CLIP is also surprisingly strong on Hateful Meme detection, where CLIP is only 0.7 points behind the current single model SOTA and several points above the best baseline from the original paper. Similar to SST-2, these other results on Hateful Memes use the ground truth text which CLIP does not have access to. Finally, we note that zero-shot CLIP outperforms the best results using fully supervised linear probes across all other 56 models included in our evaluation suite. This suggests CLIP's OCR capability is at least somewhat unique compared to existing work on self-supervised and supervised representation learning.

\begin{table}[ht]
\vskip 0.15in
\begin{center}
\small
\begin{tabular}{p{0.1cm}lcccc}
\toprule
 && UCF101 & K700 & \multicolumn{2}{c}{RareAct} \\
 && {\scriptsize Top-1} & {\scriptsize AVG } & {\scriptsize mWAP} & {\scriptsize mWSAP} \\
\midrule
\multirow{4}{*}{\rot{Finetune}} & R(2+1)D-BERT\xa & \textbf{98.7} & - & - & - \\
& NS ENet-L2$^b$ & - & \textbf{84.8} & - & - \\
& HT100M S3D\xd & 91.3 & - & - & - \\ 
& Baseline I3D\xe & - & 70.2 & - & - \\
\midrule
\multirow{3}{*}{\rot{Linear}} & MMV FAC\xf & 91.8 & - & - & - \\
& NS ENet-L2$^c$ & 89.4\xc & 68.2\xc & - & - \\
& CLIP & 92.0 & 73.0 & - & - \\
\midrule
\multirow{2}{*}{\rot{ZS}} & HT100M S3D\xd & - & - & 30.5 & 34.8 \\
& CLIP & 80.3 & 69.6 & \textbf{40.7} & \textbf{44.8} \\
\bottomrule
\end{tabular}
\caption{Action recognition performance on 3 video datasets. Single model SOTA reported to best of knowledge. Note that \textit{linear CLIP} and \textit{linear NS ENet-L2} are trained and evaluated on a single frame subsampled version of each dataset and not directly comparable to prior work. On Kinetics-700, we report the ActivityNet competition metric which is the average of top-1 and top-5 performance. $^a$\citep{kalfaoglu2020late} $^b$\citep{lu2020leveraging} $^c$\citep{xie2020self} $^d$\citep{miech2020end} $^e$\citep{carreira2019kinetics700} $^f$\citep{alayrac2020self}}
\label{table:action}
\end{center}
\vskip -0.1in
\end{table}

\subsection{Action Recognition in Videos}

For the purpose of learning, a potentially important aspect of natural language is its ability to express, and therefore supervise, an extremely wide set of concepts. A CLIP model, since it is trained to pair semi-arbitrary text with images, is likely to receive supervision for a wide range of visual concepts involving both common and proper nouns, verbs, and adjectives. ImageNet-1K, by contrast, only labels common nouns. Does the lack of broader supervision in ImageNet result in weaker transfer of ImageNet models to tasks involving the recognition of visual concepts that are not nouns?

To investigate this, we measure and compare the performance of CLIP and ImageNet models on several video action classification datasets which measure the ability of a model to recognize verbs. In Table \ref{table:action} we report results on UCF-101 \citep{soomro2012ucf101} and Kinetics-700 \citep{carreira2019kinetics700}, two common datasets for the task. Unfortunately, our CPU based linear classifier takes a prohibitively long time to evaluate on a video dataset due to the very large number of training frames. To deal with this, we aggressively sub-sample each video to only a single center frame, effectively turning it into an image classification dataset. As a result, our reported performance in a linear evaluation setting likely under estimates performance by a moderate amount.

\begin{table*}[ht]
\vskip 0.15in
\small
\begin{center}
\begin{tabular}{lcccccccccc}
\toprule
& IN & IN-V2 & IN-A & IN-R & ObjectNet & IN-Sketch & \multicolumn{2}{c}{IN-Vid} & \multicolumn{2}{c}{YTBB} \\
& Top-1 & Top-1 & Top-1 & Top-1 & Top-1 & Top-1 & PM0 & PM10 & PM0 & PM10 \\
\midrule
NS EfficientNet-L2\xa & \textbf{88.3} & \textbf{80.2} & \textbf{84.9} & 74.7 & 68.5 & 47.6 & 88.0 & 82.1 & 67.7 & 63.5 \\
FixResNeXt101-32x48d V2\xb & 86.4 & 78.0 & 68.4 & 80.0 & 57.8 & 59.1 & 85.8 & 72.2 & 68.9 & 57.7 \\
Linear Probe CLIP & 85.4 & 75.9 & 75.3 & 84.2 & 66.2 & 57.4 & 89.1 & 77.2 & 68.7 & 63.1 \\
Zero-Shot CLIP & 76.2 & 70.1 & 77.2 & \textbf{88.9} & \textbf{72.3} & \textbf{60.2} & \textbf{95.3} & \textbf{89.2} & \textbf{95.2} & \textbf{88.5} \\
\bottomrule
\end{tabular}
\caption{Detailed ImageNet robustness performance. IN is used to abbreviate for ImageNet. $^a$\citep{xie2020self} $^b$\citep{touvron2019fixing}}
\label{table:robustness}
\end{center}
\vskip -0.1in
\end{table*}

Despite this handicap, CLIP features transfer surprisingly well to this task. CLIP matches the best prior result on UCF-101 in a linear probe evaluation setting and also outperforms all other models in our evaluation suite. On Kinetics-700, CLIP also outperforms the fine-tuned I3D baseline from the original paper. Since it does not require a training stage, we report CLIP's zero-shot performance when averaging predictions across all frames. CLIP also performs well in this setting and on Kinetics-700 its performance is within 1\% of the fully supervised I3D baseline which is trained on 545000 labeled videos. Encouraged by these results, we also measure CLIP's performance on the recently introduced RareAct dataset \citep{miech2020rareact} which was designed to measure zero-shot recognition of unusual actions like ``hammering a phone'' and ``drilling an egg''. CLIP improves over the prior state of the art, a S3D model trained on automatically extracted captions from 100 million instructional videos, by 10 points.

While CLIP has encouragingly strong performance on the task of action recognition, we note that there are many differences between the models being compared beyond just their form of supervision such as model architecture, training data distribution, dataset size, and compute used. Further work is needed to more precisely determine what specific design decisions contribute to achieving high performance on this task.

\begin{table}[ht]
\vskip 0.15in
\small
\begin{center}
\begin{tabular}{lccccc}
\toprule
& 1km & 25km & 200km & 750km & 2500km \\
\midrule
ISNs\xa & \textbf{16.9} & \textbf{43.0} & \textbf{51.9} & \textbf{66.7} & \textbf{80.2} \\
CPlaNet\xb & 16.5 & 37.1 & 46.4 & 62.0 & 78.5\\
CLIP & 13.9 & 32.9 & 43.0 & 62.0 & 79.3 \\
Deep-Ret+\xc & 14.4 & 33.3 & 47.7 & 61.6 & 73.4 \\
PlaNet\xd & 8.4 & 24.5 & 37.6 & 53.6 & 71.3 \\
\bottomrule
\end{tabular}
\caption{Geolocalization performance on the IM2GPS test set. Metric is percent of images localized within a given radius. Models are ordered by average performance. $^a$\citep{muller2018geolocation} $^b$\citep{hongsuck2018cplanet} $^c$\citep{vo2017revisiting} $^c$\citep{weyand2016planet}}
\label{table:geolocalization}
\end{center}
\vskip -0.1in
\end{table}

\subsection{Geolocalization}

Another behavior we noticed during the development of CLIP was its ability to recognize many places and locations. To quantify this we created the Country211 dataset as described in Appendix \ref{sec:linear-probe} and report results on it throughout the paper. However it is a new benchmark so to compare with prior work on geolocalization we also report results on the IM2GPS test set from \citet{hays2008im2gps} in Table \ref{table:geolocalization}. Since IM2GPS is a regression benchmark, we guess the GPS coordinates of the nearest image in a set of reference images using CLIP's embedding space. This is not a zero-shot result since it uses nearest-neighbor regression. Despite querying only 1 million images, which is much less than prior work, CLIP performs similarly to several task specific models. It is not, however, competitive with the current state of the art.

\subsection{Robustness to Distribution Shift}

Section \ref{subsection:robustness} provides a high level summary and analysis of ImageNet-related robustness results. We briefly provide some additional numerical details in this appendix. Performance results per dataset are provided in Table \ref{table:robustness} and compared with the current state of the art results reported in \citet{taori2020measuring}'s evaluation suite. Zero-shot CLIP improves the state of the art on 5 of the 7 datasets, ImageNet-R, ObjectNet, ImageNet-Sketch, ImageNet-Vid, and Youtube-BB. CLIP's improvements are largest on ImageNet-Vid and Youtube-BB due to its flexible zero-shot capability and on ImageNet-R, which likely reflects CLIP's pre-training distribution including significant amounts of creative content. A similar behavior has been documented for the Instagram pre-trained ResNeXt models as discussed in \citet{taori2020measuring}.

\clearpage

\section{Model Hyperparameters}

\begin{table}[h!]
\begin{minipage}{\textwidth}
\centering

\begin{tabular}{l|c} \toprule
    Hyperparameter  & Value \\ \midrule
    Batch size & 32768 \\
    Vocabulary size & 49408 \\
    Training epochs & 32 \\
    Maximum temperature & 100.0 \\
    Weight decay & 0.2 \\
    Warm-up iterations & 2000 \\
    Adam $\beta_1$ & 0.9 \\
    Adam $\beta_2$ & 0.999 (ResNet), 0.98 (ViT)\\
    Adam $\epsilon$ & $10^{-8}$ (ResNet), $10^{-6}$ (ViT)\\
    \bottomrule
\end{tabular}
\caption{Common CLIP hyperparameters}

\vspace{2em}

\small
\begin{tabular}{l|cccccccc} \toprule
          & Learning & Embedding & Input      & \multicolumn{2}{c}{ResNet}  & \multicolumn{3}{c}{Text Transformer} \\
    Model & rate & dimension & resolution & blocks & width  & layers & width & heads  \\ \midrule
    RN50  & $5 \times 10^{-4}$ & 1024 & 224 & (3, 4, 6, 3) & 2048 & 12 & 512 & 8 \\
    RN101 & $5 \times 10^{-4}$ & 512 & 224 & (3, 4, 23, 3) & 2048 & 12 & 512 & 8 \\
    RN50x4 & $5 \times 10^{-4}$ & 640 & 288 & (4, 6, 10, 6) & 2560 & 12 & 640 & 10 \\
    RN50x16 & $4 \times 10^{-4}$ & 768 & 384 & (6, 8, 18, 8) & 3072 & 12 & 768 & 12 \\
    RN50x64 & $3.6 \times 10^{-4}$ & 1024 & 448 & (3, 15, 36, 10) & 4096 & 12 & 1024 & 16 \\ 
    \bottomrule
\end{tabular}
\caption{CLIP-ResNet hyperparameters}

\vspace{2em}

\small
\begin{tabular}{l|ccccccccc} \toprule
          & Learning & Embedding & Input      & \multicolumn{3}{c}{Vision Transformer} & \multicolumn{3}{c}{Text Transformer} \\
    Model & rate & dimension & resolution & layers & width & heads  & layers & width & heads \\ \midrule
    ViT-B/32 & $5 \times 10^{-4}$ & 512 & 224 & 12 & 768 & 12 & 12 & 512 & 8 \\
    ViT-B/16 & $5 \times 10^{-4}$ & 512 & 224 & 12 & 768 & 12 & 12 & 512 & 8 \\
    ViT-L/14 & $4 \times 10^{-4}$ & 768 & 224 & 24 & 1024 & 16 & 12 & 768 & 12 \\
    ViT-L/14-336px & $2 \times 10^{-5}$ & 768 & 336 & 24 & 1024 & 16 & 12 & 768 & 12 \\
    \bottomrule
\end{tabular}
\caption{CLIP-ViT hyperparameters}

\end{minipage}
\end{table}

\end{document}